\documentclass[final]{cvpr}

\pdfoutput=1

\usepackage{times}
\usepackage{epsfig}
\usepackage{graphicx}
\usepackage{amsmath}
\usepackage{amssymb}

\usepackage{enumitem}       
\usepackage[utf8]{inputenc} 
\usepackage[T1]{fontenc}    
\usepackage{url}            
\usepackage{booktabs}       
\usepackage{amsfonts}       
\usepackage{nicefrac}       
\usepackage{microtype}      

\usepackage{graphicx}
\usepackage{caption}
\usepackage{subcaption}
\usepackage{color}
\usepackage{bm}      	    
\usepackage{wrapfig}        
\usepackage{tabularx} 
\usepackage[numbers]{natbib} 
\usepackage{array} 
\usepackage{makecell}       

\usepackage{float} 
\usepackage{color}
\usepackage{subcaption}
\usepackage[ruled,noend]{algorithm2e} 

\usepackage{xcolor}         

\usepackage{algorithmic}    

\usepackage[pagebackref=true,breaklinks=true,colorlinks,bookmarks=false]{hyperref}

\def\cvprPaperID{3576} 
\def\confYear{CVPR 2021}

\begin{document}

\title{Towards Robust Classification Model by \\ Counterfactual and Invariant Data Generation}

\author{Chun-Hao Chang, George Alexandru Adam, Anna Goldenberg\\
University of Toronto, Vector Institute, The Hospital for Sick Children\\
{\tt\small \{kingsley.chang,alex.adam\}@mail.utoronto.ca,anna.goldenberg@utoronto.ca}
}

\maketitle

\begin{abstract}
Despite the success of machine learning applications in science, industry, and society in general, many approaches are known to be non-robust, often relying on spurious correlations to make predictions.
Spuriousness occurs when some features correlate with labels but are not causal;
relying on such features prevents models from generalizing to unseen environments where such correlations break. 
In this work, we focus on image classification and propose two data generation processes to reduce spuriousness. 
Given human annotations of the subset of the features responsible (causal) for the labels (e.g. bounding boxes), we modify this causal set to generate a surrogate image that no longer has the same label (i.e. a counterfactual image). 
We also alter non-causal features to generate images still recognized as the original labels, which helps to learn a model invariant to these features.
In several challenging datasets, our data generations outperform state-of-the-art methods in accuracy when spurious correlations break, and increase the saliency focus on causal features providing better explanations.
\end{abstract}

\section{Introduction}
What makes an image be labeled as a cat? What makes a doctor think there is a tumor in a CT scan? 
What makes a human label a movie review as positive or negative?
These questions are inherently causal, but typical machine learning models rely on associations between features and labels rather than causation.
Especially in high-dimensional feature spaces with strong correlations, learning which sets of features are right (causal) associations to predict targets becomes difficult, as different sets can result in the same best training accuracy.
Because of this, we see issues such as \emph{spurious correlations}~\citep{geirhos2020shortcut}, \emph{artifacts}~\citep{gururangan2018annotation}, \emph{lack of robustness}~\citep{agrawal2016analyzing, belinkov2018synthetic}, and \emph{discrimination}~\citep{kusner2017counterfactual} happening across many machine learning fields.


Spurious associations happen when factors correlate with labels but are not causal.
We might consider factors as spurious associations if \emph{intervening} on such factors would not change the resulting labels.
In the context of images, the backgrounds of images can be a source of spurious correlations with labels (e.g. a forest background correlates with bird label) because changing (intervening on) backgrounds should not affect the labels of the foreground classification.

\begin{figure}[tbp]
  \centering
  \includegraphics[width=\linewidth]{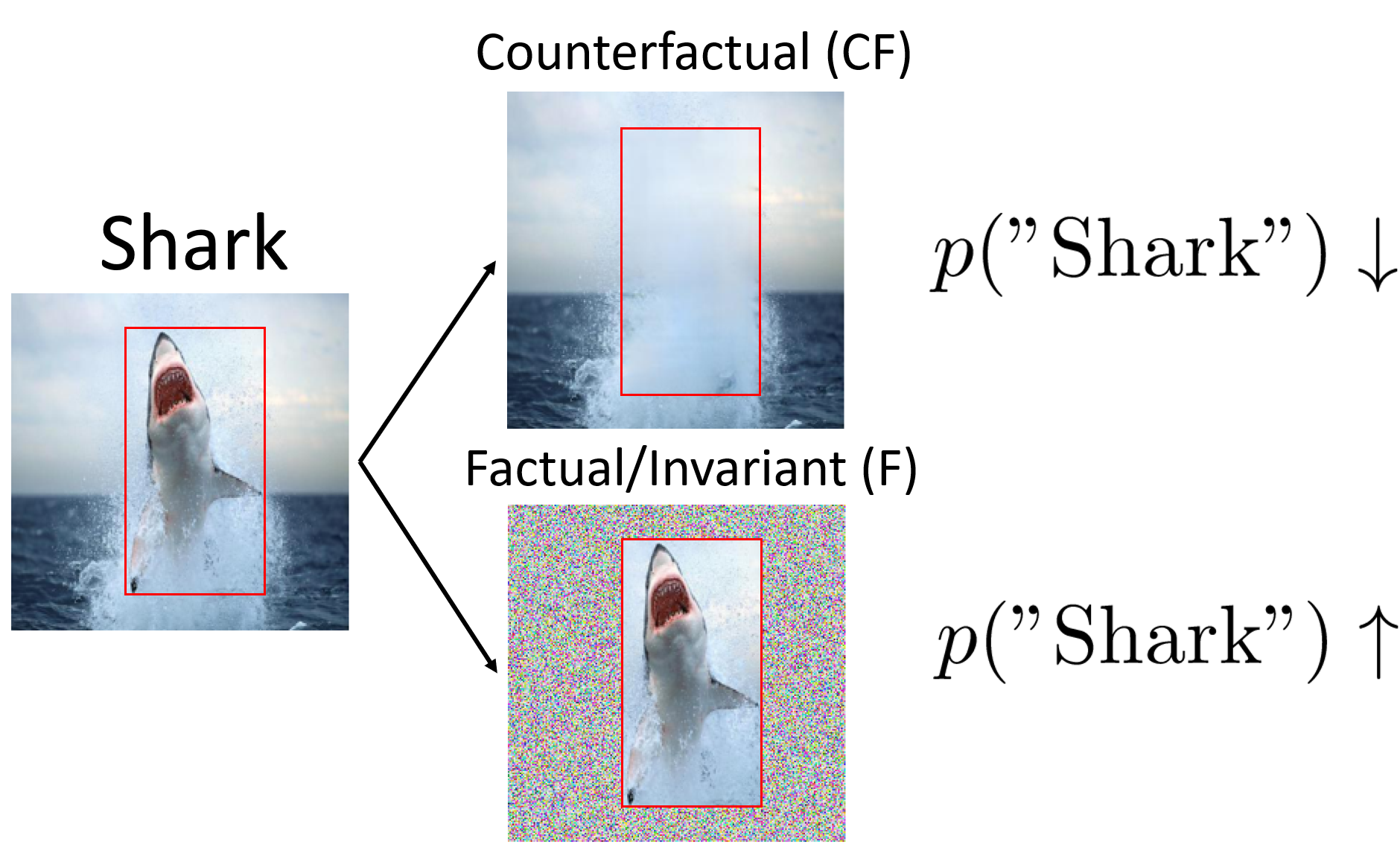}
 \caption{
    We propose $2$ data augmentations: Counterfactual (CF) and Factual (F). CF image keeps backgrounds and reduces the probability of target class, while F image keeps foreground and increases the probability.
  }
  \label{fig:fig1_diagram}
  \vspace{-5pt}
\end{figure}

In this paper, we aim to address such spurious associations in the typical ML classification framework by incorporating human causal knowledge.
Given a human rationale behind a labeling process (e.g. this part of the image is cat-like), we augment our datasets to break the correlations between backgrounds and labels in two ways.
First, we generate counterfactuals that ask "how can we modify the image such that a human would no longer label it as a cat?"
That is, by removing the causal features (foreground region containing the cat), and imputing it in a way that is consistent with the background, we generate the counterfactual image that would not be labeled by humans as a cat.
Second, we intervene on the non-causal factors (i.e. image backgrounds) to generate new images still containing the cat but with a modified background.
This helps the model be invariant to such factors.
We experiment on several large-scale datasets and show our methods consistently improve the accuracy and saliency focus on causal features.
Our contributions can be summarized as follows:
\begin{itemize}[leftmargin=*]
    \item We use various counterfactual and invariant data generations to augment training datasets which makes models more robust to spurious correlations.
    \item We show that our augmentations lead to similar or better accuracy than state-of-the-art saliency regularization and other robustness baselines on challenging datasets in the presence of background shifts. 
    We also find combining our augmentations with saliency regularization can further improve performance.
    \item Our methods have stronger salience focus on causal features that provide better explanations, although we find strong salience on causal features only correlates weakly with good generalization.
    
\end{itemize}

\section{Related Work}

Various works have found that standard machine learning models rely on spurious patterns to make predictions and do not generalize to unseen environments~\citep{geirhos2020shortcut}.
For instance, \citet{geirhos2018imagenet} found standard ImageNet-trained models classify images using object's texture rather than object's shape. 
Several medical imaging classifiers have also been shown to use spurious background to make predictions for COVID-19~\citep{maguolo2020critic} and other lung symptoms~\citep{zech2018confounding}.
Similarly, \citet{young2019deep} showed deep learning models for CT-scans, although having high accuracy, seem to produce explanations outside of the relevent regions when visualized by Grad-CAM~\citep{selvaraju2017grad} and Shap~\citep{lundberg2017unified}.
\citet{bissoto2019constructing} also found that models trained using public skin lesion datasets tend to have explanations outside of the human-labeled important region, questioning their abilities to generalize across other datasets.

Several methods have been proposed to remove known spurious correlations in concepts (e.g. gender or texture bias).
\citet{lu2020gender, zmigrod2019counterfactual} removed gender bias in text by swapping pronouns ("he" becomes "she") to augment the data.
\citet{geirhos2018imagenet} trained on augmented ImageNet datasets generated in different styles via style transfer to remove texture bias.
\citet{zemel2013learning, madras2018learning} directly penalized the model to prevent it from classifying sensitive concepts including race or sex to achieve fairness.
In this work we study a different case where the spurious and causal features are separated feature-wise which allows us to remove biases without knowing them in the first place.

Several previous works have attempted to solve the same problem with a different approach: they directly regularize the explanations (saliency) of the model to match the human-labeled important features.
\citet{ross2017right} were the first to propose regularizing the input gradients toward the causal features and showed improved robustness when the model was evaluated on a different test distribution.
\citet{erion2019learning} used the expected gradient (a stronger saliency method) and regularized it toward other forms of human priors (e.g. sparsity or smoothness).
\citet{rieger2020interpretations} proposed using Contextual Decomposition~\citep{singh2018hierarchical} which can regularize not just per-pixel saliency but the interactions of the pixels.
Several works also found regularizing saliency helps in text classification~\citep{du2019learning, ghaeini2019saliency} or medical imaging~\citep{zhuang2019care}.
In addition, \citet{ross2018improving} found that input gradient regularization improves adversarial robustness.
\citet{bao2018deriving, Mitsuhara2019EmbeddingHK} explored regularizing attention (instead of gradient) and showed improvement in text classification.
Despite all the aforementioned success, \citet{viviano2019underwhelming} reported an underwhelming relationship between controlling saliency maps and improving generalization performance in two large-scale medical imaging datasets.

Augmenting counterfactual data to remove spurious correlation has been investigated in the NLP domain \citep{Kaushik2020Learning}, but they relied on human efforts to generate counterfactual data.
Several works were also investigated in Visual Question Answering (VQA) fields by augmenting with counterfactual data that changes the answer~\citep{agarwal2020towards, chen2020counterfactual}.
Here we instead investigate the effect on the task of image classification, and explore various generation approaches including heuristics and generative models. 

\setlength\tabcolsep{0pt} 
\begin{figure*}[tbp]
  \centering
  \begin{tabular}{cccccccccc}
  & \multicolumn{5}{c}{Counterfactual (CF)} & \multicolumn{4}{c}{Factual (F)} \\
  \cmidrule(lr){2-6}\cmidrule(lr){7-10}
  None &
  CF(Grey) &
  CF(Random) &
  CF(Shuffle) &
  CF(Tile) &
  CF(CAGAN) &
  F(Random) &
  F(Shuffle) &
  F(Mixed-Rand) &
  F(FGSM) \\

  \includegraphics[width=0.1\linewidth]{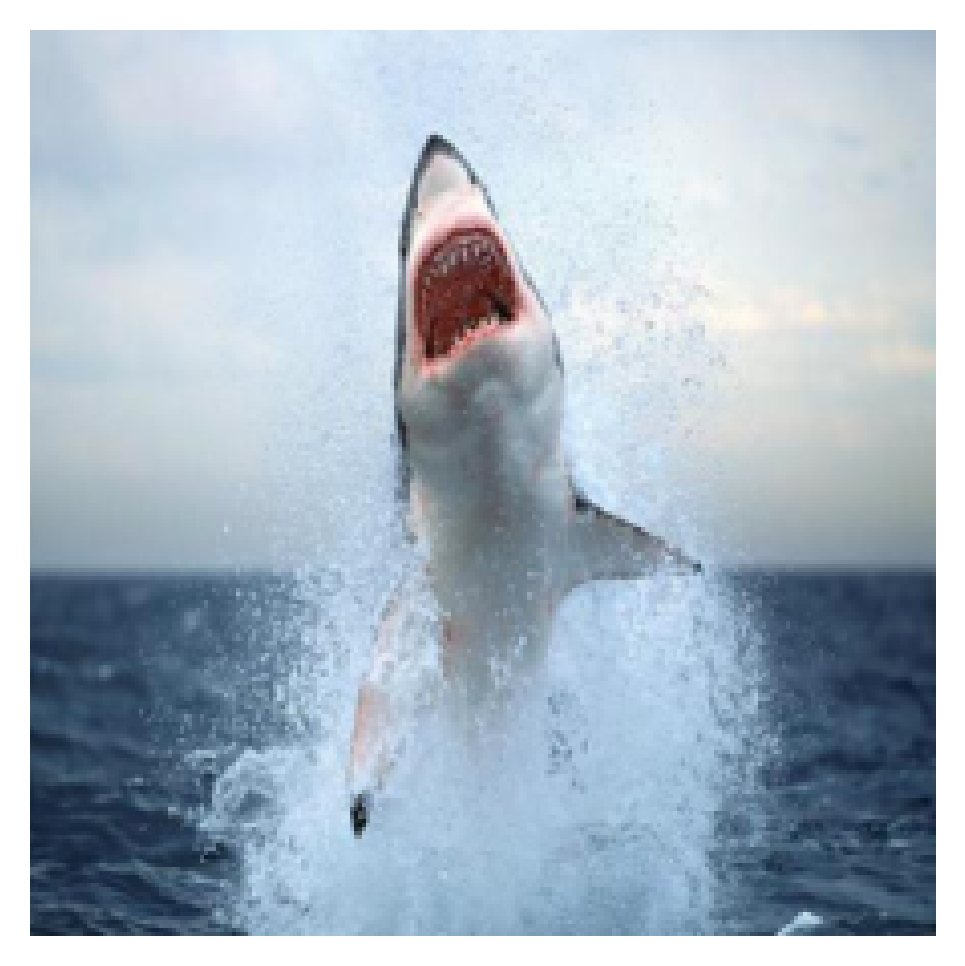} &
  \includegraphics[width=0.1\linewidth]{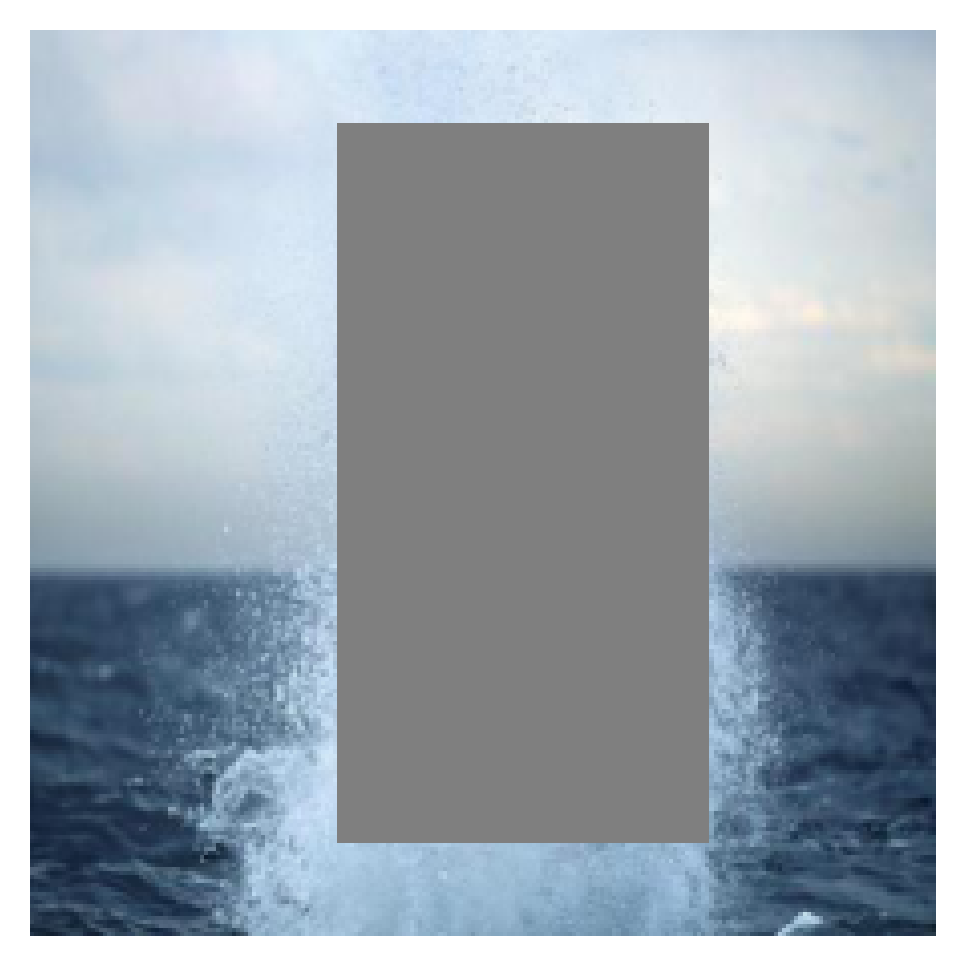} &
  \includegraphics[width=0.1\linewidth]{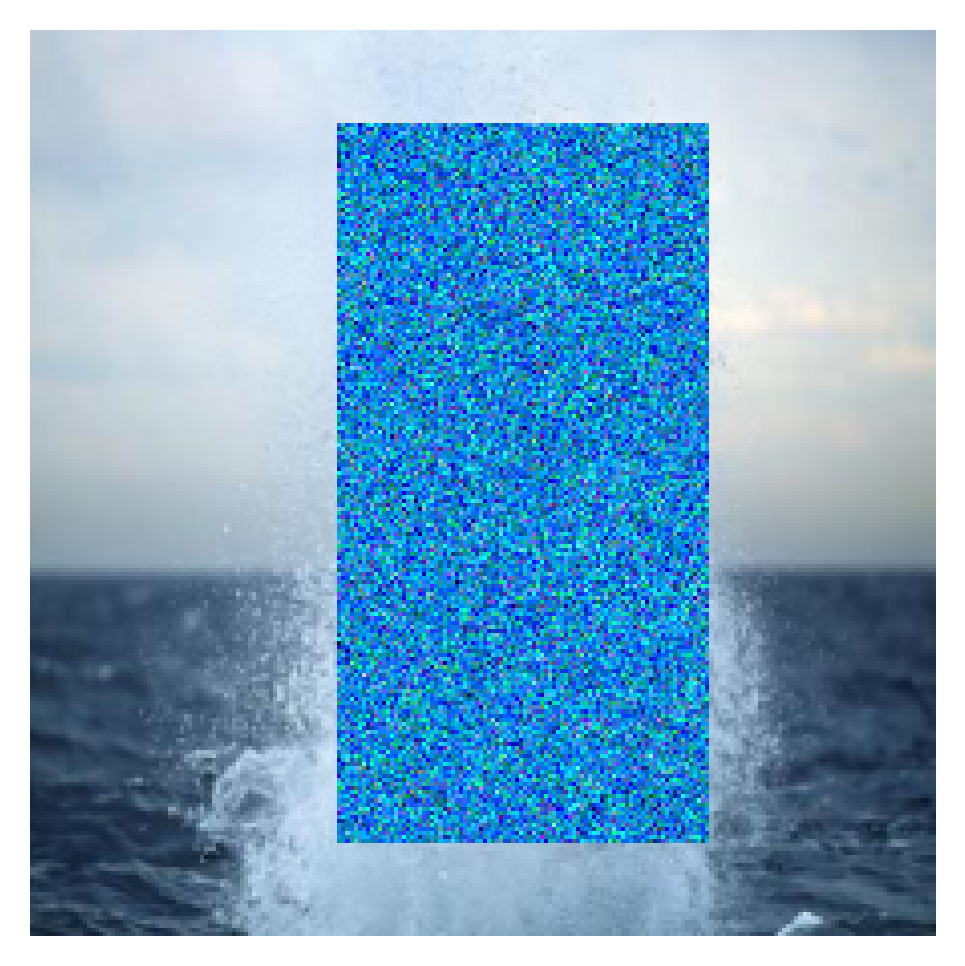} &
  \includegraphics[width=0.1\linewidth]{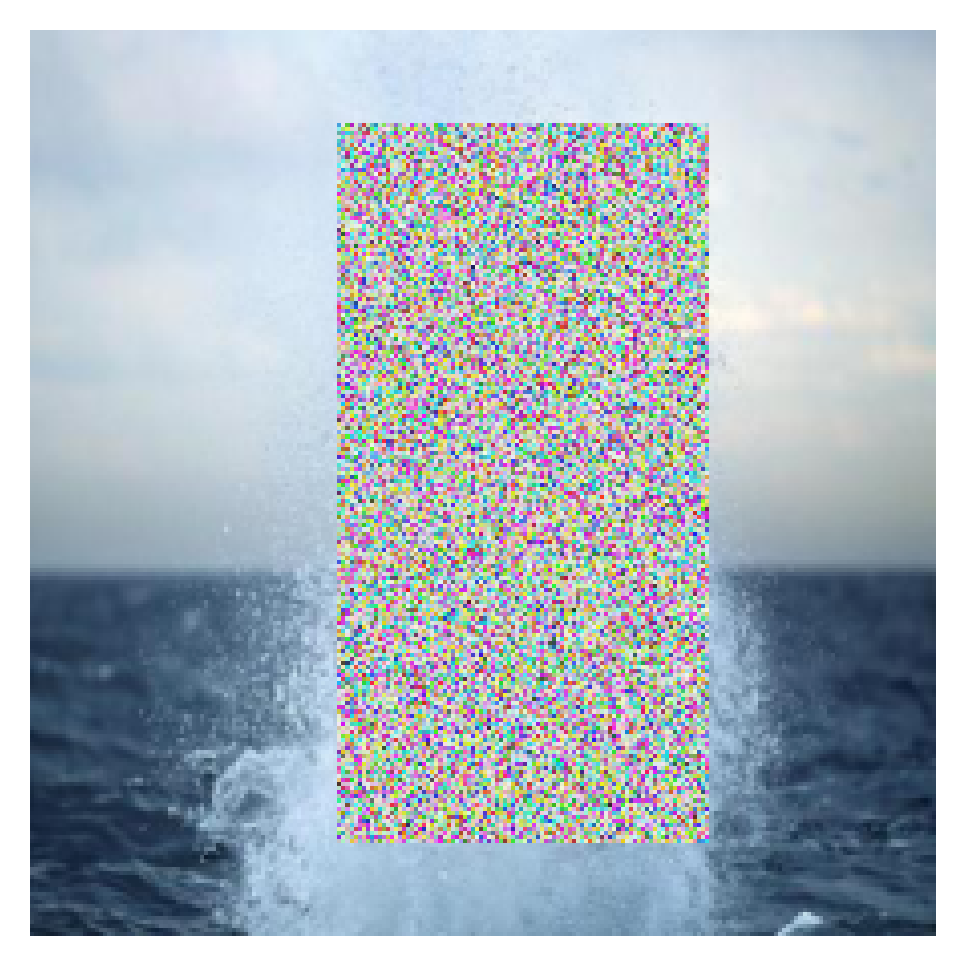} &
  \includegraphics[width=0.1\linewidth]{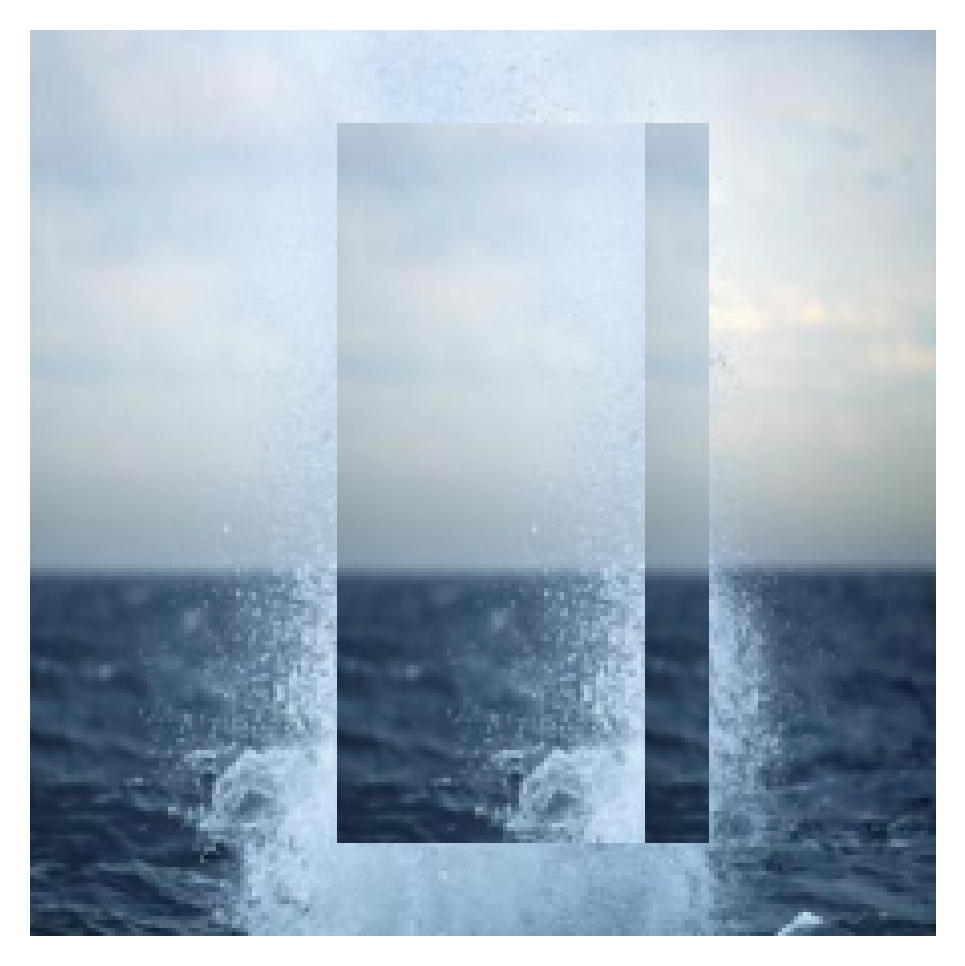} &
  \includegraphics[width=0.1\linewidth]{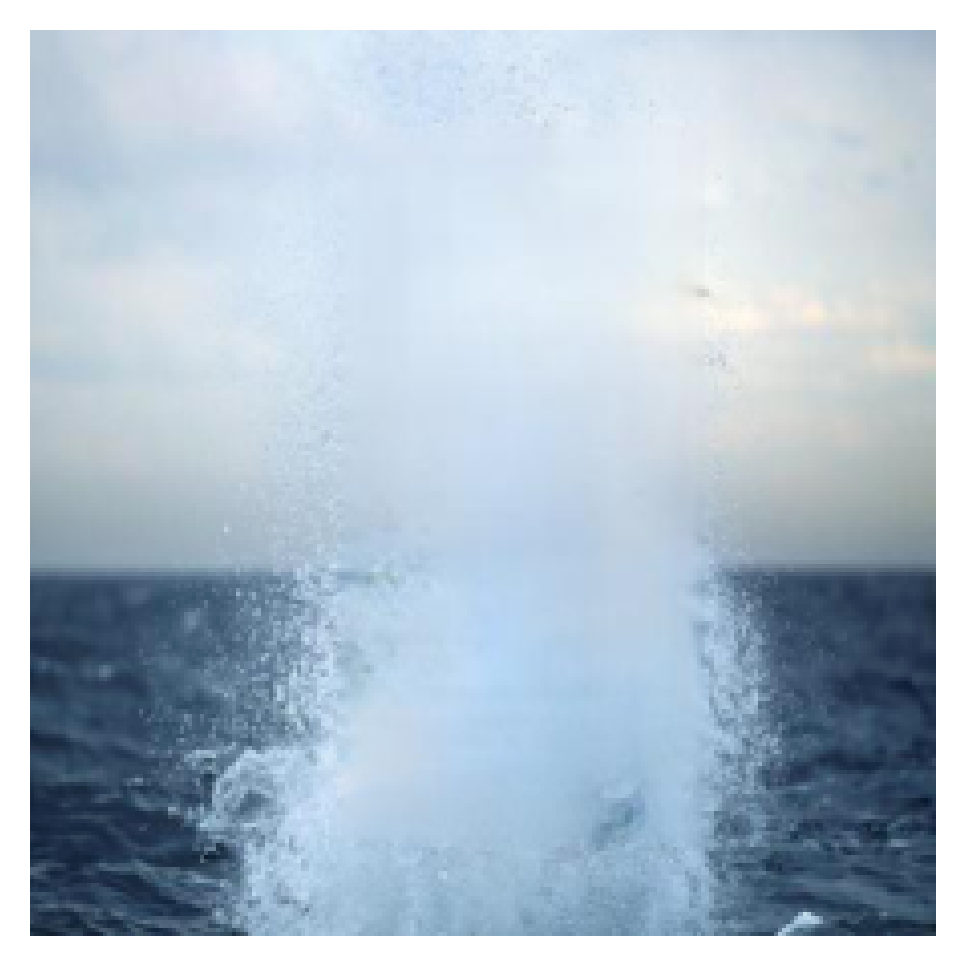} &
  \includegraphics[width=0.1\linewidth]{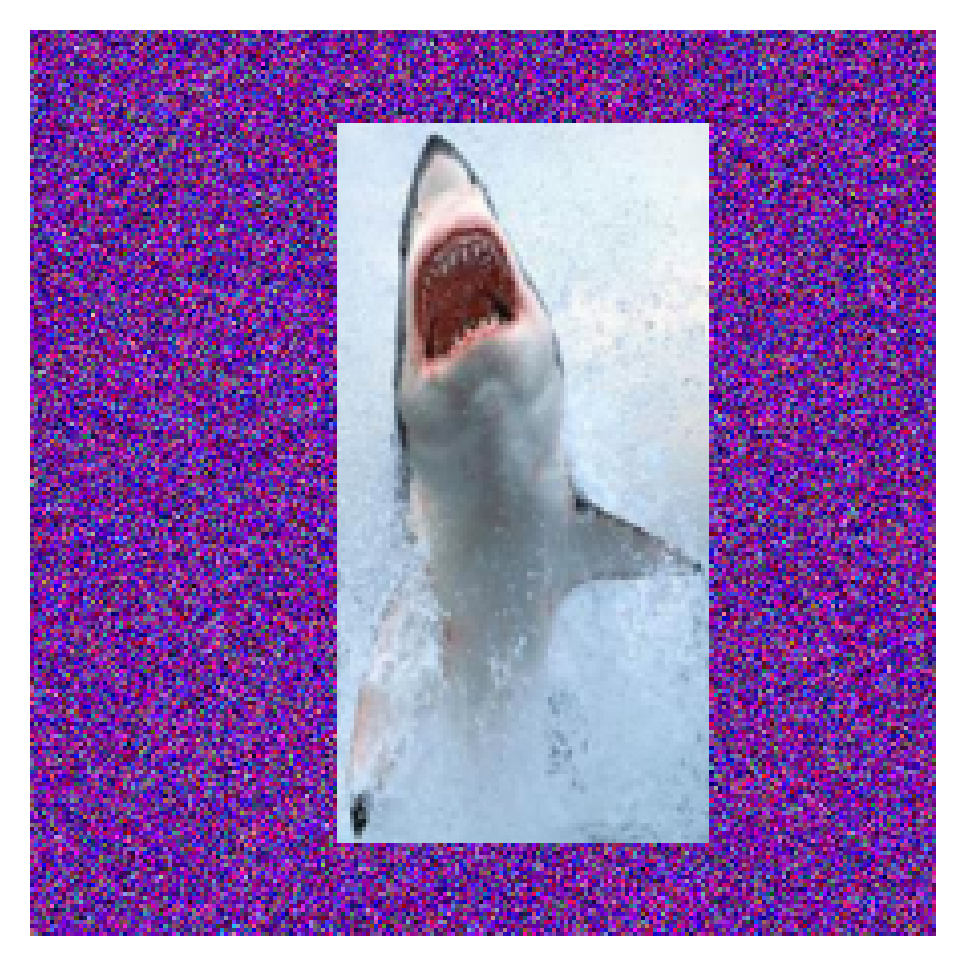} &
  \includegraphics[width=0.1\linewidth]{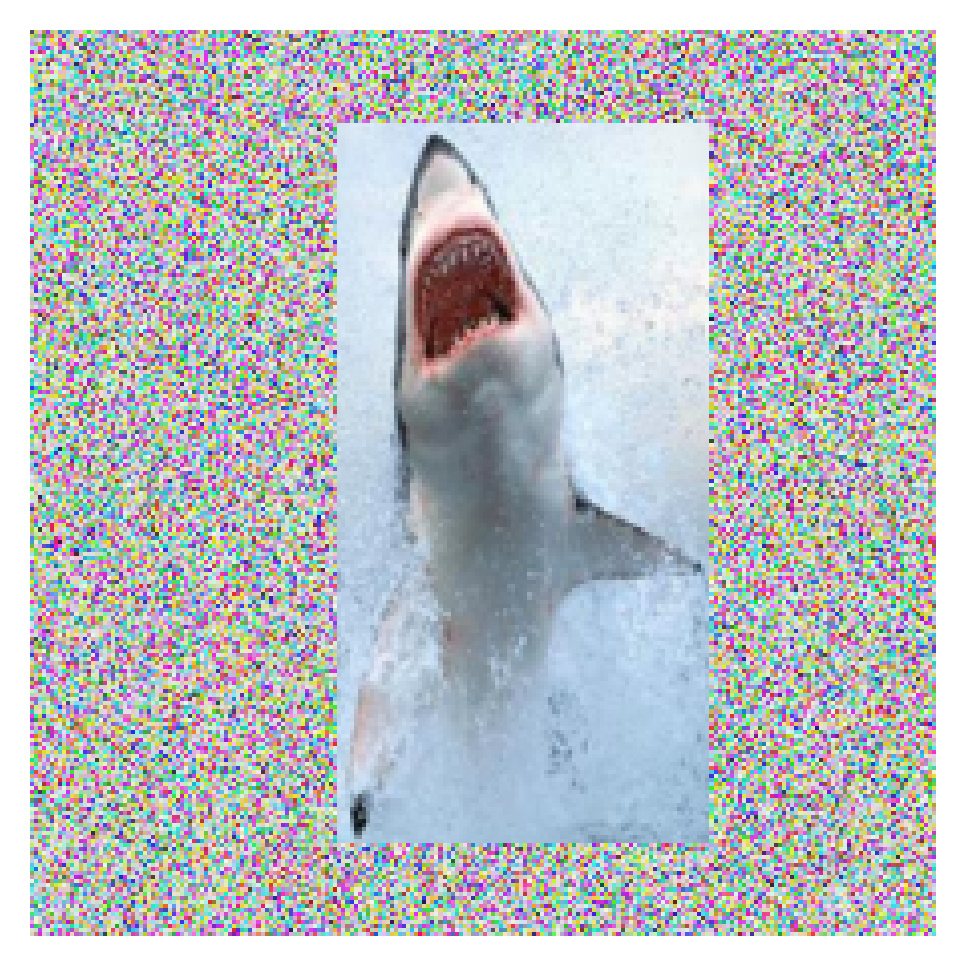} &
  \includegraphics[width=0.1\linewidth]{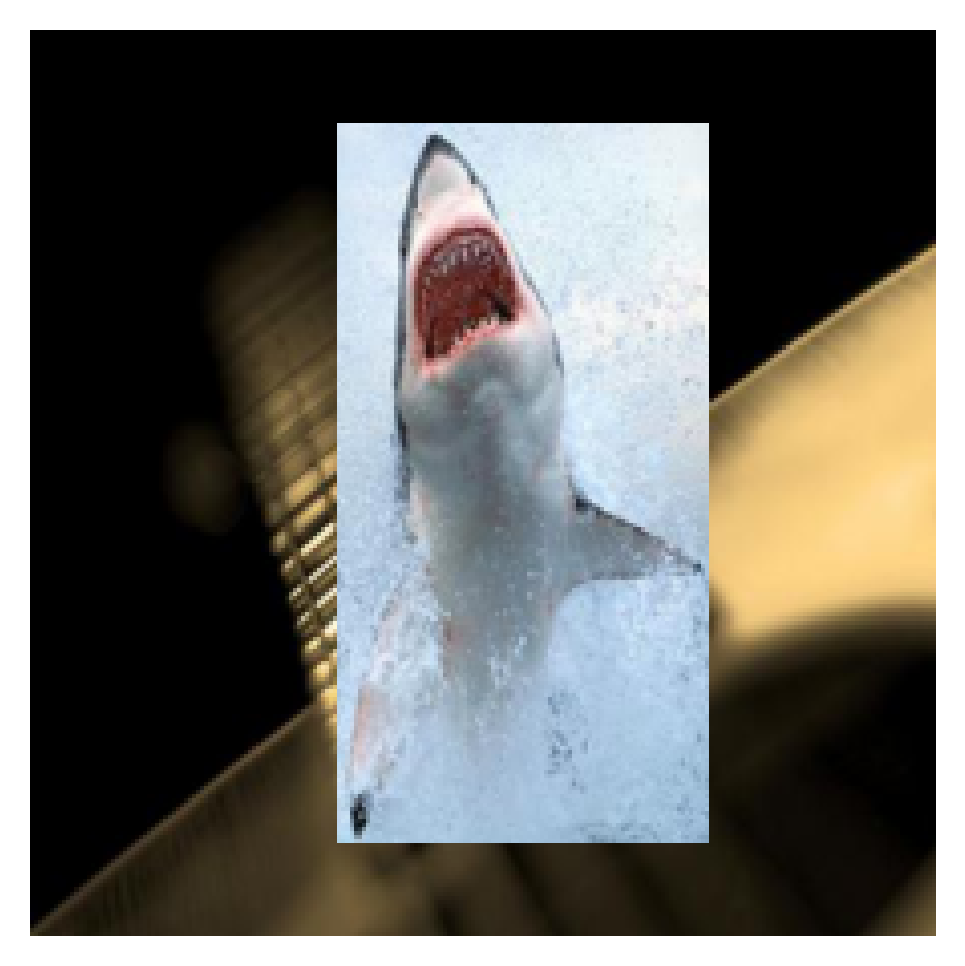} &
  \includegraphics[width=0.1\linewidth]{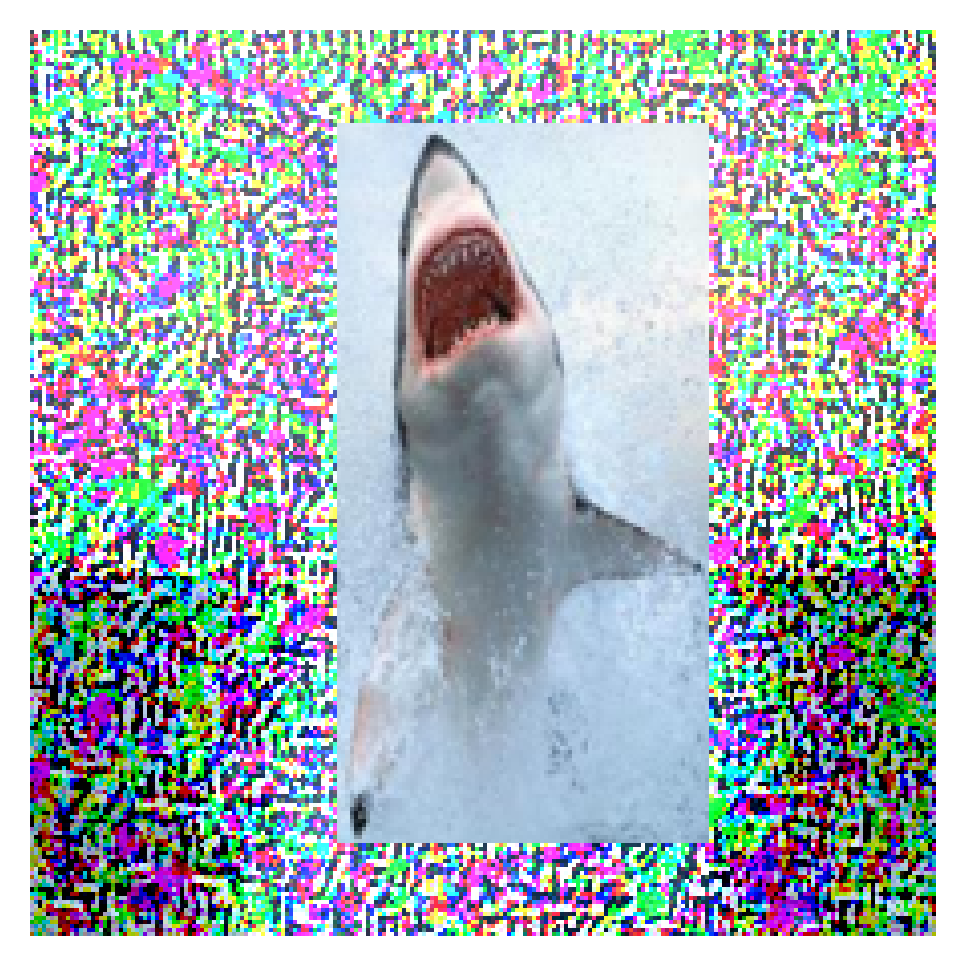} \\

  \includegraphics[width=0.1\linewidth]{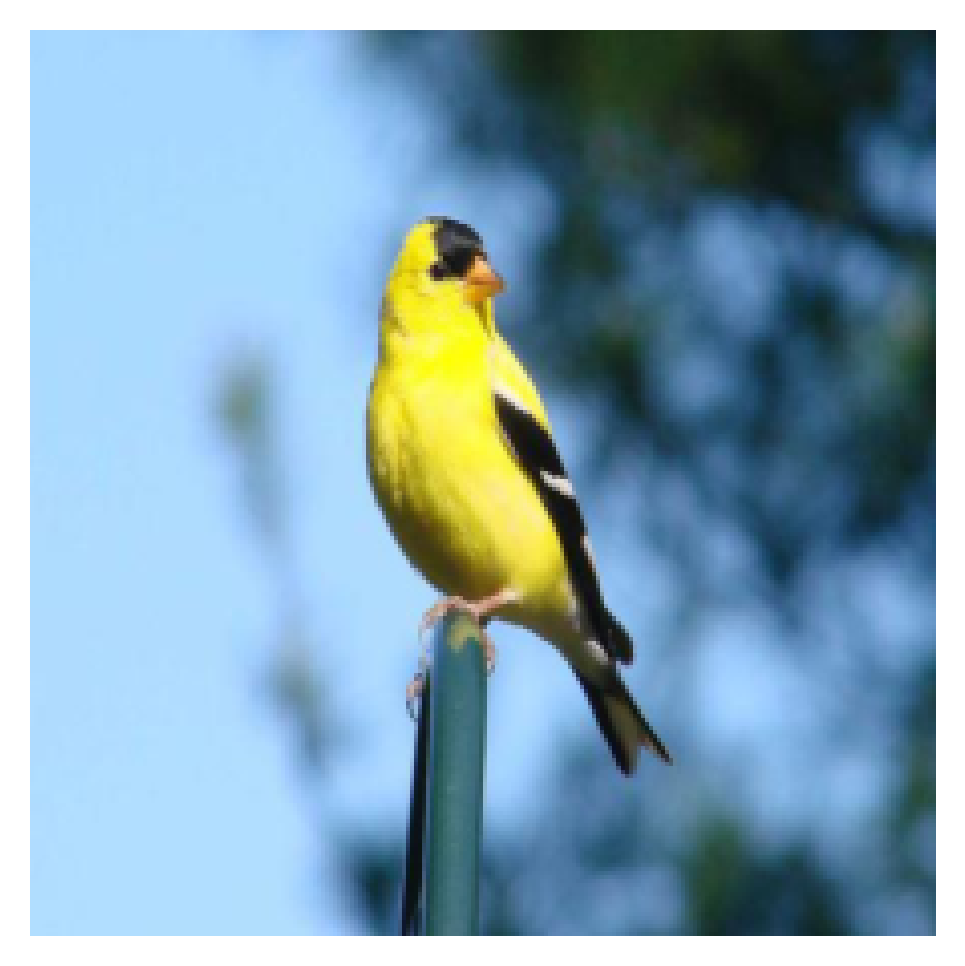} &
  \includegraphics[width=0.1\linewidth]{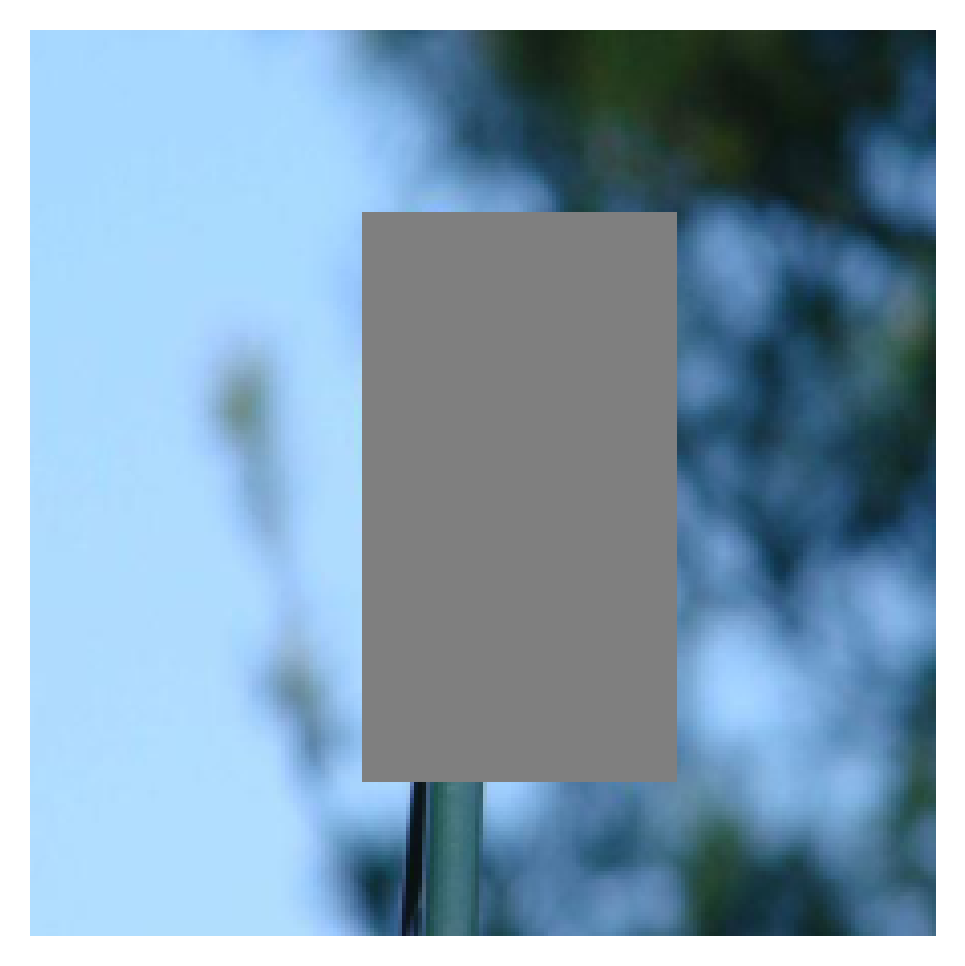} &
  \includegraphics[width=0.1\linewidth]{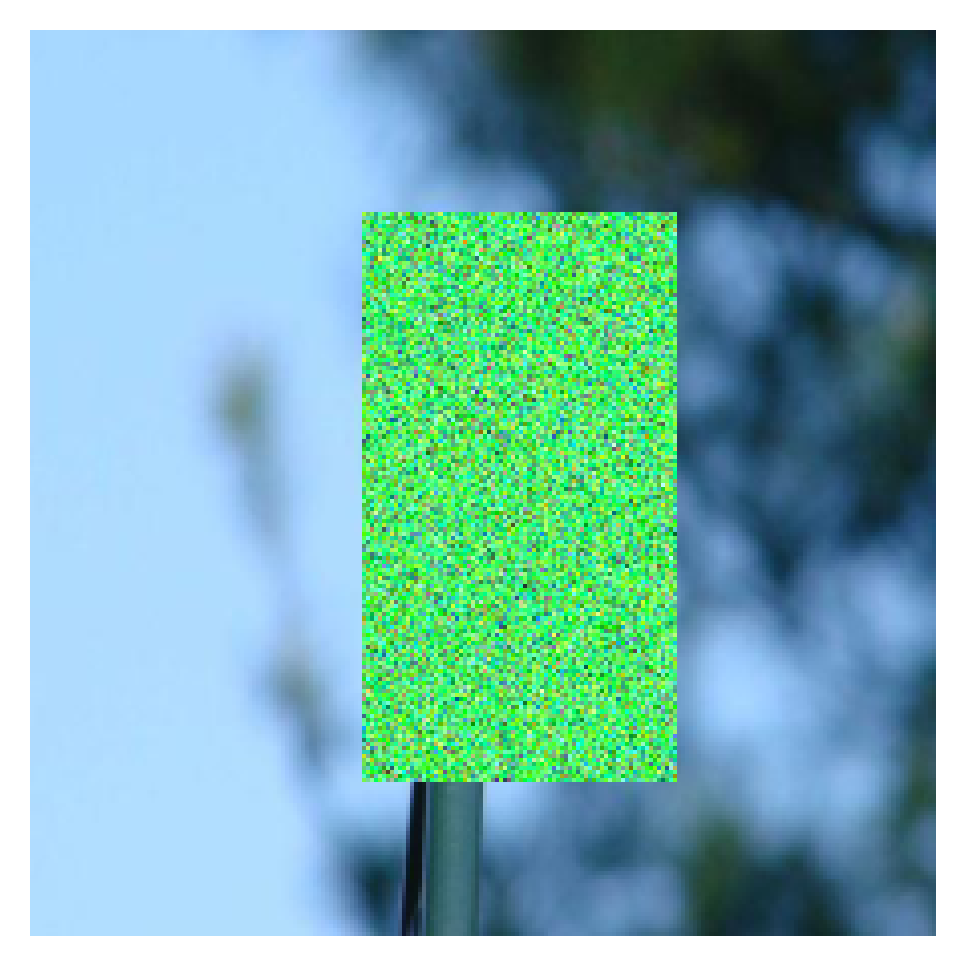} &
  \includegraphics[width=0.1\linewidth]{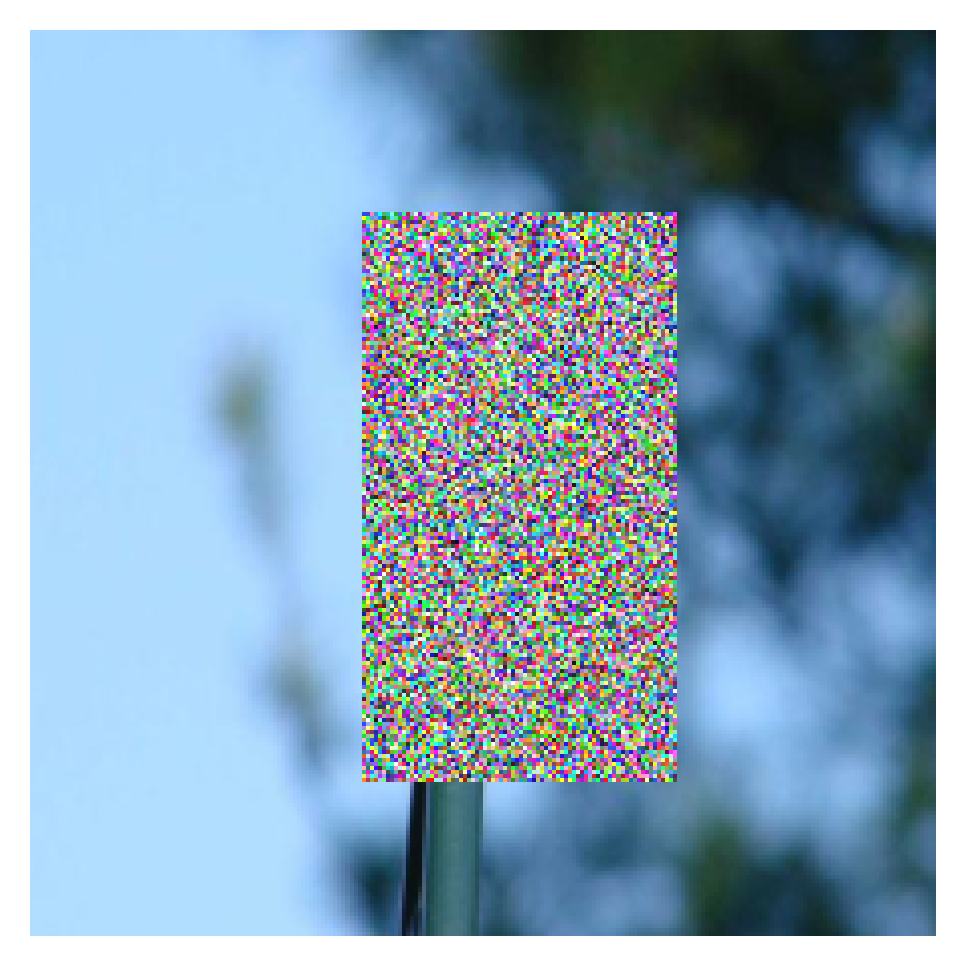} &
  \includegraphics[width=0.1\linewidth]{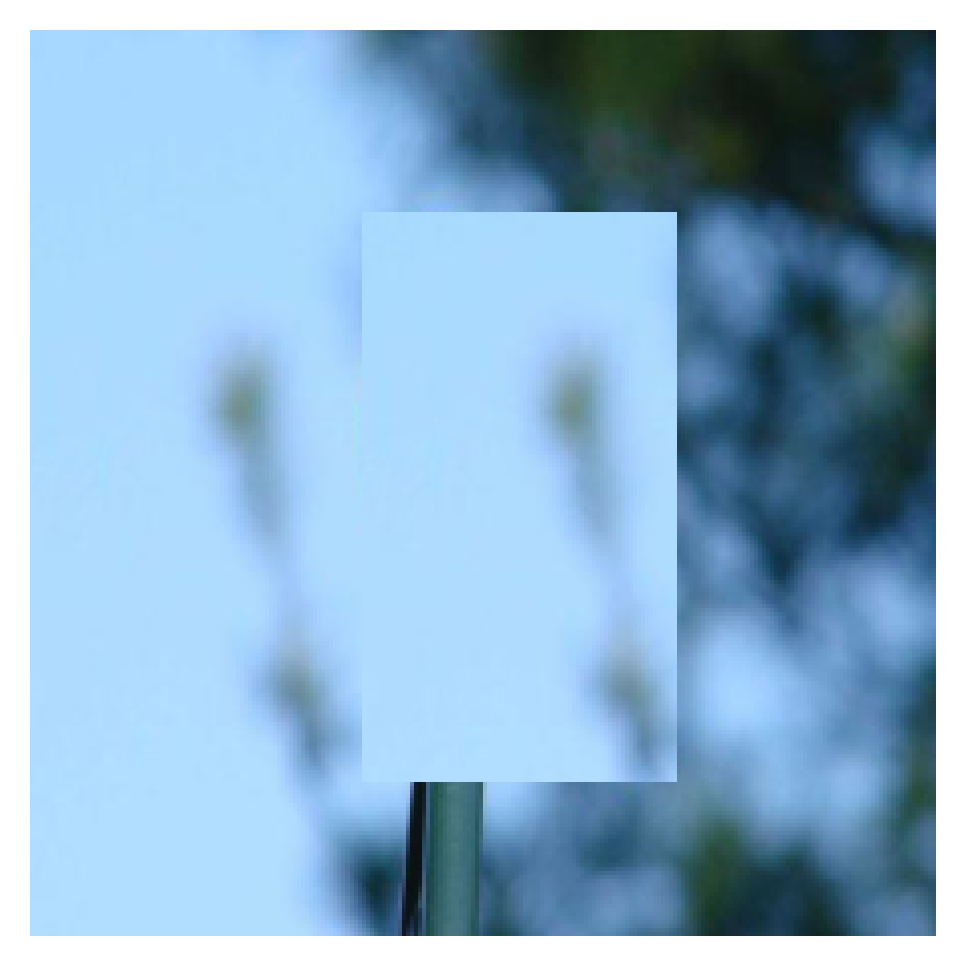} &
  \includegraphics[width=0.1\linewidth]{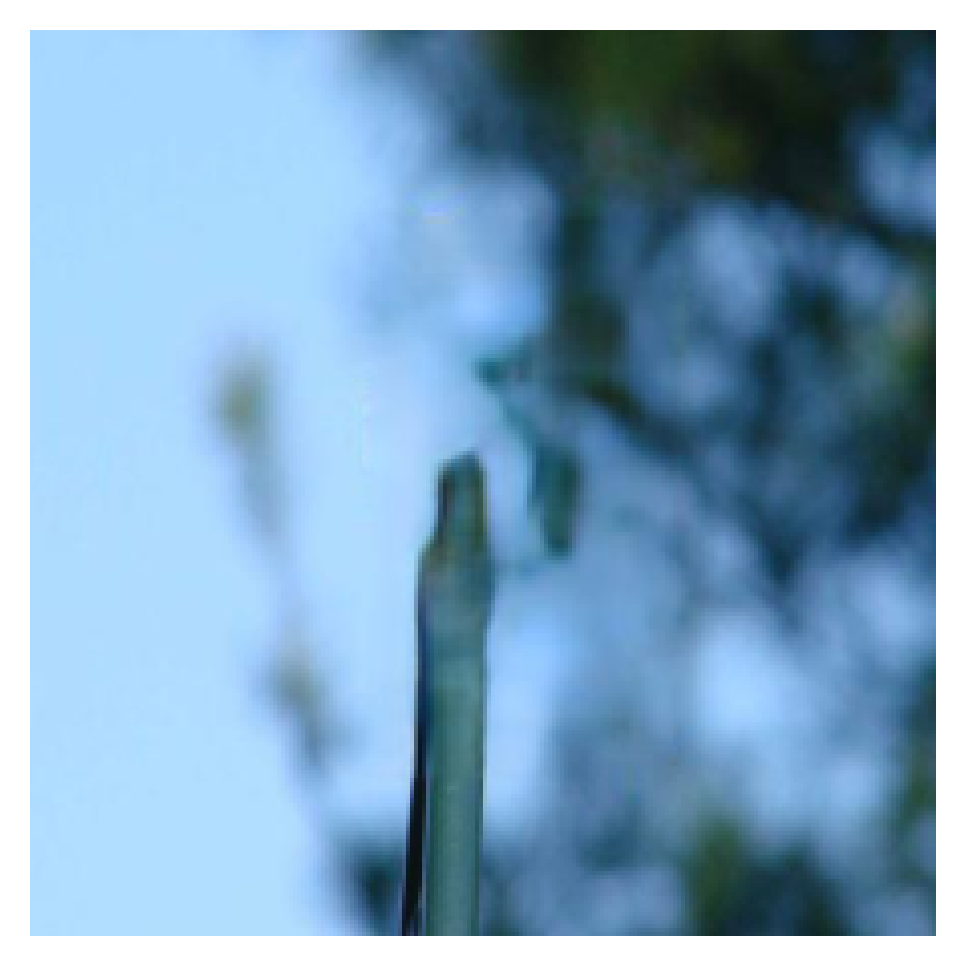} &
  \includegraphics[width=0.1\linewidth]{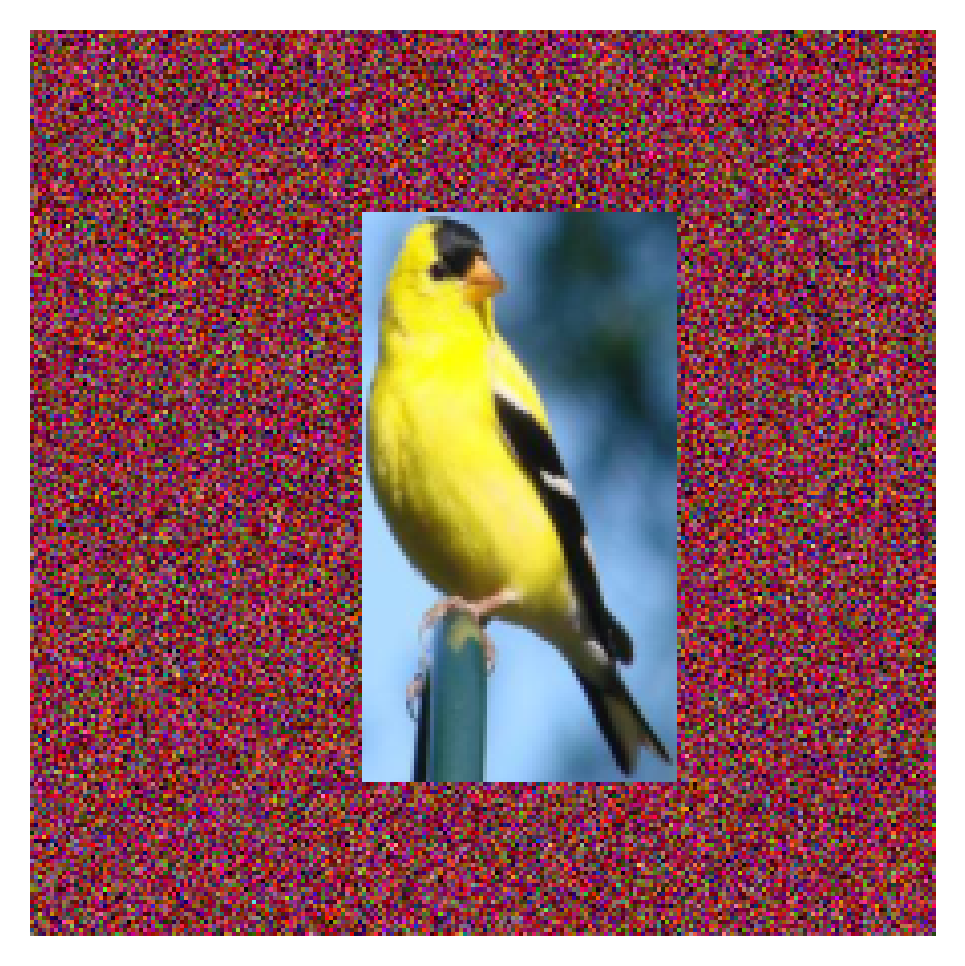} &
  \includegraphics[width=0.1\linewidth]{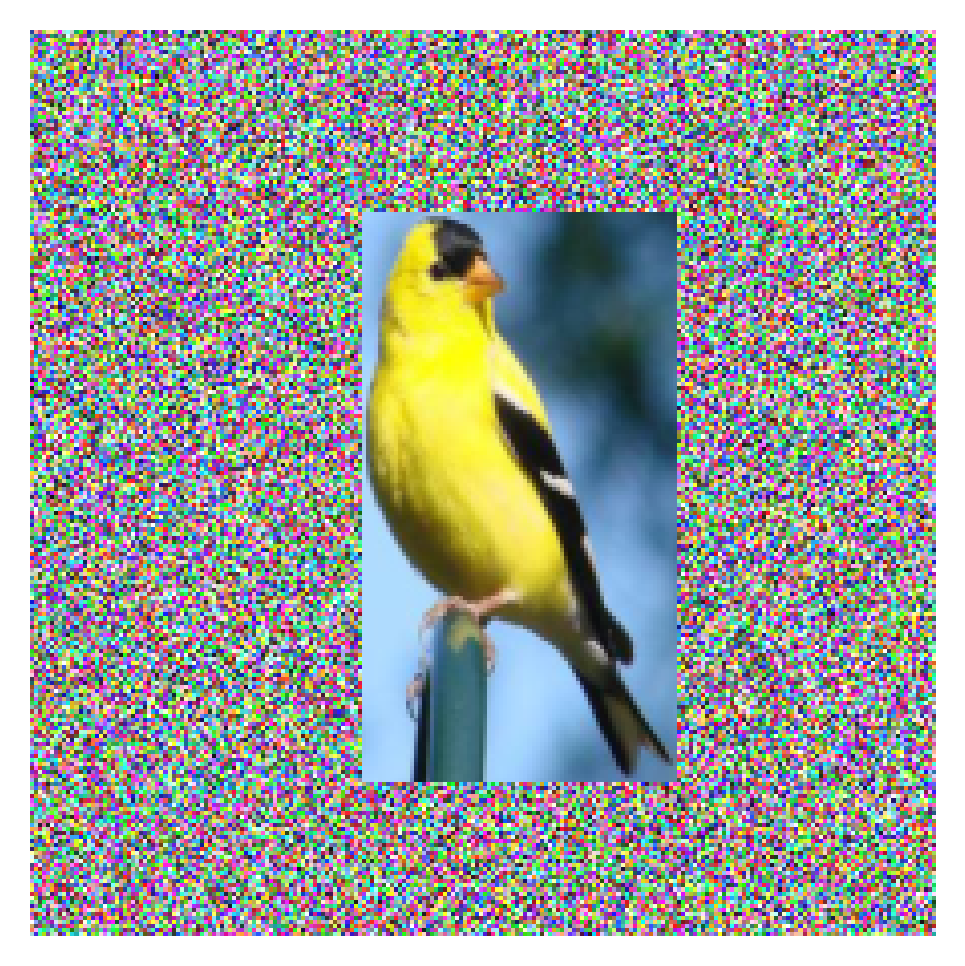} &
  \includegraphics[width=0.1\linewidth]{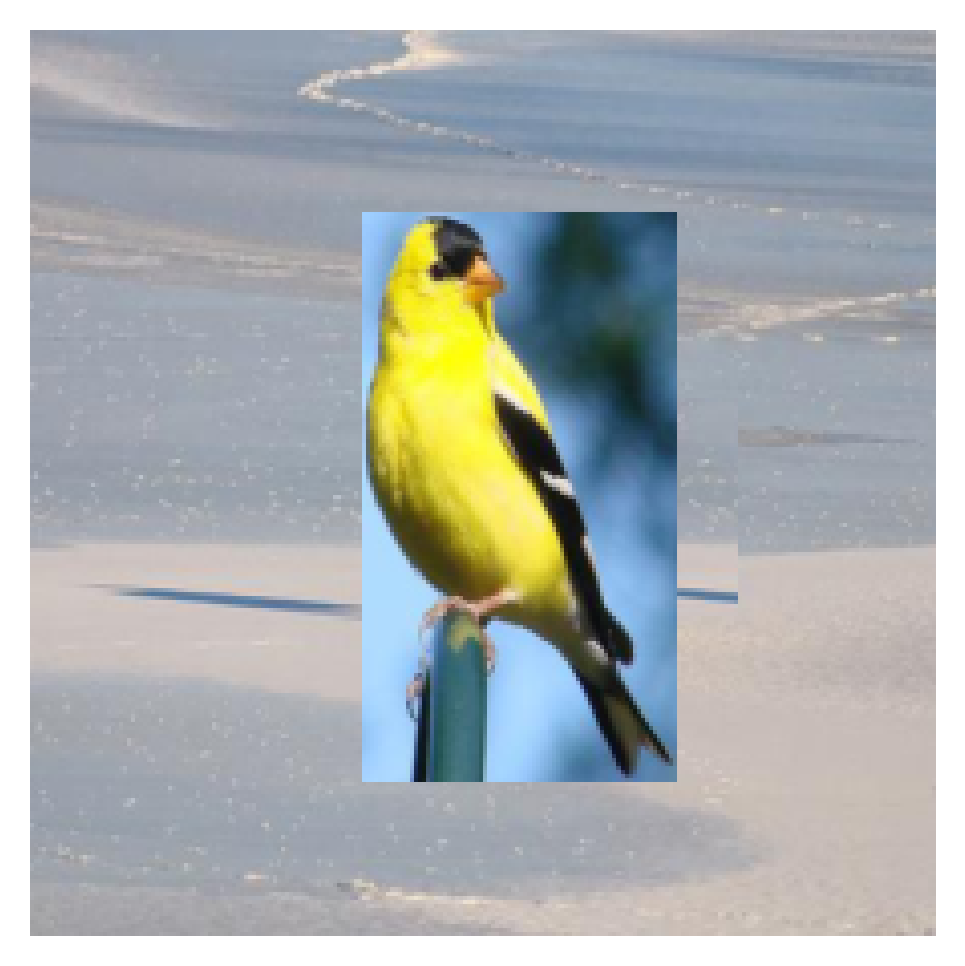} &
  \includegraphics[width=0.1\linewidth]{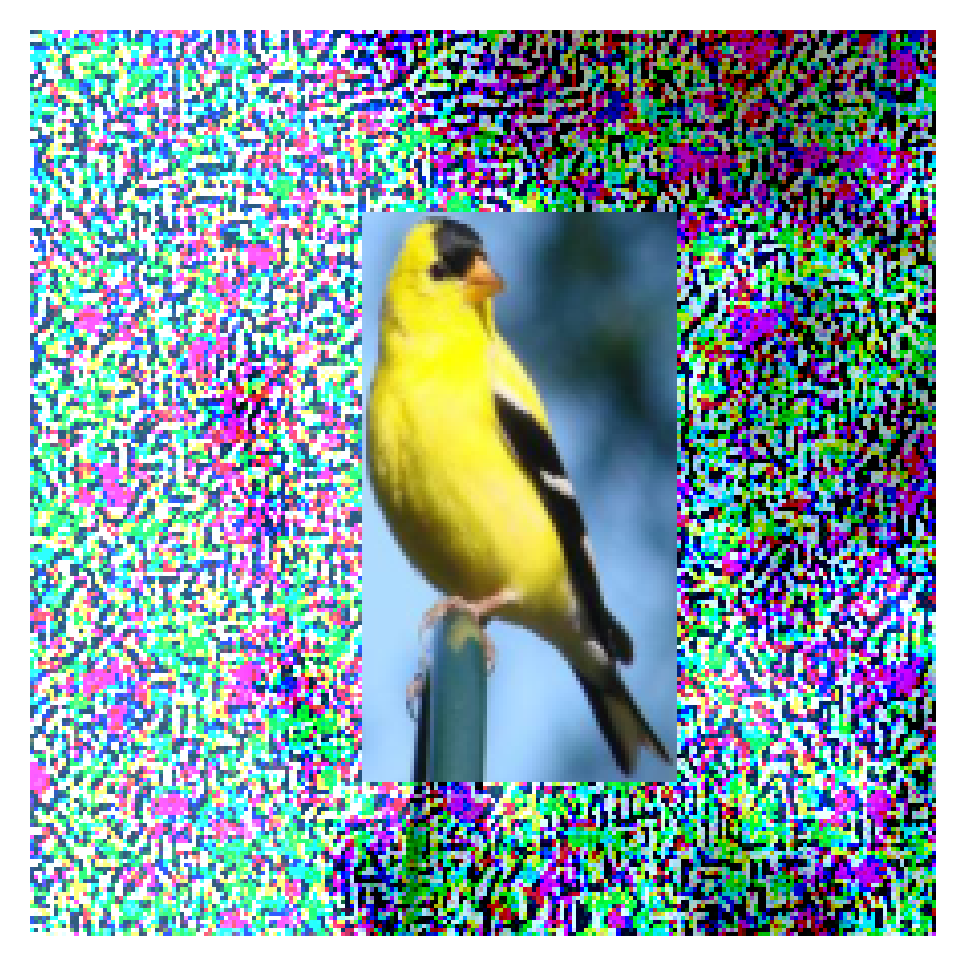} \\
  \end{tabular}
 \caption{
        Examples of Counterfactual (CF) or Factual (F) data generation.
  }

  \label{fig:fig1_inpainting_examples}
\end{figure*}
\setlength\tabcolsep{6pt} 


\section{Proposed Methods}

\paragraph{Generating counterfactual data}

To break the correlation between non-causal features (backgrounds) and labels, we generate counterfactuals that keep the backgrounds in data but remove foregrounds.

Specifically, consider an image $\bm{x}$ with $U$ pixels, a label $y$, and a causal region $\bm{r} \subseteq \{0, 1\}^U$ ($1$ means causal). We define an infilling function $\phi_{cf}$ that mixes the original image and the infilling value (specific choices are presented later):

$$
\phi_{cf}(\bm{x}, \bm{r}) = (1 - \bm{r}) \odot \bm{x} + \bm{r} \odot \bm{\hat{x}} \mathrm{ \ \ , where \ } \bm{\hat{x}} \sim p_{\mathrm{infill}}(\bm{\hat{x}} | \bm{x}_{\bm{r}=\bm{0}}).
$$

Then we label such images as "non-$y$" $\neg y$, which introduces the need for a counterfactual loss function, and we explore $3$ different options: (1) Negative log likelihood of $P(\hat{y} \neq y)$ i.e. $-log (1 - P(\hat{y} = y|x))$;
(2) KL divergence between the uniform distribution and the predicted probability i.e. KL(Uniform$(y) || P(\hat{y} | x))$. 
The intuition is that given we remove foregrounds, the model should just predict a uniform distribution among all classes, since it does not have a class for backgrounds; (3) KL divergence between the uniform distribution except original class $y$, and the predicted probability. 
We found (2) worked poorly for removing the spurious correlations, and that (1) worked better than (3).
We choose (1) as our final objective.
Then we augment our training objective with additional counterfactual loss function:
$$
    L_{cf} = (-log (1 - P_f(\hat{y} = c|\phi_{cf}(x))))
$$

\paragraph{Generating factual data}
To make a classifier immune to background shifts, we augment our data by perturbing the backgrounds which generates new images with unchanged labels.
We define another factual infilling function $\Phi_{f}$ that mixes the foreground with background value $\bm{\hat{x}}$:

$$
    \Phi_{f}(\bm{x}, \bm{r}) = \bm{r} \odot \bm{x} + (1 - \bm{r}) \odot \bm{\hat{x}} \mathrm{ \ \ where \ } \bm{\hat{x}} \sim p_{\mathrm{infill}}(\bm{\hat{x}} | \bm{x}_{\bm{r}=\bm{1}}).
$$

And the final objective (with cross entropy loss) is:
$$
    L = L_{CE} (y, \hat{y}(x)) + L_{cf} +  \overbrace{L_{CE} (y, \hat{y}(\Phi_{f}(\bm{x}, \bm{r}))}^{\mathrm{Factual\ Loss}}
$$

\paragraph{Choice of infilling value}
We describe some methods for producing \emph{counterfactual} infilling values $\hat{x}$ (some choices are inspired by~\citep{chang2018explaining}).
\textbf{Grey} sets each pixel of $\hat{\bm{x}}$ to $0.5$, which is $0$ after being normalized between $-1$ and $1$.
\textbf{Random} first samples $\hat{\bm{x}}$ from a uniform distribution that resembles low-frequency noise, and adds Gaussians of $\sigma=0.2$ per-channel per-pixel as high-frequency noise, and truncates it between $0$ and $1$.
\textbf{Shuffle} randomly shuffles all the pixel values in the specified region; it keeps the marginal distribution but breaks the joint distribution.
\textbf{Tile} first extracts the largest rectangle from the background that does not intersect with the foreground, and uses that to tile the foreground region.
Finally, we use a generative model; \textbf{CAGAN} which is the Contextual Attention GAN  \citep{yu2018cagan}; we use the authors' pre-trained ImageNet model to inpaint the removed foreground.

For \emph{factual} infilling, aside from using \textbf{Random} and \textbf{Shuffle} infilling, we propose \textbf{Mixed-Rand} that swaps the background with another randomly-chosen tiled background from images of other classes within the same training batch. 
We also propose using adversarial attacks to manipulate the non-causal  features, i.e. we perform an adversarial attack only on the background region. 
We adopt the $\ell_{\inf}$ norm and FGSM attack~\citep{goodfellow2014explaining} for its fast computation.
We tried the PGD attack~\citep{madry2018towards} but found that it performs similarly to FGSM yet is more computationally demanding, so we use FGSM for all experiments.
See Figure~\ref{fig:fig1_inpainting_examples} for examples. We use CF abbreviated for Counterfactual and F for Factual.

\paragraph{Baseline}
We compare our work with approaches that penalize the model's saliency (input gradients) outside of bounding boxes.
We find the original form of RRR~\citep{ross2017right} that takes the gradient of sum of the log probabilities across classes do not perform well (also found in~\citet{viviano2019underwhelming}), and thus instead we use GradMask~\citep{simpson2019gradmask} of uncontrast form for multi-class settings that uses the target logit $\hat{\bm{y}}$:
$$
    L_{Sal} = \lambda_{Sal} \sum_{j=1}^U (\frac{\partial \hat{\bm{y}}_{c=y}}{\partial \bm{x_j}})^2 * (1 - \bm{r_j}) / \sum_{j=1}^U (1 - \bm{r_j})
$$


Saliency regularization (\textbf{Sal}) explicitly attempt to break the correlation between non-casual features and labels via model, whereas our augmentations achieve this goal via data.

We also compare with two other methods: \textbf{Mix-up}~\citep{zhang2018mixup} and Label Smoothing (\textbf{LS})~\citep{muller2019does}. 
These techniques were not designed to address spurious correlations, but have nonetheless been shown to improve test set accuracy in image classification tasks by non-trivial margins.

\paragraph{Hyperparameters}
For models, we use a variant of ResNet-50~\citep{kolesnikov2019large} as our architecture for all our datasets.
We call this Original in our experimental section.
For data preprocessing, We scales and center-crops images to 224x224 with horizontal flipping and normalizes to $[-1,1]$.
For Sal, we try $\lambda$ from $\{1$e$-4$, $1$e$-3$, ... to $1000\}$.
For FGSM, we try $\epsilon$ of $\ell_{inf}$ norm for $\{0.15, 0.25, 0.35, 0.5, 0.6, 0.75, 0.9\}$.
For Mix-up, we try $\alpha=\{0.1,0.2,0.3\}$. For LS, we try $\epsilon=\{0.05,0.1,0.2\}$.
We run $3$ different random seeds.
More details are in Appendix A.

\section{Experimental Results}

We aim to answer the following questions for our augmentations: (i) Do they improve the accuracy under shifted distributions? (ii) Do they make model focus more on foregrounds instead of backgrounds measured by saliency map? (iii) Does focusing on foregrounds indicate better accuracy? 
(iv) Do our augmentations make models' predictions less affected by changed backgrounds?
We experiment on two controlled datasets that explicitly swaps the foreground and background, and a real-world dataset that has very different backgrounds between train and test images.



\subsection{IN-9 dataset: a synthetic perturbed background challenge dataset}
\label{sec:in9}

\setlength\tabcolsep{0pt} 
\begin{figure}[tbp]
\centering
\begin{tabular}{cccc}
  (a)Original & (b)Mixed-Same & (c)Mixed-Rand & (d)Mixed-Next    \\
  \includegraphics[width=0.25\linewidth]{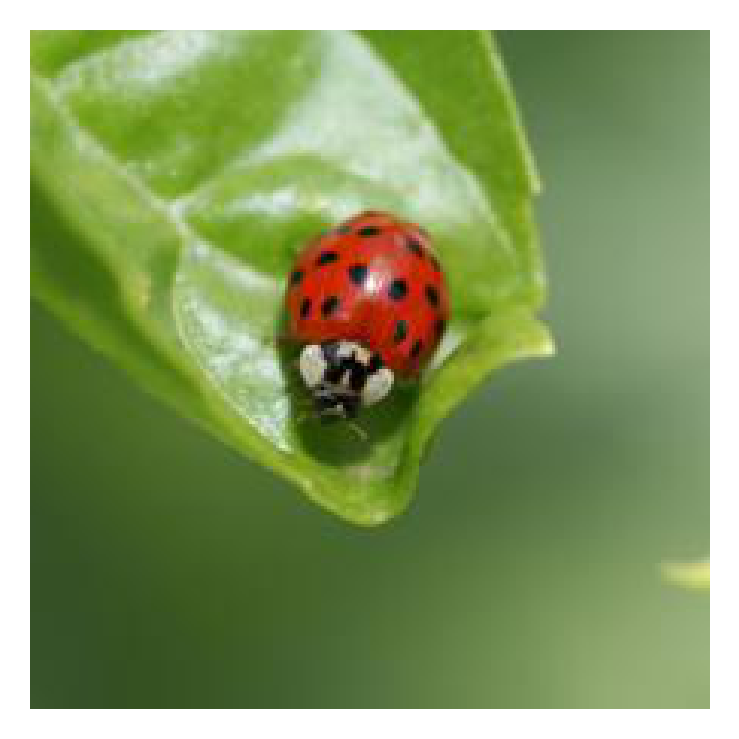} & 
  \includegraphics[width=0.25\linewidth]{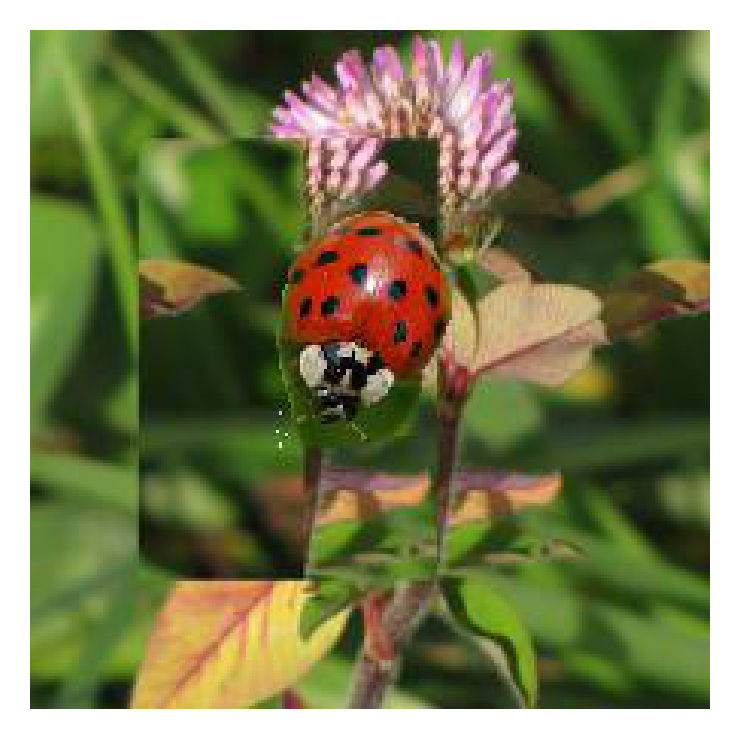} &
  \includegraphics[width=0.25\linewidth]{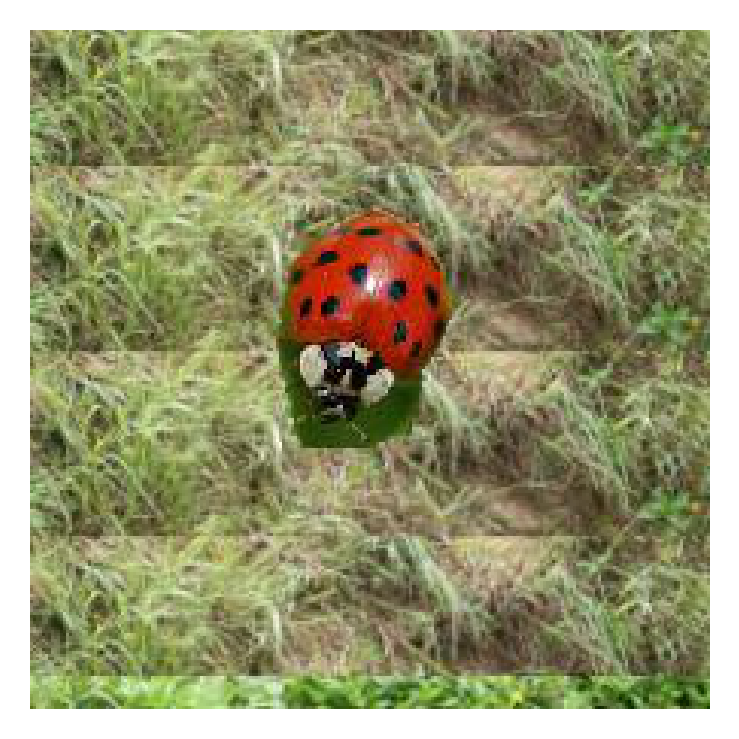} &  
  \includegraphics[width=0.25\linewidth]{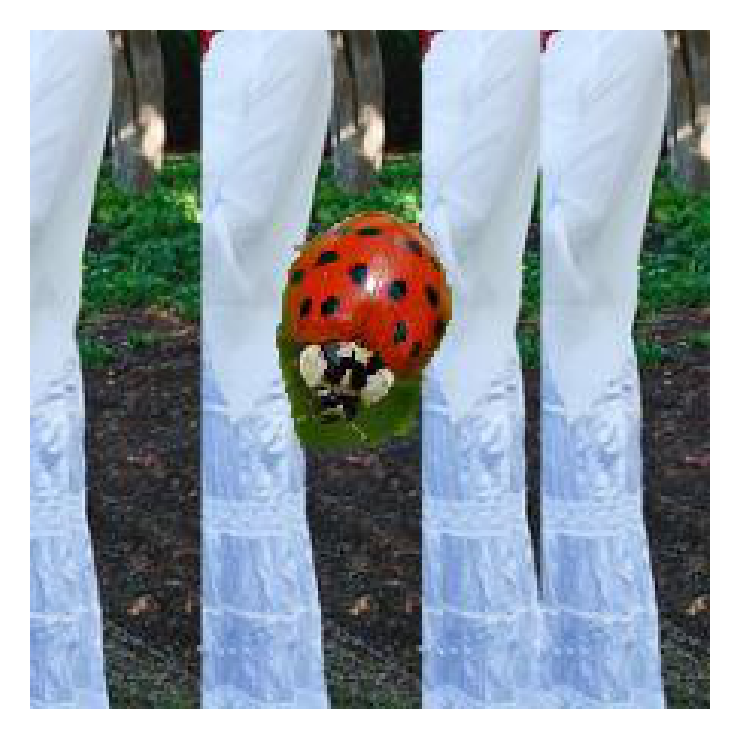} \\
\end{tabular}
\caption{Examples of IN9 test sets. 
Original and Mixed-Same have backgrounds of the same class; Mixed-Rand and Mixed-Next have backgrounds of different classes.}
\label{fig:in9_datasets_examples}
\end{figure}
\setlength\tabcolsep{6pt} 

ImageNet-9 (IN-9) is a dataset proposed by~\citet{xiao2020noise} to disentangle the relationship between foreground and background.
It groups the ImageNet classes into $9$ broad classes and filters the images with bounding box annotations, resulting in each class having $5,045$ training images and $450$ test images.
To disentangle the background and foreground, for each test image they mix the background and foreground in various ways:
(1) \textbf{Mixed-Same}: the background is swapped with the background of another image belonging to the same class. (2) \textbf{Mixed-Rand}: the background is swapped with the background of another image belonging to a different random class. (3) \textbf{Mixed-Next}: the background swapped with the background of another image belonging to the next class, i.e. if the class index for the image is 5, then we swap backgrounds with an image from class 6.
See Figure~\ref{fig:in9_datasets_examples} for an example.
For Sal and F methods, we use the provided foreground segmentation masks as important regions.
For CF methods, since the shape of the mask still leaks the information of the foreground objects, we instead use rectangular bounding box.
We do not compare to F(Mixed-Rand) since our test sets Mixed-Rand and Mixed-Next are constructed the same way and will give F(Mixed-Rand) an unfair advantage.

\setlength\tabcolsep{1.5pt} 
\begin{table}[tbp]
\caption{The performance results ($\%$ of accuracy) of IN9 datasets.
We show mean and standard deviation for $3$ runs.}
\centering
\small
\begin{tabular}{l|cccc}
 & Original           & \makecell{Mixed\\Same}      & \makecell{Mixed\\Rand}    & \makecell{Mixed\\Next} \\ \toprule
Orig.              & $82.0\ (0.7)$ & $71.0\ (0.4)$ & $51.5\ (0.4)$ & $47.5\ (0.6)$ \\ \midrule
+ Sal($\lambda=10^{-4}$)                      & $82.5\ (0.6)$ & $71.4\ (1.6)$ & $52.4\ (2.4)$      & $48.5\ (1.4)$      \\ \midrule
+ CF(Grey)         & $82.4\ (0.3)$ & $68.8\ (0.9)$ & $51.4\ (0.3)$ & $48.8\ (0.3)$ \\
+ CF(Random)       & $82.5\ (0.2)$ & $69.8\ (0.7)$ & $52.1\ (0.3)$ & $49.3\ (1.1)$ \\
+ CF(Shuffle)      & $82.6\ (0.6)$ & $71.5\ (0.3)$ & $51.8\ (0.3)$ & $48.4\ (0.5)$ \\
+ CF(Tile)         & $81.9\ (0.8)$ & $58.8\ (1.3)$ & $46.5\ (0.5)$ & $45.9\ (0.1)$ \\
+ CF(CAGAN)        & $80.5\ (0.6)$ & $68.7\ (0.2)$ & $53.3\ (0.2)$ & $50.0\ (0.4)$ \\ \midrule
+ F(Random)        & $83.2\ (0.9)$ & $73.9\ (1.7)$ & $56.4\ (2.8)$ & $53.0\ (2.5)$ \\
+ F(Shuffle)       & $\bm{83.5}\ (0.0)$ & $73.8\ (0.3)$ & $57.1\ (0.8)$ & $53.3\ (0.7)$ \\
+ F(FGSM $\epsilon=0.5$) & $82.6\ (1.3)$ & 	$70.6\ (1.5)$ & 	$51.6\ (2.2)$ & $47.9\ (2.5)$ \\ \midrule
\makecell[l]{+ CF(CAGAN)\\+ F(Shuffle)}                    & $81.2\ (0.7)$ & $71.5\ (1.3)$ & $\bm{58.2}\ (1.7)$ & $54.2\ (1.0)$      \\ \midrule
\makecell[l]{+ CF(CAGAN)+Sal}  & $80.5\ (0.2)$ & $68.0\ (0.9)$ & $53.2\ (0.8)$ & $49.9\ (0.9)$ \\ \midrule
\makecell[l]{+ F(Shuffle) + Sal} & $83.0\ (0.7)$ & $\bm{74.5}\ (0.1)$ & $57.1\ (0.9)$ & $52.8\ (1.6)$ \\ \midrule
\makecell[l]{+ CF(CAGAN)\\+ F(Shuffle) + Sal} & $81.9\ (1.2)$ & $72.5\ (0.5)$ & $\bm{58.0}\ (1.2)$      & $\bm{55.5}\ (2.1)$ \\ \midrule
+ Mixup($\alpha=0.2$)  & $82.8\ (1.7)$ & $72.1\ (1.0)$ & $52.2\ (1.7)$ & $48.2\ (1.8)$ \\
+ LS($\epsilon=0.1$)  & $82.9\ (0.4)$ & $71.8\ (0.7)$ & $52.5\ (0.5)$ & $48.5\ (0.7)$ \\
\end{tabular}
\label{table:in9}
\end{table}

We compare our methods in Table~\ref{table:in9}.
For models that rely less on backgrounds, we expect worse performance in Original and Mixed-Same where leveraging backgrounds is beneficial at test time, but expect improvement in Mixed-Rand or Mixed-Next where backgrounds contradict labels.
Indeed, Sal and CF methods (except CF(Tile)) perform as expected, with CF(CAGAN) as the best method in Mixed-Rand and Mixed-Next while doing slightly worse in Original and Mixed-Same.
To our surprise, F methods like F(Shuffle) perform better in all $4$ test sets.
We think this is because F methods increase the sample size of images with foregrounds which helps learn more generalizable features. 
We further combine the best methods in CF, F and Sal and show their combinations improve accuracy even more, suggesting their gains have different causes.

\begin{table}[tbp]
\caption{The model's AUPR of using saliency map to predict binary foregrounds, and its corresponding accuracy in IN-9 Original and Mixed-Next. High Saliency AUPR does not necessarily indicate high accuracy in both datasets.}
\centering
\begin{tabular}{l|cc|cc}
                        & \multicolumn{2}{c}{Saliency AUPR} & \multicolumn{2}{c}{Accuracy} \\
                        \cmidrule(lr){2-3}\cmidrule(lr){4-5}
         & Original              & Mixed-next    & Original      & Mixed-next    \\ \toprule
Orig.     & $56.7\ (0.8)$         & $58.2\ (0.5)$ & $82.0\ (0.7)$ & $47.5\ (0.6)$ \\
+ Sal($\lambda=10^{-4}$)      & $56.0\ (1.6)$         & $58.8\ (0.5)$ & $82.5\ (0.6)$ & $48.5\ (1.4)$ \\
+ CF(CAGAN) & $57.5\ (0.9)$         & $58.7\ (0.4)$ & $80.5\ (0.6)$ & $50.0\ (0.4)$ \\
+ F(shuffle) & $59.8\ (0.9)$ & $\bm{61.6}\ (0.5)$ & $\bm{83.5}\ (0.0)$ & $53.3\ (0.7)$ \\
+ Combined & $\bm{60.8}\ (1.0)$         & $61.2\ (0.9)$ & $81.9\ (1.2)$ & $\bm{55.5}\ (2.1)$ \\ \midrule
+ Sal($\lambda=100$) & $60.5\ (0.7)$         & $61.2\ (0.4)$ & $80.8\ (0.6)$ & $45.9\ (1.2)$            \\
\end{tabular}
\label{table:in9_sal_map}
\end{table}

\setlength\tabcolsep{0pt} 
\begin{figure}[tbp]
\centering
\begin{tabular}{ccc}
  & Original & Mixed-Next   \\
  \raisebox{5\normalbaselineskip}[0pt][0pt]{\rotatebox[origin=c]{90}{Accuracy}} & \includegraphics[width=0.5\linewidth]{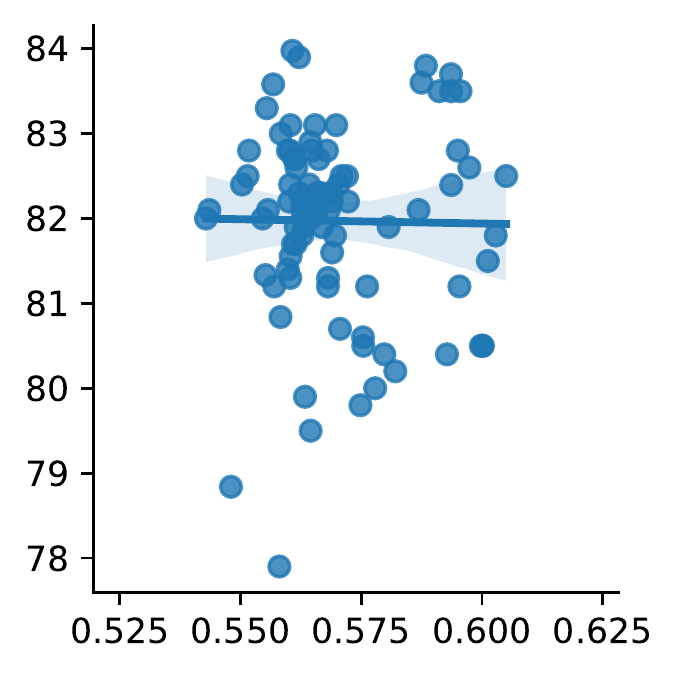} & 
  \includegraphics[width=0.5\linewidth]{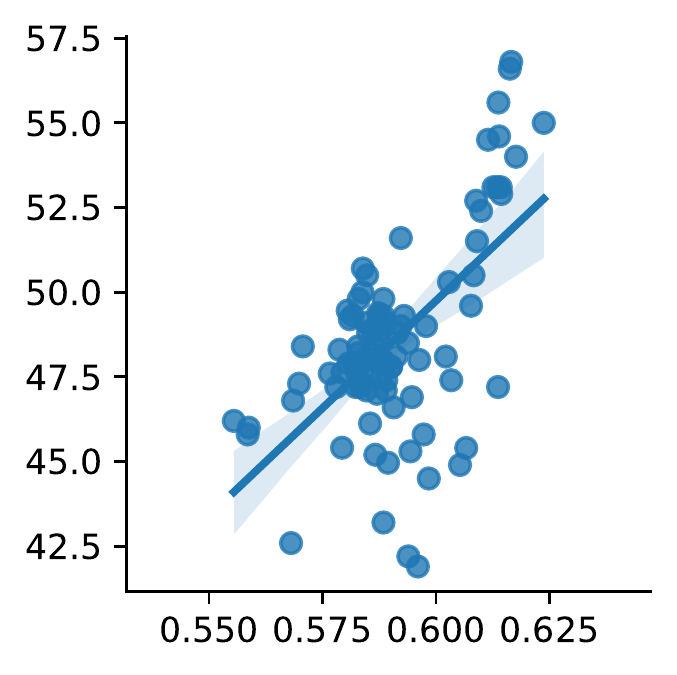}  \\ \vspace{-5pt}
  & \ \ \ \ Saliency AUPR & \ \ \ \ Saliency AUPR
\end{tabular}
\caption{The scatter plot between Saliency map (DeepLiftShap) AUPR and accuracy for all models we trained in IN-9. Only in Mixed-Next there is a positive correlation ($R^2=0.4$) between saliency AUPR and accuracy.}
\label{fig:in9_sal_acc_scatter_plot}
\end{figure}
\setlength\tabcolsep{6pt} 

To determine if our augmentations cause models to focus more on foregrounds thereby resulting in higher accuracy, we measure the saliency map of each model and quantify its overlap with provided foreground regions.
We try both DeepLiftShap~\citep{lundberg2017unified} and input gradient as choices of saliency maps and find their ranking is similar, and thus only show results of DeepLiftShap.
Specifically, given an image, we take the $\ell_2$ norm of the saliency map across channels per pixel as prediction score, and set targets as $1$ for pixels in foregrounds and $0$ in backgrounds, and compute the binary Area Under the Precision-Recall curve (AUPR).
We show saliency AUPR and accuracy in Table~\ref{table:in9_sal_map}.
Overall F(Shuffle) increases the foreground focus the most compared to Sal and CF(CAGAN), suggesting its large accuracy improvement comes from better focus on the foreground.
But, CF(CAGAN) in Mixed-Next with almost the same foreground AUPR still achieves higher accuracy.
On the other hand, Sal with high penalty ($\lambda=100$) which explicitly encourages higher foreground focus has similar AUPR as F(Shuffle) while having much lower accuracy.
To further understand the relationship, in Figure~\ref{fig:in9_sal_acc_scatter_plot} we plot the Saliency AUPR v.s. Accuracy for all models we have trained.
We find no positive correlation in the Original test set which makes sense because getting high accuracy on this partition does not require focusing on foregrounds.
In Mixed-Next we do find a stronger positive correlation ($R^2=0.4$), although we still have a few outliers in the lower right corner;
they are Sal with high $\lambda$ or F(FGSM) with high $\epsilon$.
These results show that accuracy only correlates with foreground AUPR when backgrounds disagree with labels (e.g. Mixed-Next), but does not necessarily correlate well. 
In fact there may be a tradeoff when strong regularization is used.

\setlength\tabcolsep{3pt} 
\begin{table}[tbp]
\caption{The difference of the probability ($\%$) of the next class between Mixed-Next and Original IN9 test sets. The lower the number the model is less affected by backgrounds.}
\centering

\begin{tabular}{ccccc}
Orig          & Sal          & CF(CAGAN)     & F(shuffle)    & Combined     \\ \toprule
$10.1\ (0.3)$ & $9.9\ (0.4)$ & $7.4\ (0.2)$ & $9.2\ (0.1)$ & $\bm{6.8}\ (0.3)$
\end{tabular}
\label{table:in9_mixed_next_prob_increase}
\end{table}


To investigate if our models indeed learn to ignore spurious correlations (backgrounds), we measure the difference of probability for the next class between Mixed-Next and Original test sets.
Given that these two test sets have identical foregrounds, a model relying more on backgrounds to predict will increase the probability more for the next class when backgrounds are swapped with next-class backgrounds like Mixed-Next does.
We show results in Table~\ref{table:in9_mixed_next_prob_increase}. 
F(Shuffle), although being the most accurate model, is not the least reliant on backgrounds. 
Instead CF(CAGAN) is the best from this perspective.
This confirms our intuition that F(Shuffle) improves accuracy not only by ignoring backgrounds but also generalizing better in foregrounds (and thus have stronger foreground focus).
Both CF and F augmentations outperform Sal, and the combination of the three further decreases the reliance on backgrounds.

\setlength\tabcolsep{0pt} 
\begin{figure}[tbp]
\centering
\begin{tabular}{ccc}
   & Original & Mixed-Next  \\
\raisebox{3.5\normalbaselineskip}[0pt][0pt]{\rotatebox[origin=c]{90}{\small{Accuracy}}}
   & \includegraphics[width=0.5\linewidth]{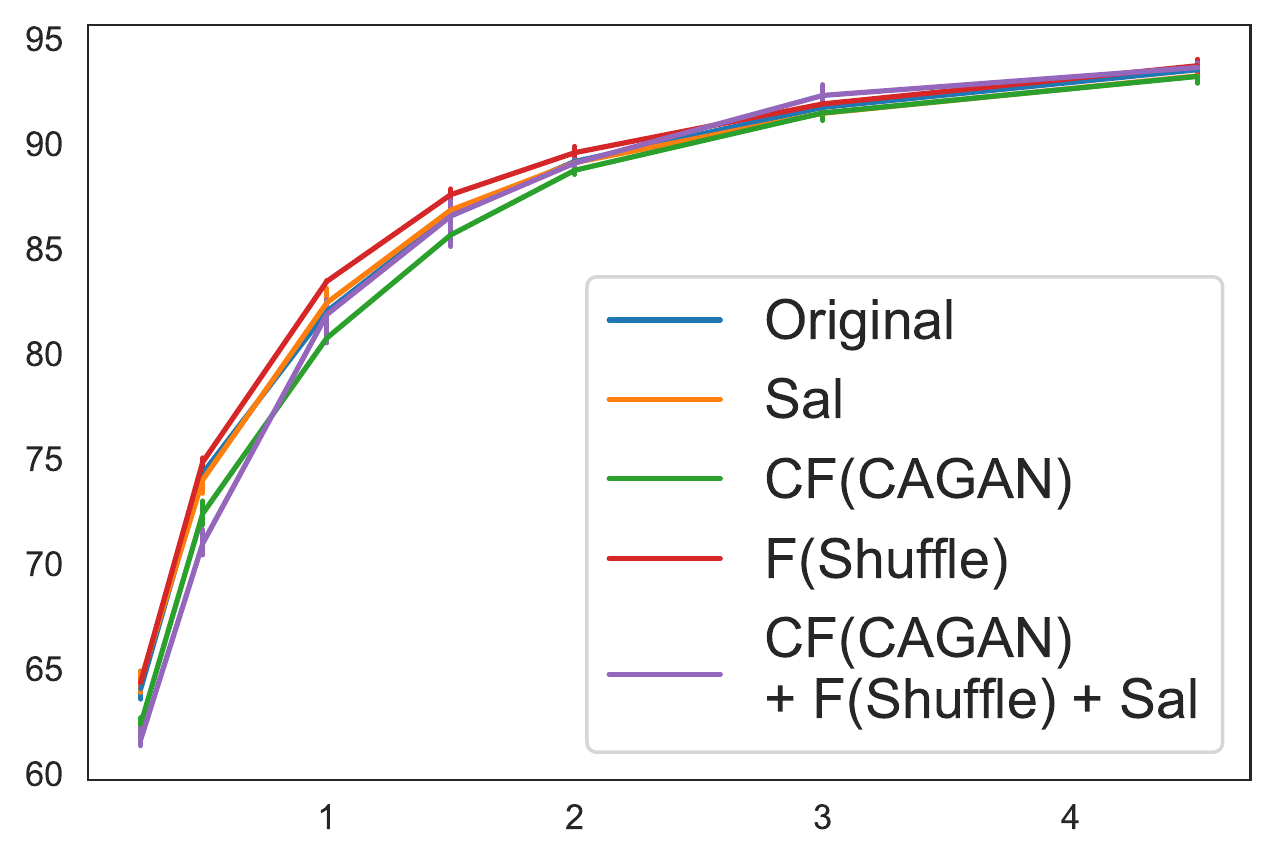}
   & \includegraphics[width=0.5\linewidth]{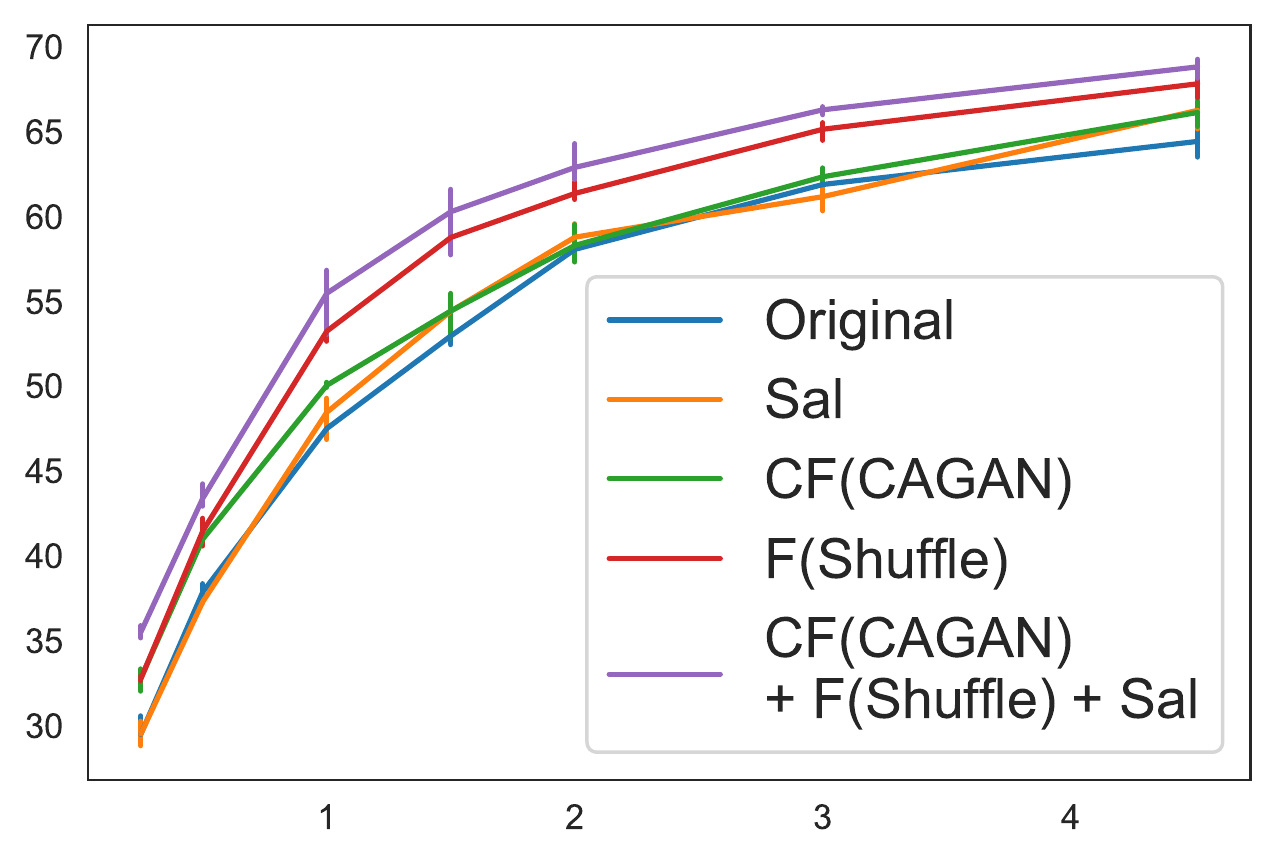} \\
   & Training Data Ratio & Training Data Ratio  \\

\end{tabular}
\caption{Our methods improve the accuracy under various training data size. Note that we only augment maximum data ratio $1$ data since only these images have bounding box annotations, and thus gaps decrease after $1$.}
\label{fig:in9_data_ratio}
\end{figure}
\setlength\tabcolsep{6pt} 

To understand how training set size affects our improvements, in Figure~\ref{fig:in9_data_ratio} we train models using various training data sizes.
Here data ratio as $1$ means using the original IN-9 dataset around $45$k images, and we subset images to make ratio less than $1$, or include remaining ImageNet images without bounding boxes to make it bigger than $1$.
And in Figure~\ref{fig:in9_data_ratio} we measure the performance gain across different data ratio.
When the data ratio is $<= 1$,  our methods continue improving as more data becomes available.
When the data ratio is $> 1$, the performance gap narrows since the size for our additional data augmentation stays fixed as $1$.
In summary, our methods improve over baselines across different training sizes, and the more data the better.

\setlength\tabcolsep{0pt} 
\begin{figure*}[tbp]

\centering
\begin{tabular}{cccccc}
  \includegraphics[width=0.166\linewidth]{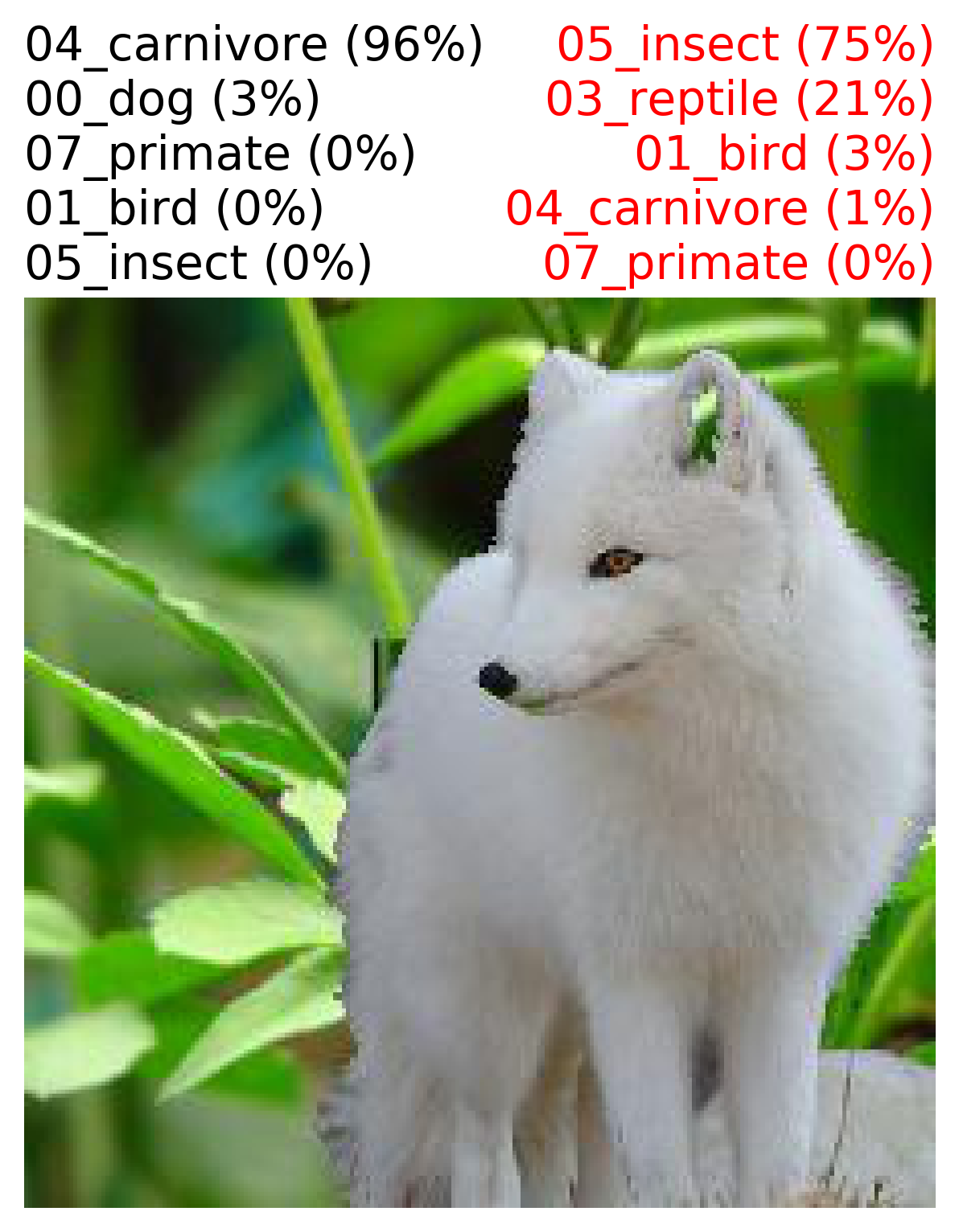} & 
  \includegraphics[width=0.166\linewidth]{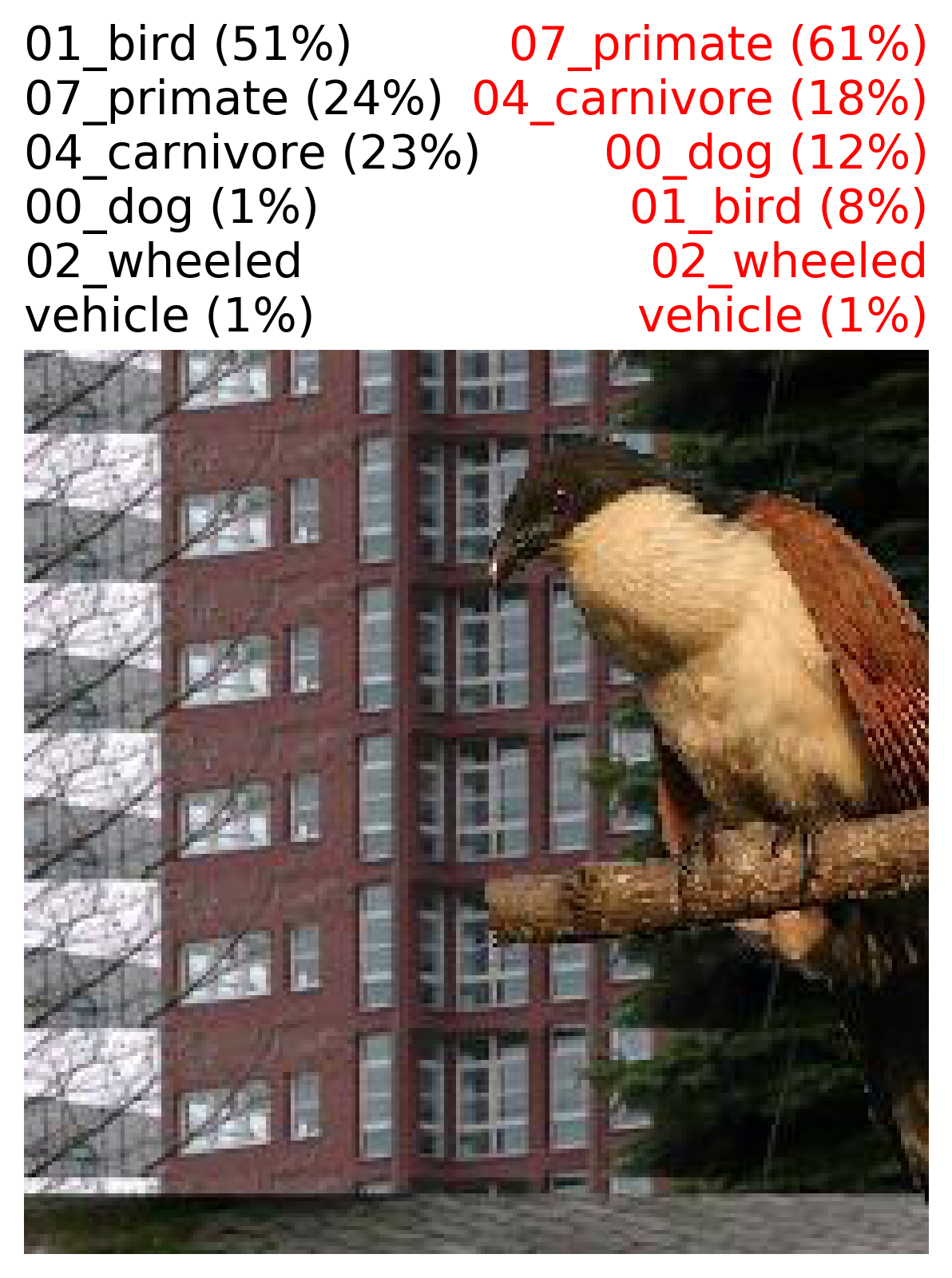} & 
  \includegraphics[width=0.166\linewidth]{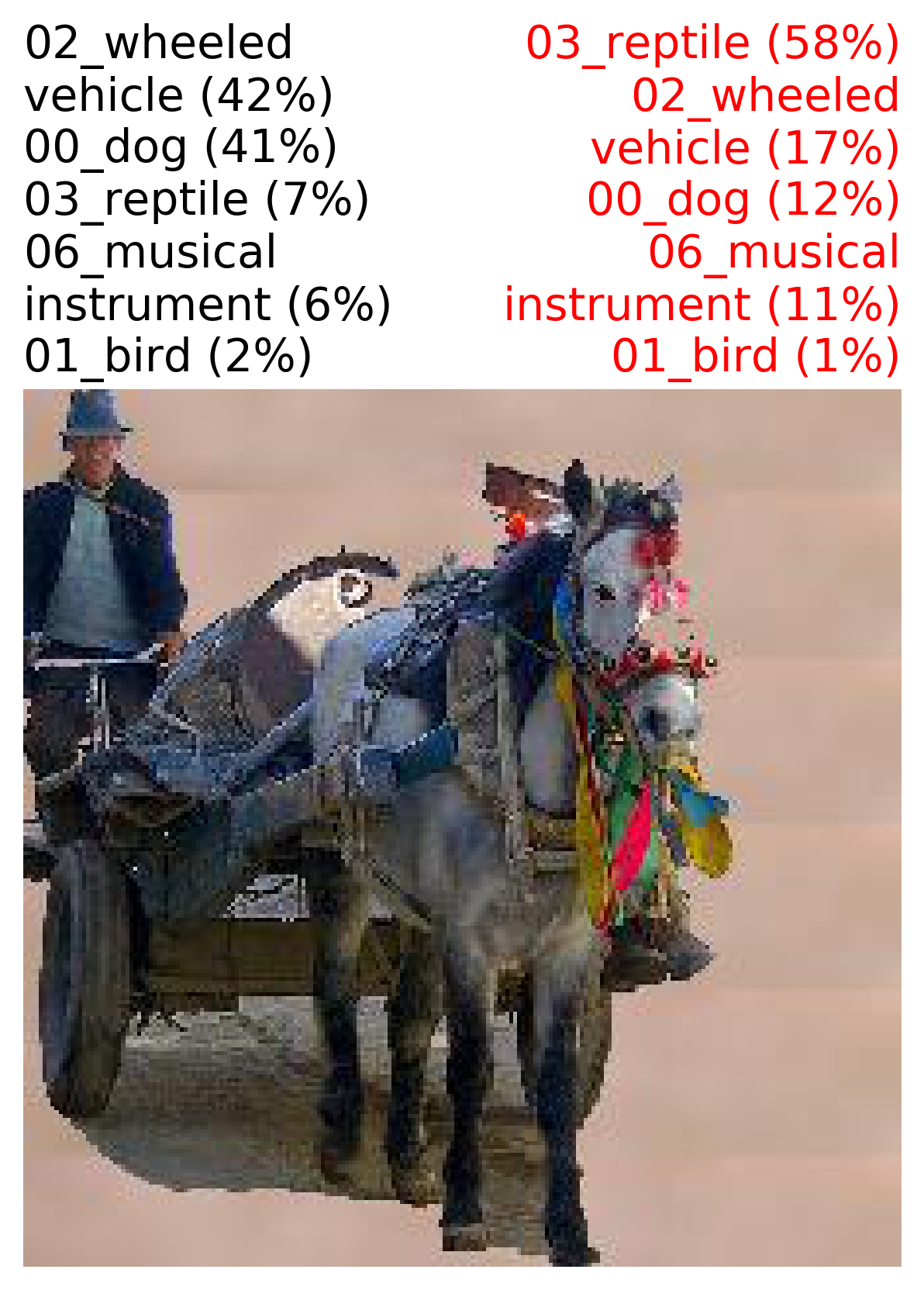} & 
  \includegraphics[width=0.166\linewidth]{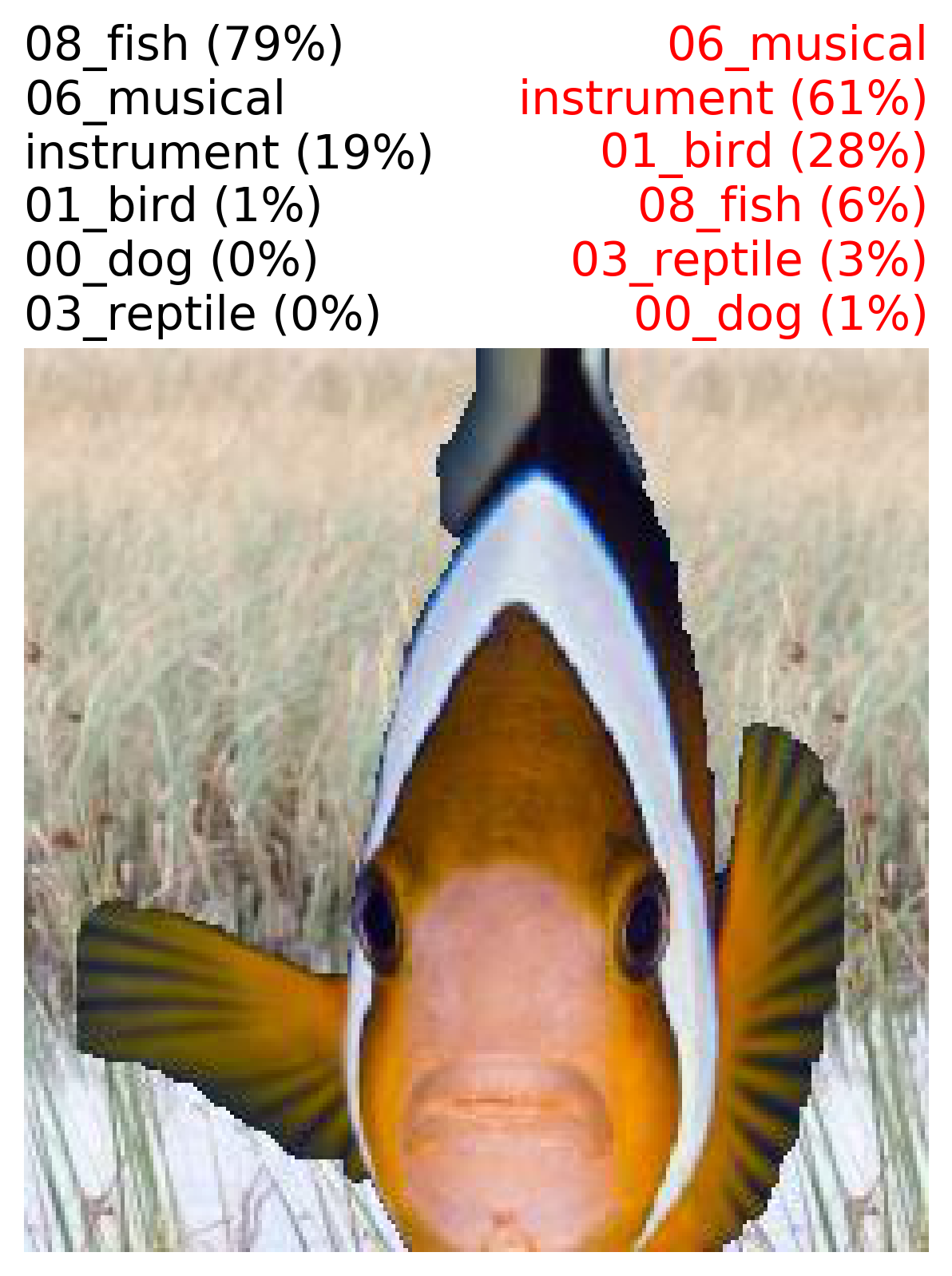} &
  \includegraphics[width=0.166\linewidth]{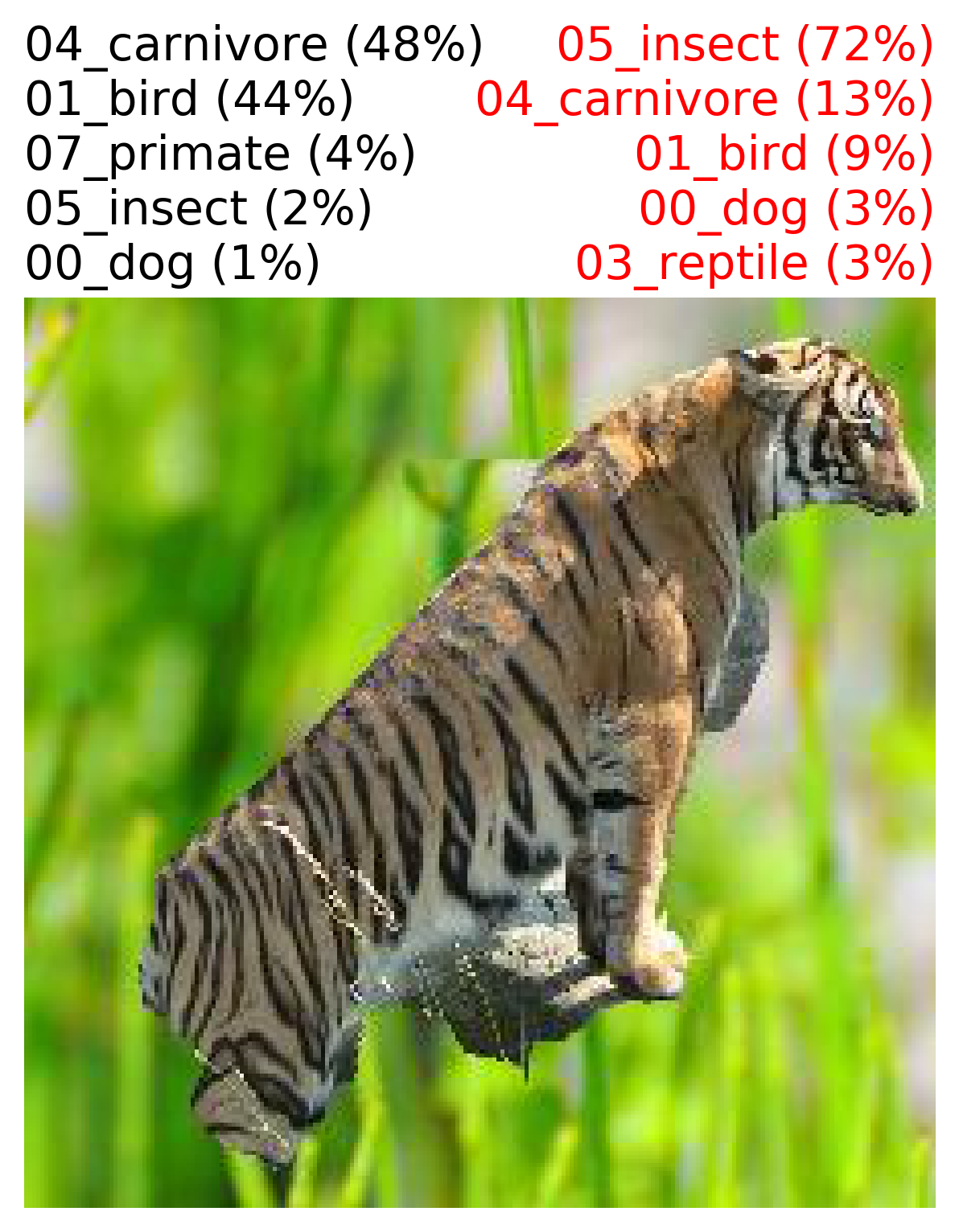} & 
  \includegraphics[width=0.166\linewidth]{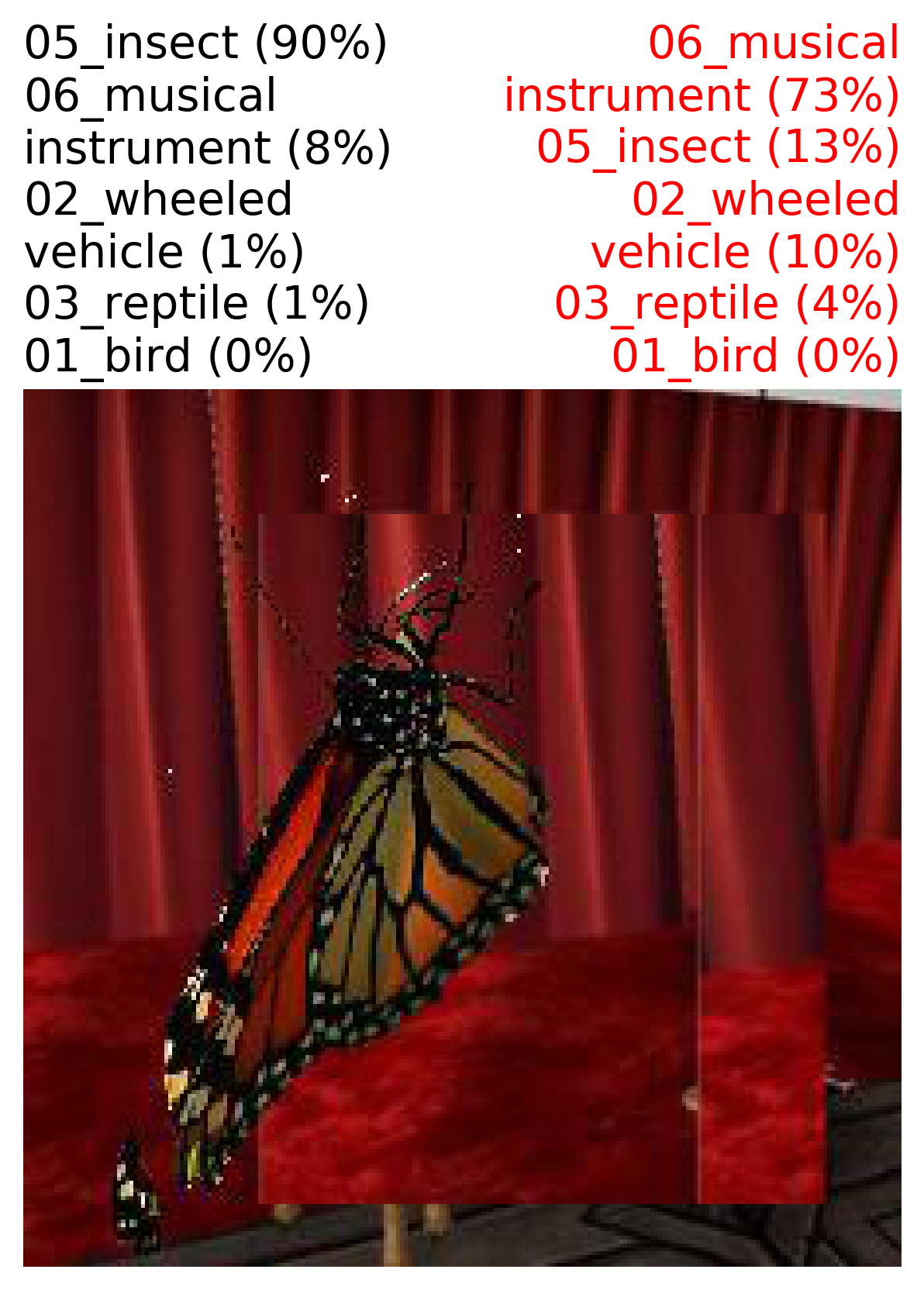}
\end{tabular}
\caption{Examples in the IN9 Mixed-Next dataset that the best model (CF(CAGAN)+F(Shuffle)+Sal) predicts correctly while Original model fails.
Grey texts are the predictions from the best model, and Red text are from Original model.}
\label{fig:in9_winning_cases}
\end{figure*}
\setlength\tabcolsep{6pt} 

In Figure~\ref{fig:in9_winning_cases} we show examples in the Mixed-Next dataset which the best model (CF(CAGAN)+F(Shuffle)+Sal) predicts correctly and the Original model fails on.
We find that their top-5 predictions can be very different.

\subsection{Waterbirds dataset}

Waterbirds is proposed by~\citet{Sagawa2020Distributionally} to study the effect of spurious correlations.
It combines bird photographs from the Caltech-UCSD Birds-200-2011 (CUB) dataset~\citep{WahCUB_200_2011} with image backgrounds from the Places dataset~\citep{zhou2014learning}. 
They label each bird as one of Y = \{waterbird, landbird\} and place it on one of A = \{water background, land background\}
, with waterbirds (landbirds) more frequently appearing against a water (land) background.
In the training set, they place 95\% of waterbirds against a water background and the remaining 5\% against a land background.
Similarly, 95\% of all landbirds are
placed against a land background with the remaining 5\% against water.
The number is balanced for the validation set.
For test set, we divide test images into Original and Flip, where Original contains images with waterbirds on water background and landbirds on land background, while Flip has opposite bird/backgrounds mix.
There are $4,795$, $1199$, $2897$ and $2897$ examples in the training, validation, Original Test, and Flip Test set.
We follow the original paper to finetune a pretrained ResNet-50 on this dataset.

We show the accuracy of each method in Table~\ref{table:waterbird}.
Similar to what we found in IN-9, Sal maintains the accuracy in Original Test set and improves slightly in Flip Test. 
CF methods decrease accuracy slightly in the Original Test set while improving accuracy in the Flip Test set, with our most natural augmentations CF(CAGAN) improving the most.
F methods remain similarly accurate in Original Test while improving Flip accuracy quite a bit with F(Random) as the best.
Further combining the best methods - CF(CAGAN), F(Random) and Sal - improves our Flip accuracy even more up to $19\%$ relative improvement compared to Original method.

\setlength\tabcolsep{3pt} 
\begin{table}[tbp]
\caption{The accuracy on the Waterbird. The Improv(\%). is the relative improvement of Flip Test w.r.t. the Orig. method. We show mean and standard deviation for $3$ runs.}
\centering

\begin{tabular}{l|ccc}
 & Orig. Test     & Flip Test      & Improv(\%) \\  \toprule
Orig.                    & $\bm{98.4}\ (0.1)$ & $76.1\ (0.4)$ & $0.0$         \\  \midrule
+ Sal($\lambda=10$)    & $98.4\ (0.1)$ & $77.6\ (0.9)$ & $1.9$         \\  \midrule
+ CF(Grey)               & $96.0\ (1.8)$ & $81.9\ (4.9)$ & $7.6$         \\
+ CF(Random)             & $98.1\ (0.1)$ & $73.2\ (1.2)$ & $-3.8$        \\
+ CF(Shuffle)            & $98.1\ (0.1)$ & $71.7\ (0.6)$ & $-5.8$        \\
+ CF(Tile)               & $96.1\ (1.1)$ & $81.6\ (3.6)$ & $7.2$         \\
+ CF(CAGAN)              & $95.2\ (0.1)$ & $85.6\ (0.3)$ & $12.5$        \\  \midrule
+ F(Random)              & $98.4\ (0.2)$ & $80.7\ (0.2)$ & $6.0$         \\
+ F(Shuffle)             & $98.4\ (0.0)$ & $79.6\ (0.5)$ & $4.5$         \\
+ F(FGSM)                & $98.2\ (0.1)$ & $69.3\ (1.5)$ & $-8.9$        \\  \midrule
\makecell[l]{+ CF(CAGAN)\\+ F(Random)} & $94.8\ (0.4)$ & $89.6\ (0.7)$ & $17.7$        \\  \midrule
\makecell[l]{+ CF(CAGAN)\\+ F(Random)+Sal} & $94.8\ (0.1)$ & $\bm{90.6}\ (0.0)$ & $\bm{19.0}$        \\  \midrule
+ Mixup($\alpha=0.1$)    & $98.4\ (0.0)$ & $76.7\ (1.5)$ & $0.8$         \\
+ LS($\epsilon=0.1$)     & $98.4\ (0.0)$ & $78.7\ (0.5)$ & $3.4$        
\end{tabular}
\label{table:waterbird}
\end{table}

\setlength\tabcolsep{1pt} 
\begin{table}[tbp]
\caption{The AUPR of the saliency map and the accuracy on Waterbirds Original and Flip test sets.}
\centering
\begin{tabular}{l|cc|cc}
                        & \multicolumn{2}{c}{Saliency AUPR} & \multicolumn{2}{c}{Accuracy} \\
                        \cmidrule(lr){2-3}\cmidrule(lr){4-5}
         & Original              & Flip    & Original      & Flip    \\ \toprule
Orig.                 & $56.2\ (0.4)$ & $48.8\ (1.0)$ & $\bm{98.4}\ (0.1)$ & $76.1\ (0.4)$ \\
+ Sal($\lambda=10$)   & $60.3\ (0.7)$ & $55.0\ (0.7)$ & $98.4\ (0.1)$ & $77.6\ (0.9)$ \\
+ CF(CAGAN)           & $56.7\ (0.8)$ & $48.4\ (1.0)$ & $95.2\ (0.1)$ & $85.6\ (0.3)$ \\
+ F(Random)           & $58.3\ (0.3)$ & $50.6\ (0.1)$ & $98.4\ (0.2)$ & $80.7\ (0.2)$ \\
+ Combined & $58.2\ (0.6)$ & $50.2\ (1.3)$ & $94.8\ (0.1)$ & $\bm{90.6}\ (0.0)$ \\ \midrule
+ Sal($\lambda=10^3$) & $\bm{62.0}\ (0.4)$ & $\bm{57.7}\ (0.9)$ & $97.6\ (0.1)$ & $72.4\ (1.1)$ \\
\end{tabular}
\label{table:wb_sal_map}
\vspace{-10pt}
\end{table}

\setlength\tabcolsep{0pt} 
\begin{figure}[tbp]
\centering
\begin{tabular}{ccc}
  & Original & Flip   \\
  \raisebox{5\normalbaselineskip}[0pt][0pt]{\rotatebox[origin=c]{90}{Accuracy(\%)}} & \includegraphics[width=0.5\linewidth]{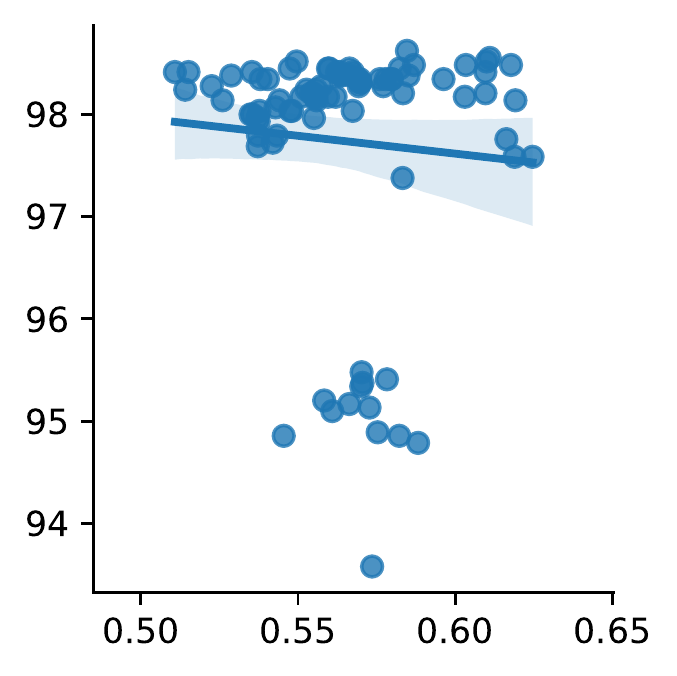} & 
  \includegraphics[width=0.5\linewidth]{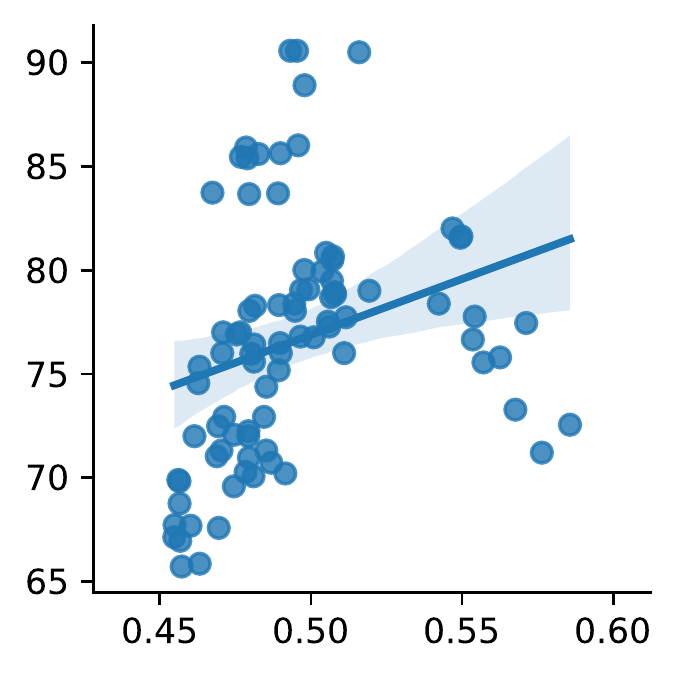}  \\ \vspace{-5pt}
  & \ \ \ \ Saliency AUPR & \ \ \ \ Saliency AUPR
\end{tabular}
\caption{The scatter plot between Saliency map (DeepLiftShap) AUPR and accuracy for all models we trained in Waterbirds. In Flip test set there is a stronger positive correlation ($R^2=0.08$) between saliency AUPR and accuracy.}
\label{fig:wb_sal_acc_scatter_plot}
\end{figure}
\setlength\tabcolsep{3pt} 

\setlength\tabcolsep{0pt} 
\begin{figure*}[tbp]

\centering
\begin{tabular}{cccccc}
  \includegraphics[width=0.166\linewidth]{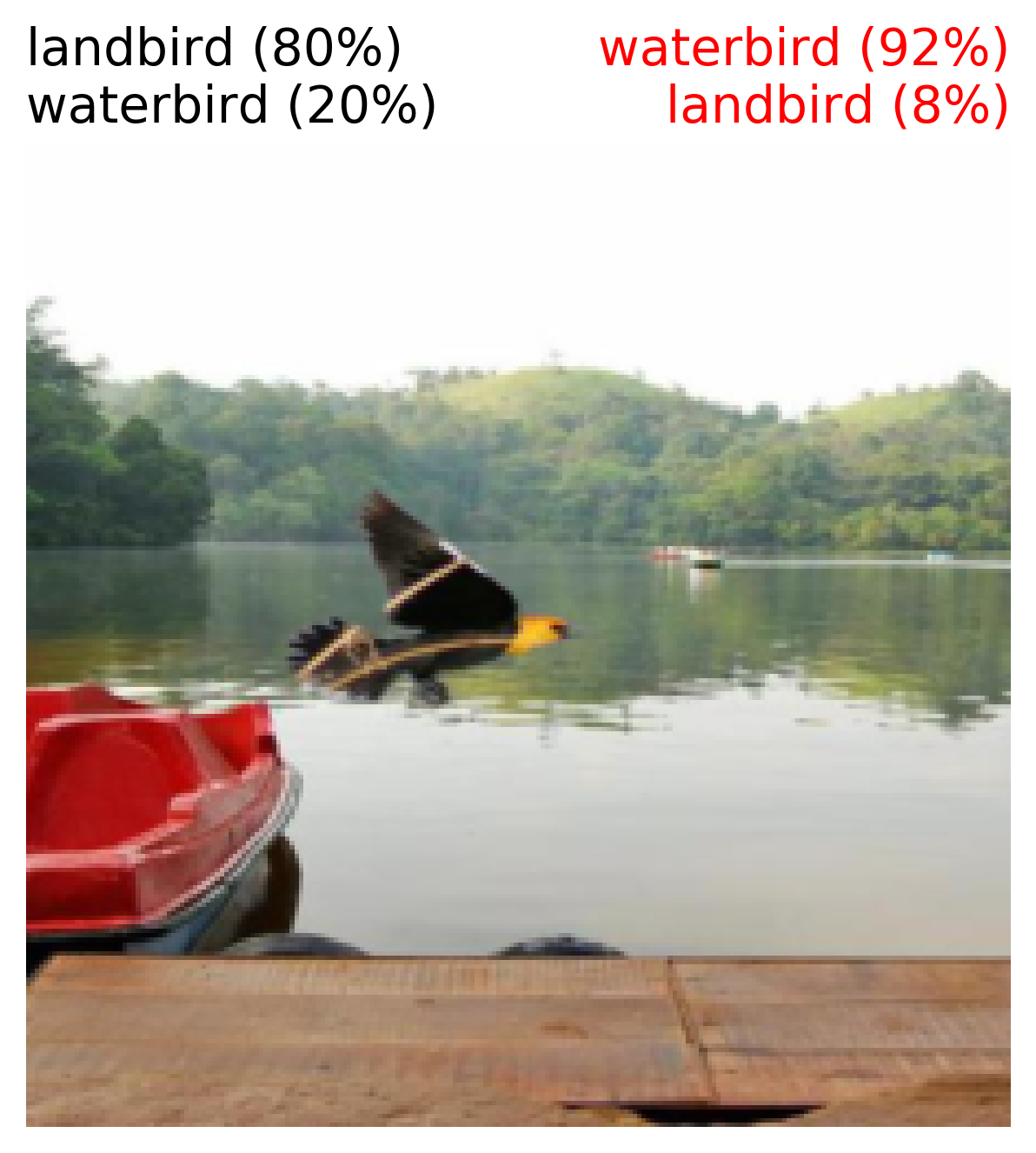} & 
  \includegraphics[width=0.166\linewidth]{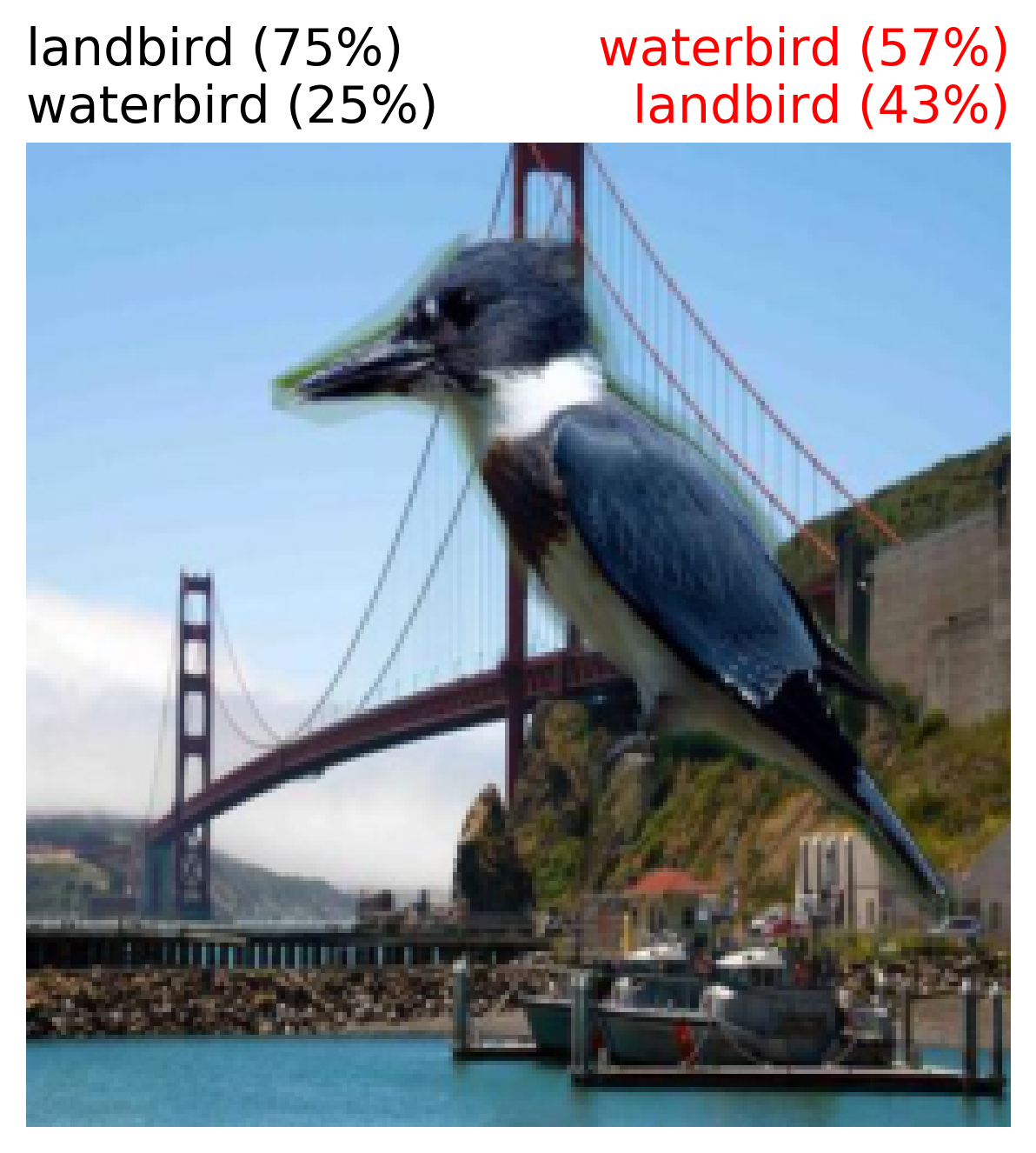} & 
  \includegraphics[width=0.166\linewidth]{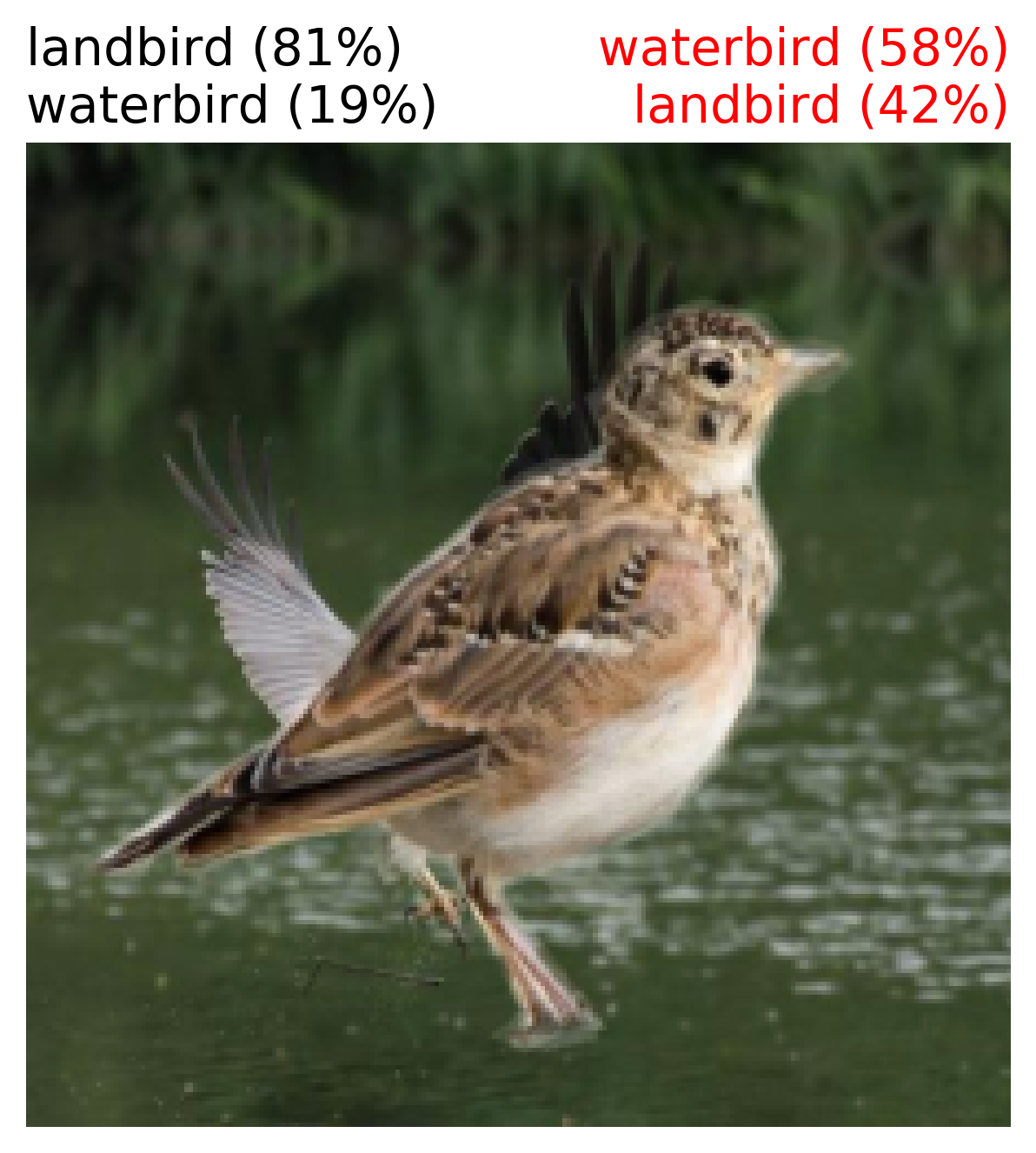} & 
  \includegraphics[width=0.166\linewidth]{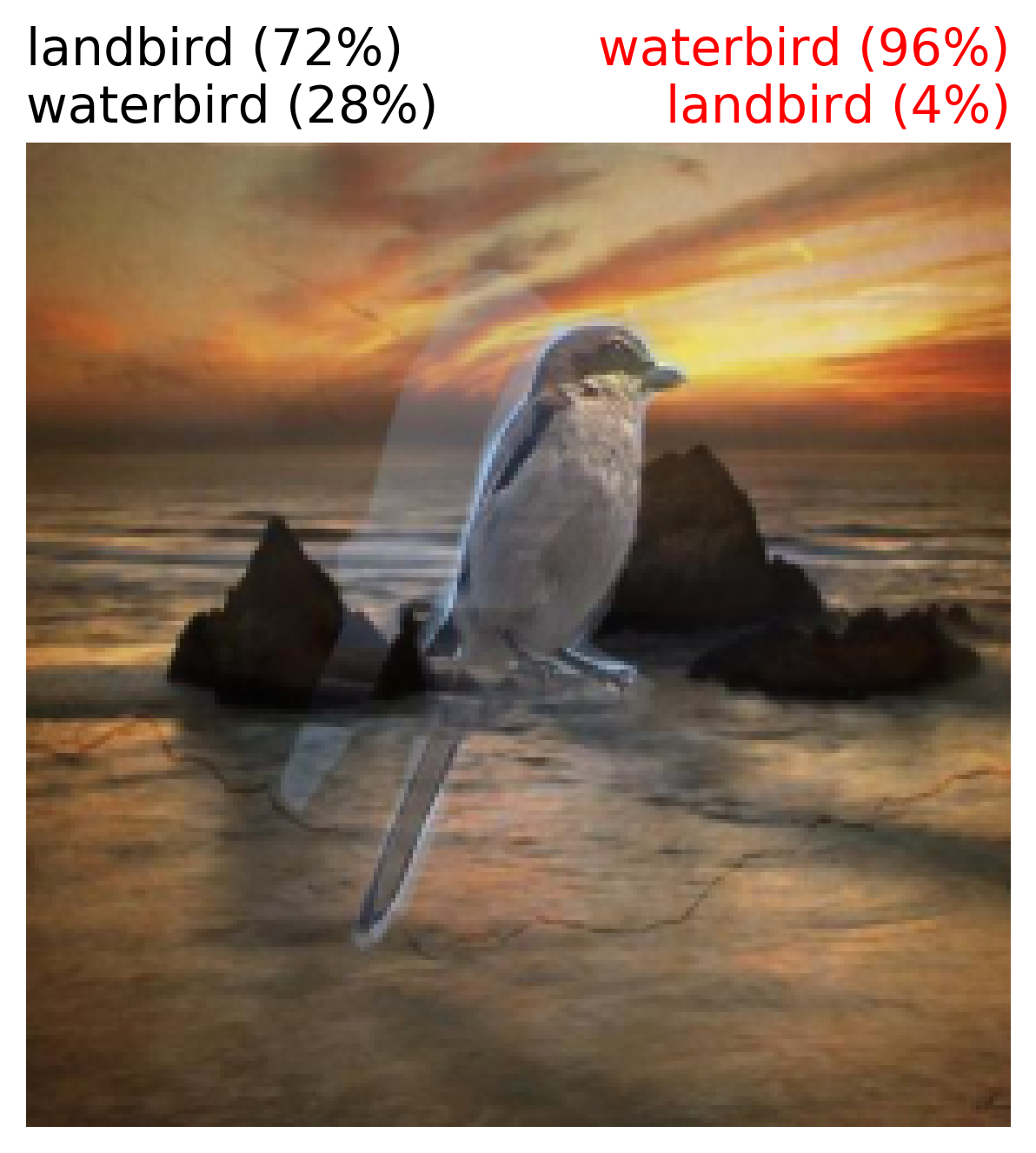} &
  \includegraphics[width=0.166\linewidth]{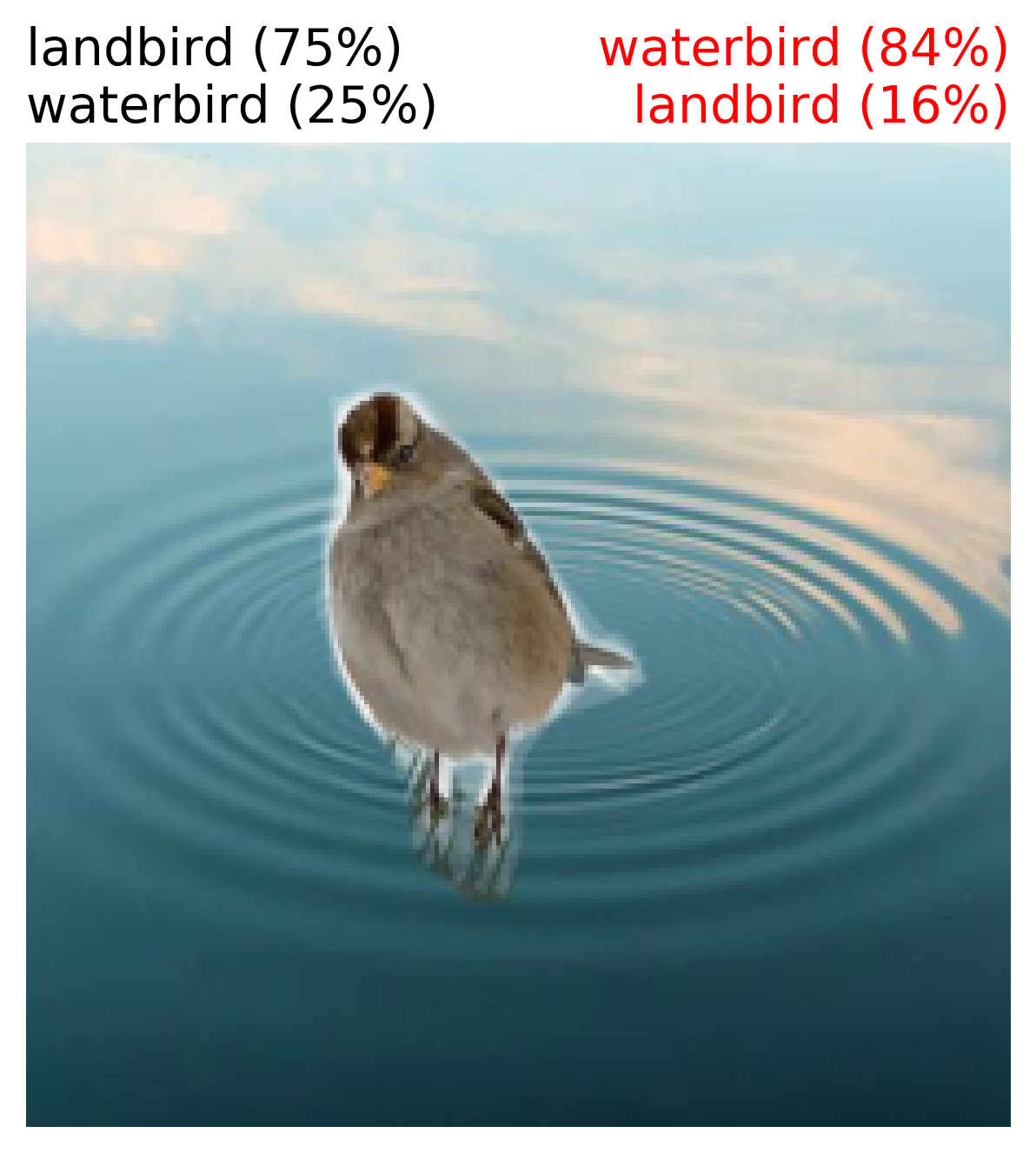} & 
  \includegraphics[width=0.166\linewidth]{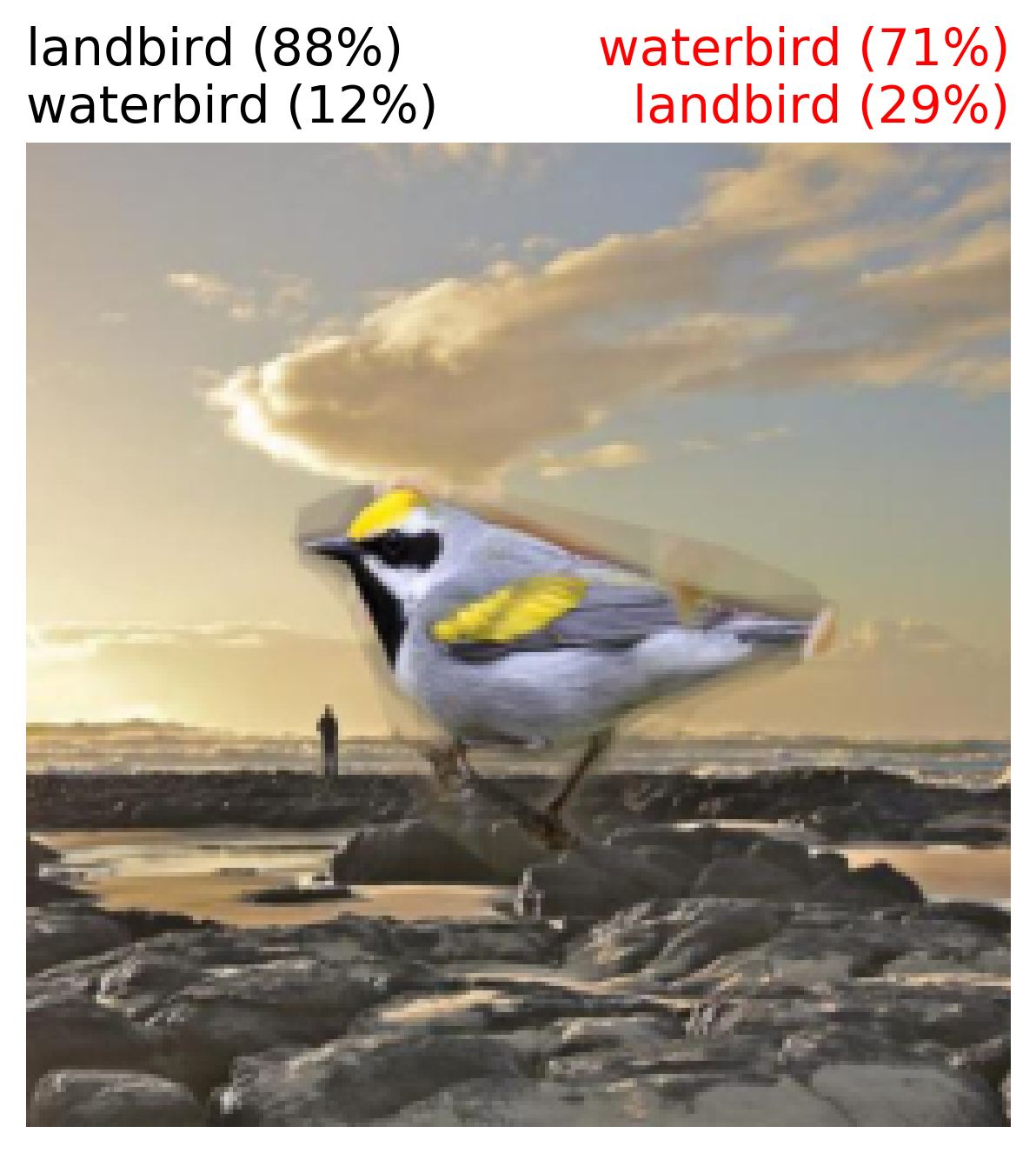}
\end{tabular}
\caption{Examples in the Waterbird Flip Test set that the best model (CF(CAGAN)+F(Random)+Sal) predicts correctly while Original model fails.
Grey texts are the predictions from the best model, and Red text are from Original model.}
\label{fig:wb_winning_cases}
\end{figure*}
\setlength\tabcolsep{6pt} 

To investigate the relationship between foregrounds focus and accuracy (details in Section~\ref{sec:in9}), we again show saliency AUPR and accuracy in Table~\ref{table:wb_sal_map}, and scatter plot all models in both Original and Flip test sets in Figure~\ref{fig:wb_sal_acc_scatter_plot}.
In Table~\ref{table:wb_sal_map}, all methods except CF(CAGAN) improve saliency AUPR (left column) in both Original and Flip test set with Sal($\lambda=10^3$) as the best method.
However, the improvement in saleincy AUPR does not come with better accuracy in both Original and Flip. For example, Sal($\lambda=10^3$) gets the best saliency AUPR while having the worst accuracy in Flip.
In Figure~\ref{fig:wb_sal_acc_scatter_plot}, we also find that in Flip there is a slightly better correlation between saliency focus and accuracy than in Original ($R^2=0.08 > 0.004$), but it is also not very strong.

In Figure~\ref{fig:wb_winning_cases}, we show some example images that our best model predicts correctly while the Original model fails.

\setlength\tabcolsep{0pt} 
\begin{figure*}[tbp]
\centering
\begin{tabular}{cccccc}

  \includegraphics[width=0.166\linewidth]{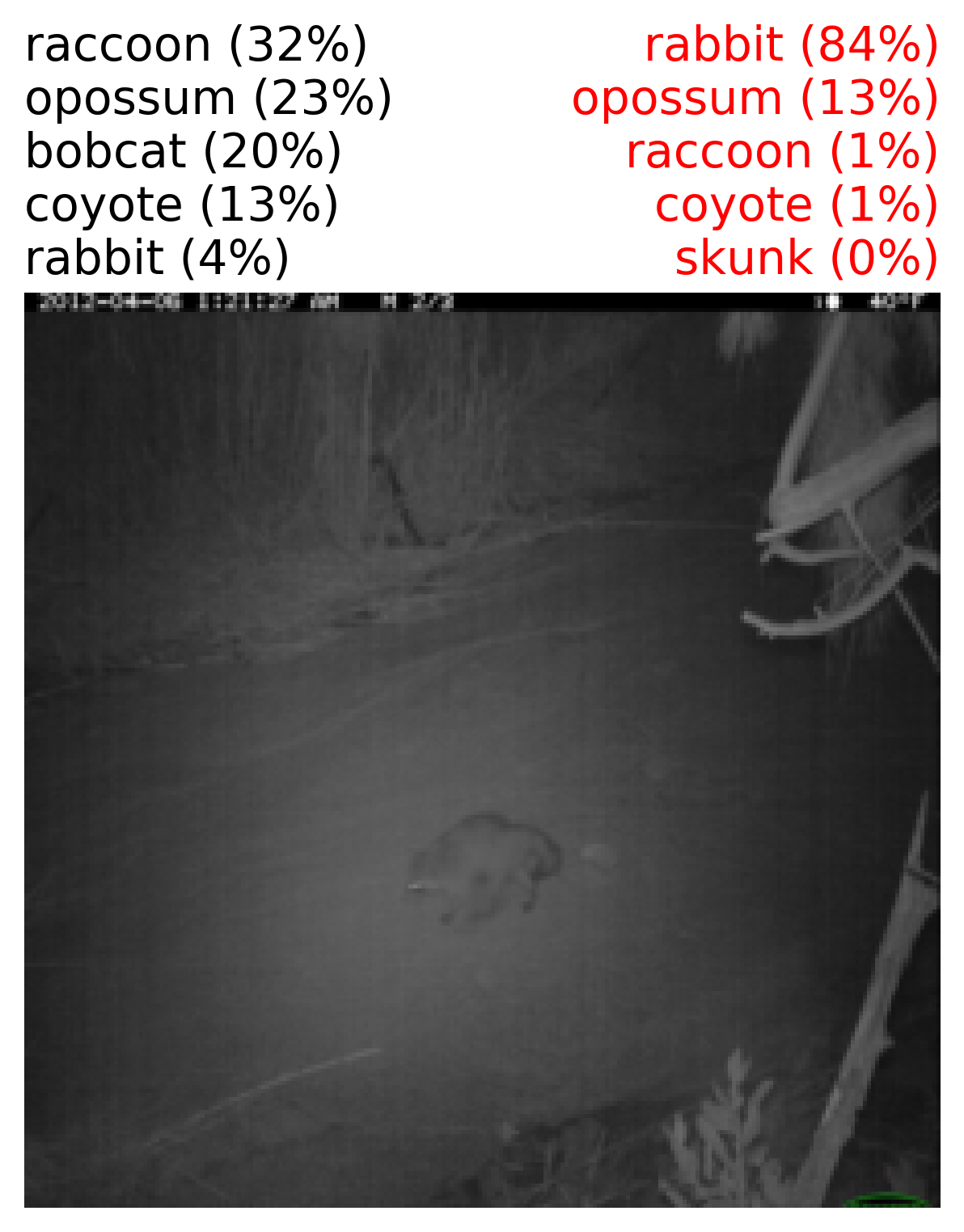} & 
  \includegraphics[width=0.166\linewidth]{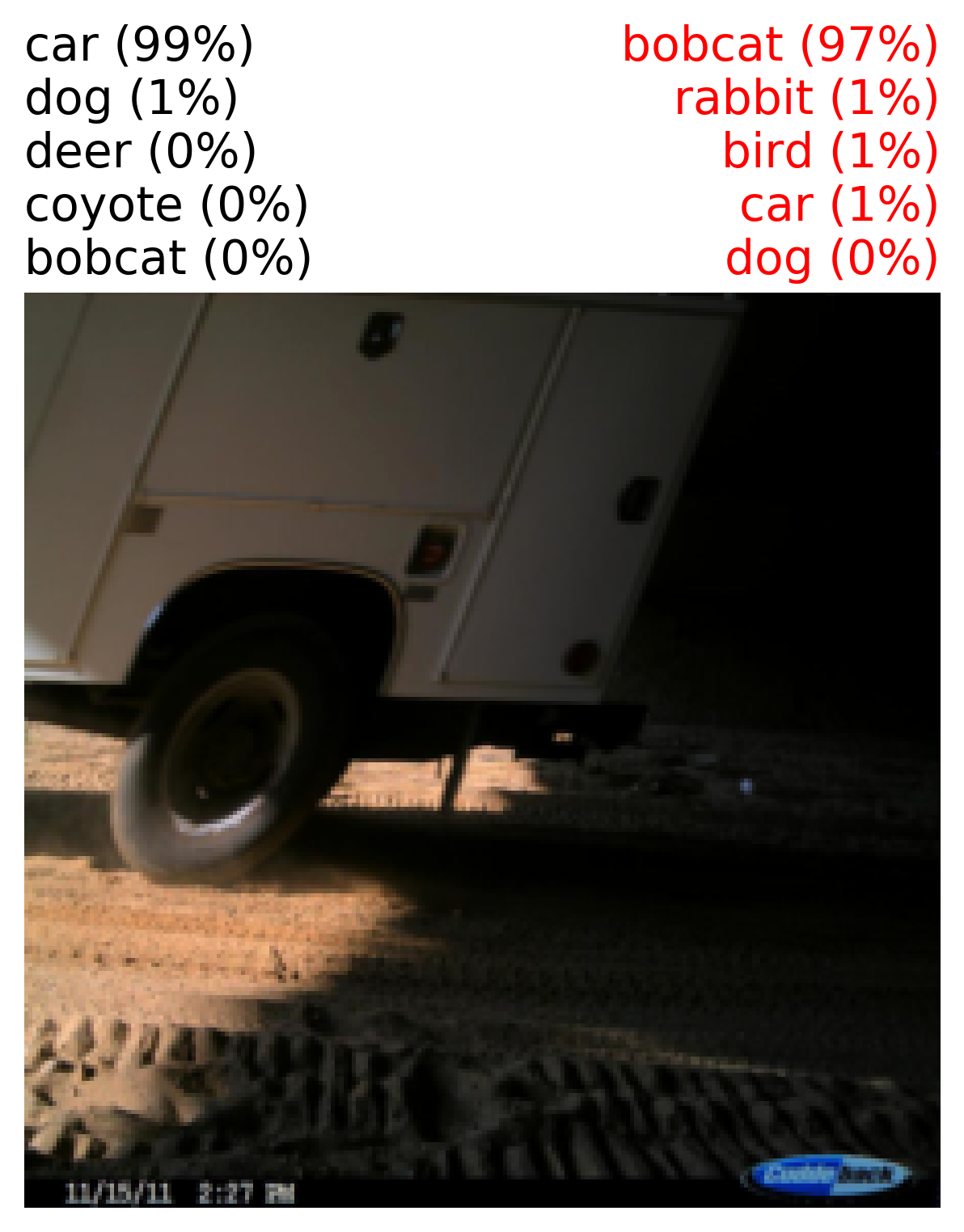} &
  \includegraphics[width=0.166\linewidth]{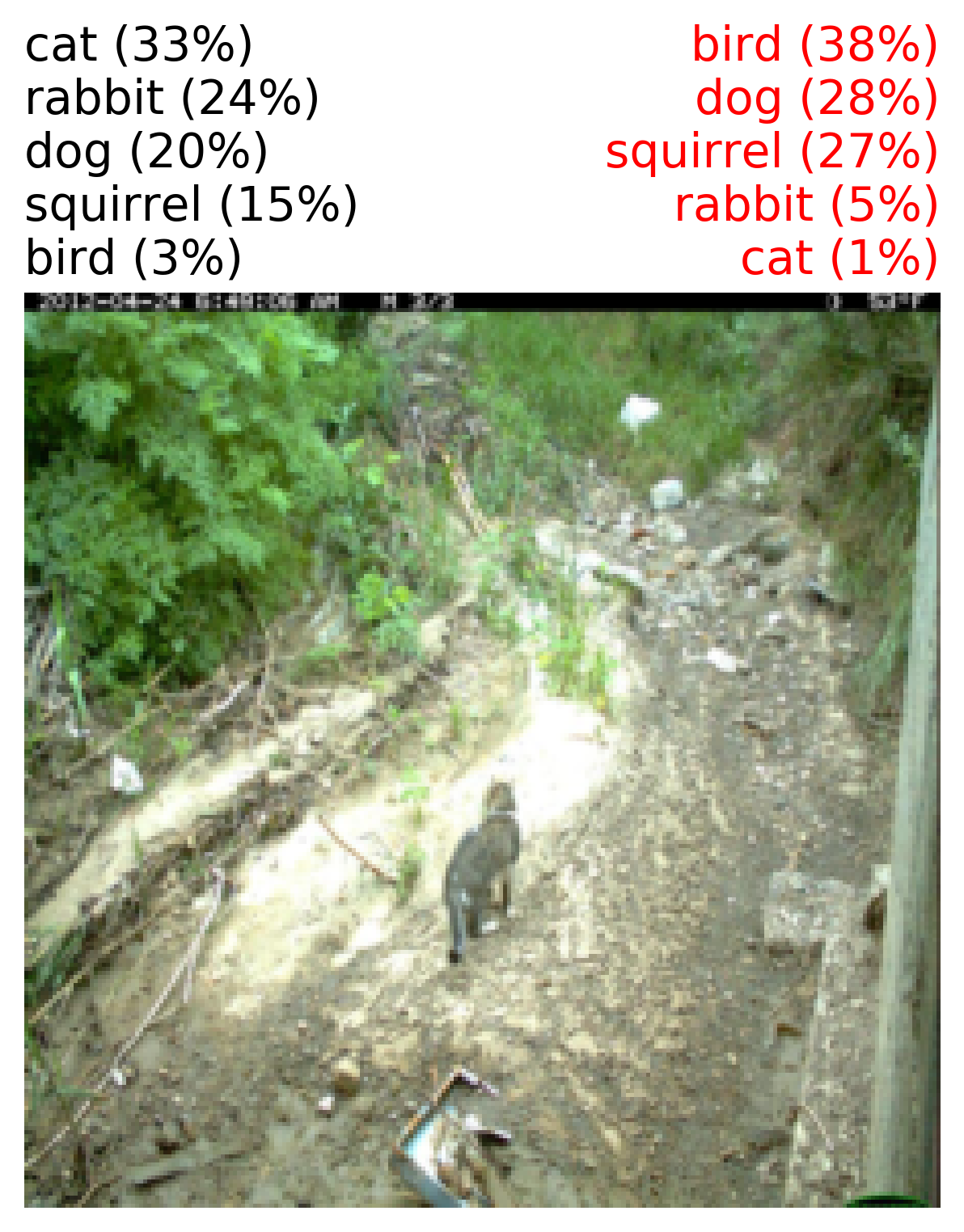} & 
  \includegraphics[width=0.166\linewidth]{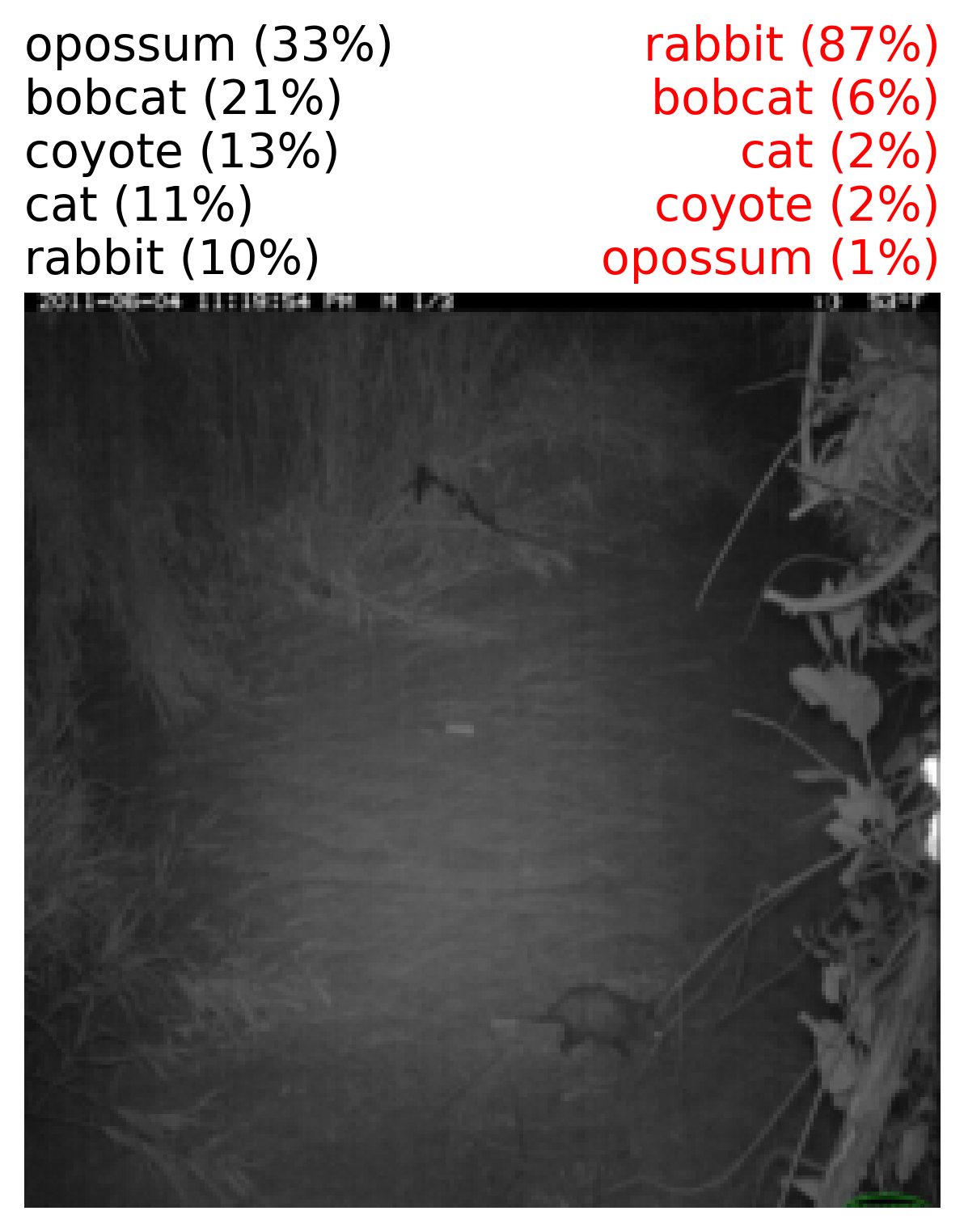} & 
  \includegraphics[width=0.166\linewidth]{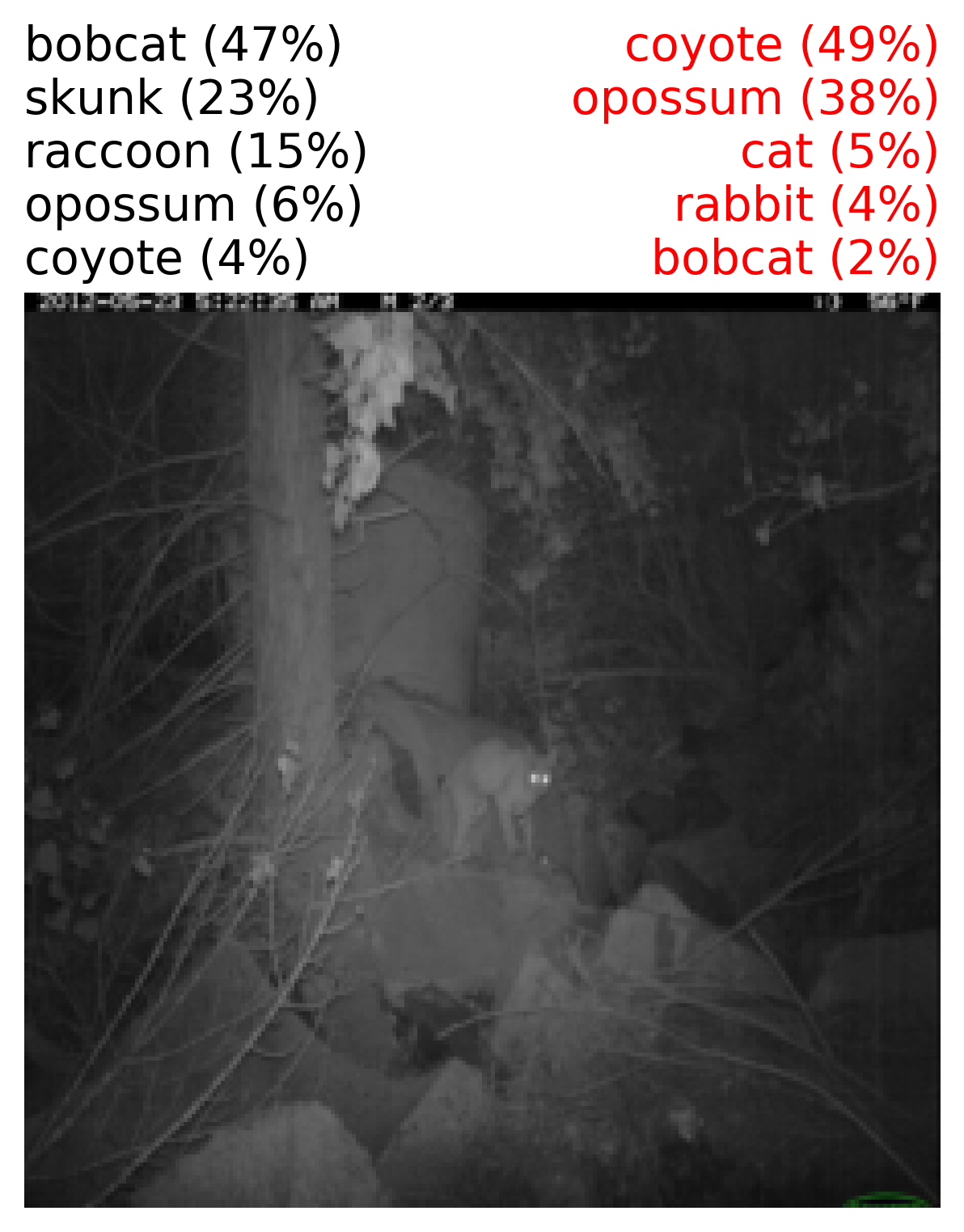} &
  \includegraphics[width=0.166\linewidth]{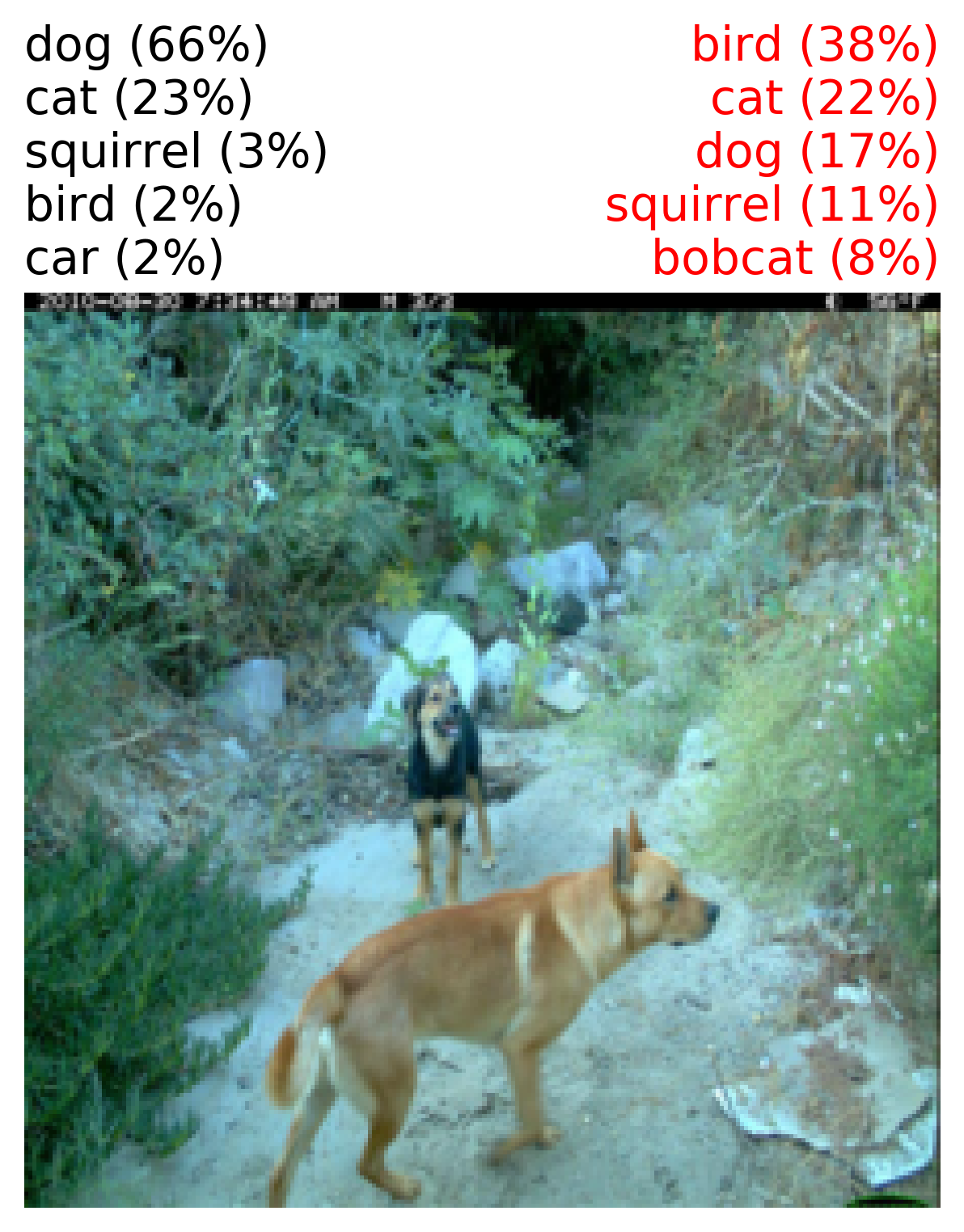} 
\end{tabular}
\caption{Examples of the CCT dataset that the best model (+CF(Tile)+Sal) predicts correctly but Original model fails. 
Grey texts are the predictions from the best model, and Red text are from Original model.}
\label{fig:cct_winning_cases}
\end{figure*}
\setlength\tabcolsep{6pt} 

\subsection{Caltech Camera Trap (CCT) dataset}


\setlength\tabcolsep{3pt} 
\begin{table}[tbp]
\caption{The performance results ($\%$ of AUC) of CCT datasets.
The Improv(\%). is the relative improvement of Trans-Test w.r.t. the Orig. method. We show mean and standard deviation for $3$ runs.}
\centering
\begin{tabular}{l|ccc}
 & Cis-Test           & Trans-Test & Improv(\%) \\ \toprule
Orig.                                        & $82.1\ (1.9)$ & $70.2\ (1.2)$  & $0.0$          \\ \midrule
+ Sal($\lambda=100$)                         & $86.6\ (1.2)$ & $72.2\ (1.1)$  & $2.8$          \\ \midrule
+ CF(Grey)                                   & $83.8\ (2.1)$ & $69.7\ (2.5)$  & $-0.7$         \\
+ CF(Random)                                 & $85.7\ (1.8)$ & $70.8\ (2.7)$  & $0.9$          \\
+ CF(Shuffle)                                & $84.2\ (0.2)$ & $68.9\ (1.5)$  & $-1.9$         \\
+ CF(Tile)                                   & $86.2\ (1.6)$ & $70.4\ (2.8)$  & $0.3$          \\
+ CF(CAGAN)                                  & $84.4\ (2.3)$ & $70.1\ (2.1)$  & $-0.1$         \\ \midrule
+ F(Random)                                  & $82.9\ (1.1)$ & $70.9\ (1.6)$  & $1.0$          \\
+ F(Shuffle)                                 & $84.2\ (2.4)$ & $70.8\ (1.9)$  & $0.9$          \\
+ F(Mixed-Rand)                              & $82.7\ (1.7)$ & $70.8\ (0.9)$  & $0.9$          \\
+ F(FGSM $\epsilon=0.5$)                     & $84.9\ (2.1)$ & $71.9\ (1.8)$  & $2.4$          \\ \midrule
+ CF(Tile) + Sal                             & $86.9\ (0.9)$ & $\bm{74.0}\ (1.0)$  & \bm{$5.4$} \\
+ CF(Tile) + F(Shuffle)                      & $85.0\ (1.3)$ & $72.3\ (2.1)$  & $3.0$          \\
+ F(Shuffle) + Sal                           & $\bm{88.5}\ (0.5)$ & $73.2\ (2.3)$  & $4.3$          \\ \midrule
\makecell[l]{+ CF(Tile)\\+ F(Shuffle) + Sal} & $87.0\ (1.4)$ & $73.2\ (1.1)$  & $4.3$          \\ \midrule
+ Mixup($\alpha=0.2$)                        & $85.0\ (0.5)$ & $69.4\ (1.7)$  & $-1.1$         \\
+ LS ($\epsilon=0.2$)                        & $81.2\ (0.7)$ & $67.8\ (3.0)$  & $-3.4$         \\         
\end{tabular}
\label{table:cct}
\end{table}

\setlength\tabcolsep{1pt} 
\begin{table}[tbp]
\caption{The AUPR of the saliency map (DeepLiftShap) of the ground truth class on CCT Cis and Trans datasets. The higher the model has better foreground focus.}
\centering
\begin{tabular}{l|cccc}
                        & \multicolumn{2}{c}{Saliency AUPR} & \multicolumn{2}{c}{AUC} \\
                        \cmidrule(lr){2-3}\cmidrule(lr){4-5}
         & Cis-Test              & Trans-Test    & Cis-Test      & Trans-Test    \\ \toprule
Orig     & $5.7\ (0.1)$          & $7.2\ (0.0)$ & $82.1\ (1.9)$ & $70.2\ (1.2)$ \\ \midrule
Sal($\lambda=100$)   & $6.4\ (0.5)$ & $7.5\ (0.3)$ & $86.6\ (1.1)$  & $72.2\ (1.1)$ \\
CF(Tile) & $6.2\ (0.5)$          & $7.8\ (0.5)$ & $86.2\ (1.6)$ & $70.4\ (2.8)$ \\
F(Shuffle)          & $7.3\ (1.1)$ & $9.3\ (1.8)$ & $84.2\ (2.4)$  & $70.8\ (1.9)$ \\
CF(Tile) + Sal    & $7.9\ (0.8)$ & $8.8\ (0.7)$ & $86.9\ (0.9)$  & $\bm{74.0}\ (1.0)$ \\
F(Shuffle) + Sal   & $\bm{8.1}\ (0.2)$ & $\bm{9.7}\ (0.5)$ & $\bm{88.5}\ (0.5)$  & $73.2\ (2.3)$ \\ \midrule
Sal($\lambda=10^4$) & $6.5\ (0.8)$ & $7.6\ (0.5)$ & $78.8\ (15.8)$ & $69.4\ (7.8)$ \\
\end{tabular}
\label{table:cct_sal_map}
\end{table}

\setlength\tabcolsep{0pt} 
\begin{figure}[tbp]
\centering
\begin{tabular}{ccc}
  & Cis-Test & Trans-Test   \\
  \raisebox{5\normalbaselineskip}[0pt][0pt]{\rotatebox[origin=c]{90}{AUC}} & \includegraphics[width=0.5\linewidth]{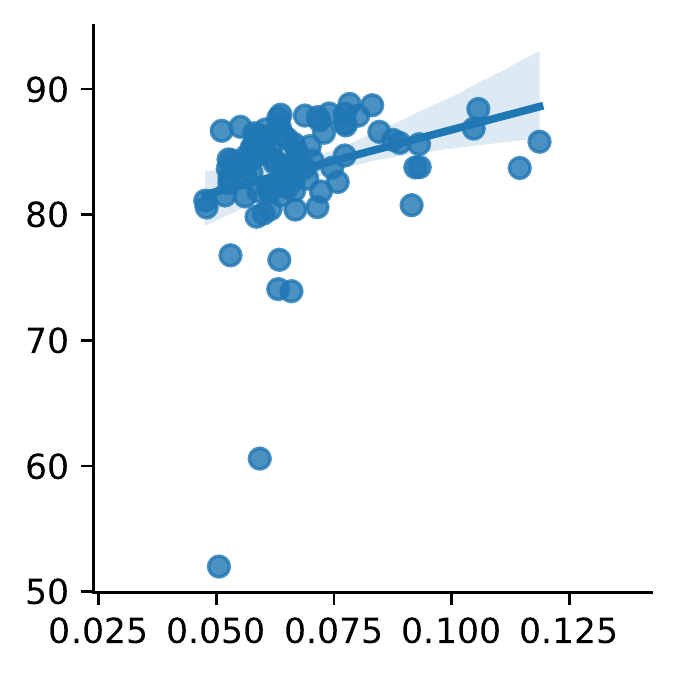} & 
  \includegraphics[width=0.5\linewidth]{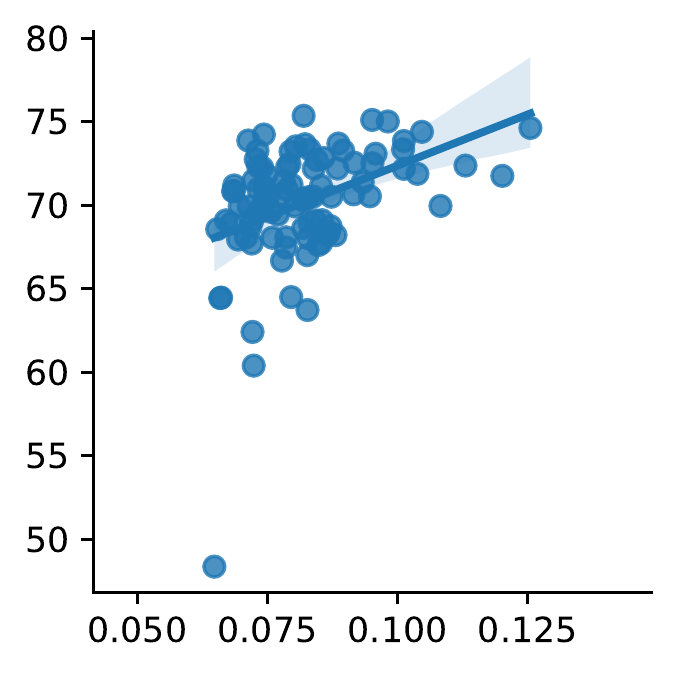}  \\ \vspace{-5pt}
  & \ \ \ \ Saliency AUPR & \ \ \ \ Saliency AUPR
\end{tabular}
\caption{The scatter plot between Saliency map (DeepLiftShap) AUPR and accuracy for all models we trained in CCT. In Trans-Test there is a stronger positive correlation ($R^2=0.16$) than Cis-Test ($R^2=0.07$) between saliency AUPR and accuracy.}
\label{fig:cct_sal_acc_scatter_plot}
\end{figure}
\setlength\tabcolsep{6pt} 

We test on a real-world dataset faced with background shifts across training and test sets.
Camera traps are motion- or heat-triggered cameras placed in locations of interest by biologists to monitor and study animal populations and behavior. 
The goal is to train a classifier that recognizes the same species of animals but with different camera backgrounds.
Caltech Camera Traps-20 (CCT) dataset~\citep{beery2018recognition}, consists of $57,868$ images across $20$ locations in the American Southwest, each labeled with one of $15$ classes of animals. 
We follow the setup of the original paper that divides test images into "cis-locations" and "trans-locations", where "cis-locations" are images with locations seen during training, and "trans" are new locations not seen before.
This gives us $13,553$ training images, $3,484$ validation and $15,827$ test images from cis-locations, and $23,275$ from trans-locations.
Since cis and trans locations have imbalanced classes, we use multi-class AUC to measure the performance. 

In Table~\ref{table:cct}, we find Sal($\lambda=100$) performs better than CF and F augmentations alone in both Cis and Trans split, although our combined method CF(Tile)+F(Shuffle) performs similarly with Sal in Trans-Test ($72.3$ v.s. $72.2$).
For CF augmentations, they mainly improve on Cis rather than Trans;
this shows that even within the same camera location like Cis dataset, there still exists spurious correlations probably due to wide variety of backgrounds (lightning, occulusions), relatively small bounding boxes, and small amount of training data.
Note that the pretrained generative model CAGAN is not fine-tuned on the CCT dataset, and further fine-tuning on this dataset can potentially give better inpaintings that may further improve the performance. 
For F augmentations, F(FGSM) helps the most in the Trans split.
Surprisingly, although F(FGSM) works best alone, when combining with CF(Tile) or Sal it gets worse performance; we think the adversarial nature of F(FGSM) might make the optimization harder.
Instead we try combinations with F(Shuffle).
We find both CF(Tile) + Sal and F(Shuffle) + Sal can further improve the performance in both Cis and Trans splits, again suggesting that CF, F and Sal improve in different ways.

We again analyze the relationship between saliency AUPR and accuracy (details in Section~\ref{sec:in9}).
In Table~\ref{table:cct_sal_map}, Sal, CF(Tile) and F(Shuffle) focus better on foregrounds with F(Shuffle) as the best.
The combination F(Shuffle)+Sal further improves focus.
Sal($\lambda=1$e$4$) with too strong regularization again increases saliency focus at the expense of lower accuracy.
In Figure~\ref{fig:cct_sal_acc_scatter_plot}, we scatter plot the relationship between test set AUC and foreground saliency AUPR.
They do not necessarily correlate well, and Trans-Test has better correlations than Cis-Test.

In Figure~\ref{fig:cct_winning_cases}, we show examples where the best model (CF(Tile)+Sal) succeeds but Original model fails in the Trans split.
Although it is unclear if the changed camera locations make the Original model wrongly classify, it can have very different top predictions from our best model.

\section{Discussions}

In this paper we focus on a particular type of spuriousness where the spurious and causal features are separated feature-wise which enables us to remove spuriousness using foreground annotations alone without knowing what the spurious factors are.
In the case where spuriousness happens in both foreground and background such as color or texture bias, our data augmentations would still work if we know what the spurious factors are, and generate corresponding Factual and Counterfactual data.
For example, if color blue is correlated with label, we can generate blue images without shape as Counterfactual images and generate random color images as Factual images.

We find our augmentations may vary performance from dataset to dataset, e.g. CF(Tile) does worse than no augmentation in IN-9, yet is one of the better performing methods for CCT. 
Overall we find CF methods perform best when imputation is natural as evidenced by better performance of CF(CAGAN) in IN-9 and Waterbirds. And we recommend F(Shuffle) or F(Random) for their better performance in our experiments.
The best approach is to try all different imputations to see what works well.

Lastly, we recognize that requiring additional annotations such as bounding boxes or segmentation maps can be costly for some datasets. This limitation can be overcome by using pre-trained segmentation models or heatmaps of pretrained models to obtain reasonable annotations. 
If some classes are novel, then few-shot semantic (FSS) segmentation can be used instead such that only a few images of novel class require manual segmentations, and the rest can be handled by the FSS model.

We believe that developing methods which make models more robust to spurious correlation is essential to overcoming the inherent obstacles to generalization posed by ambiguities in real-world datasets.

\section*{Acknowledgement}

Resources used in preparing this research were provided, in part, by the Province of Ontario, the Government of Canada through CIFAR, and companies sponsoring the Vector Institute \url{www.vectorinstitute.ai/#partners}.

{\small
\bibliographystyle{plainnat}
\bibliography{egbib}

\begin{thebibliography}{43}
\providecommand{\natexlab}[1]{#1}
\providecommand{\url}[1]{\texttt{#1}}
\expandafter\ifx\csname urlstyle\endcsname\relax
  \providecommand{\doi}[1]{doi: #1}\else
  \providecommand{\doi}{doi: \begingroup \urlstyle{rm}\Url}\fi

\bibitem[Agarwal et~al.(2020)Agarwal, Shetty, and Fritz]{agarwal2020towards}
Vedika Agarwal, Rakshith Shetty, and Mario Fritz.
\newblock Towards causal vqa: Revealing and reducing spurious correlations by
  invariant and covariant semantic editing.
\newblock In \emph{Proceedings of the IEEE/CVF Conference on Computer Vision
  and Pattern Recognition}, pages 9690--9698, 2020.

\bibitem[Agrawal et~al.(2016)Agrawal, Batra, and Parikh]{agrawal2016analyzing}
Aishwarya Agrawal, Dhruv Batra, and Devi Parikh.
\newblock Analyzing the behavior of visual question answering models.
\newblock In \emph{Proceedings of the 2016 Conference on Empirical Methods in
  Natural Language Processing}, pages 1955--1960, 2016.

\bibitem[Bao et~al.(2018)Bao, Chang, Yu, and Barzilay]{bao2018deriving}
Yujia Bao, Shiyu Chang, Mo~Yu, and Regina Barzilay.
\newblock Deriving machine attention from human rationales.
\newblock In \emph{EMNLP}, 2018.

\bibitem[Beery et~al.(2018)Beery, Van~Horn, and Perona]{beery2018recognition}
Sara Beery, Grant Van~Horn, and Pietro Perona.
\newblock Recognition in terra incognita.
\newblock In \emph{Proceedings of the European Conference on Computer Vision
  (ECCV)}, pages 456--473, 2018.

\bibitem[Belinkov and Bisk(2018)]{belinkov2018synthetic}
Yonatan Belinkov and Yonatan Bisk.
\newblock Synthetic and natural noise both break neural machine translation.
\newblock In \emph{International Conference on Learning Representations}, 2018.

\bibitem[Bissoto et~al.(2019)Bissoto, Fornaciali, Valle, and
  Avila]{bissoto2019constructing}
Alceu Bissoto, Michel Fornaciali, Eduardo Valle, and Sandra Avila.
\newblock (de) constructing bias on skin lesion datasets.
\newblock In \emph{Proceedings of the IEEE Conference on Computer Vision and
  Pattern Recognition Workshops}, pages 0--0, 2019.

\bibitem[Chang et~al.(2018)Chang, Creager, Goldenberg, and
  Duvenaud]{chang2018explaining}
Chun-Hao Chang, Elliot Creager, Anna Goldenberg, and David Duvenaud.
\newblock Explaining image classifiers by counterfactual generation.
\newblock In \emph{International Conference on Learning Representations}, 2018.

\bibitem[Chen et~al.(2020)Chen, Yan, Xiao, Zhang, Pu, and
  Zhuang]{chen2020counterfactual}
Long Chen, Xin Yan, Jun Xiao, Hanwang Zhang, Shiliang Pu, and Yueting Zhuang.
\newblock Counterfactual samples synthesizing for robust visual question
  answering.
\newblock In \emph{Proceedings of the IEEE/CVF Conference on Computer Vision
  and Pattern Recognition}, pages 10800--10809, 2020.

\bibitem[Du et~al.(2019)Du, Liu, Yang, and Hu]{du2019learning}
Mengnan Du, Ninghao Liu, Fan Yang, and Xia Hu.
\newblock Learning credible deep neural networks with rationale regularization.
\newblock In \emph{2019 IEEE International Conference on Data Mining (ICDM)},
  pages 150--159. IEEE, 2019.

\bibitem[Erion et~al.(2019)Erion, Janizek, Sturmfels, Lundberg, and
  Lee]{erion2019learning}
Gabriel Erion, Joseph~D Janizek, Pascal Sturmfels, Scott Lundberg, and Su-In
  Lee.
\newblock Learning explainable models using attribution priors.
\newblock \emph{arXiv preprint arXiv:1906.10670}, 2019.

\bibitem[Geirhos et~al.(2018)Geirhos, Rubisch, Michaelis, Bethge, Wichmann, and
  Brendel]{geirhos2018imagenet}
Robert Geirhos, Patricia Rubisch, Claudio Michaelis, Matthias Bethge, Felix~A
  Wichmann, and Wieland Brendel.
\newblock Imagenet-trained cnns are biased towards texture; increasing shape
  bias improves accuracy and robustness.
\newblock In \emph{International Conference on Learning Representations}, 2018.

\bibitem[Geirhos et~al.(2020)Geirhos, Jacobsen, Michaelis, Zemel, Brendel,
  Bethge, and Wichmann]{geirhos2020shortcut}
Robert Geirhos, J{\"o}rn-Henrik Jacobsen, Claudio Michaelis, Richard Zemel,
  Wieland Brendel, Matthias Bethge, and Felix~A Wichmann.
\newblock Shortcut learning in deep neural networks.
\newblock \emph{arXiv preprint arXiv:2004.07780}, 2020.

\bibitem[Ghaeini et~al.(2019)Ghaeini, Fern, Shahbazi, and
  Tadepalli]{ghaeini2019saliency}
Reza Ghaeini, Xiaoli Fern, Hamed Shahbazi, and Prasad Tadepalli.
\newblock Saliency learning: Teaching the model where to pay attention.
\newblock In \emph{Proceedings of the 2019 Conference of the North American
  Chapter of the Association for Computational Linguistics: Human Language
  Technologies, Volume 1 (Long and Short Papers)}, pages 4016--4025, 2019.

\bibitem[Goodfellow et~al.(2014)Goodfellow, Shlens, and
  Szegedy]{goodfellow2014explaining}
Ian~J Goodfellow, Jonathon Shlens, and Christian Szegedy.
\newblock Explaining and harnessing adversarial examples.
\newblock \emph{arXiv preprint arXiv:1412.6572}, 2014.

\bibitem[Gururangan et~al.(2018)Gururangan, Swayamdipta, Levy, Schwartz,
  Bowman, and Smith]{gururangan2018annotation}
Suchin Gururangan, Swabha Swayamdipta, Omer Levy, Roy Schwartz, Samuel~R
  Bowman, and Noah~A Smith.
\newblock Annotation artifacts in natural language inference data.
\newblock \emph{arXiv preprint arXiv:1803.02324}, 2018.

\bibitem[Kaushik et~al.(2020)Kaushik, Hovy, and Lipton]{Kaushik2020Learning}
Divyansh Kaushik, Eduard Hovy, and Zachary Lipton.
\newblock Learning the difference that makes a difference with
  counterfactually-augmented data.
\newblock In \emph{International Conference on Learning Representations}, 2020.
\newblock URL \url{https://openreview.net/forum?id=Sklgs0NFvr}.

\bibitem[Kolesnikov et~al.(2019)Kolesnikov, Beyer, Zhai, Puigcerver, Yung,
  Gelly, and Houlsby]{kolesnikov2019large}
Alexander Kolesnikov, Lucas Beyer, Xiaohua Zhai, Joan Puigcerver, Jessica Yung,
  Sylvain Gelly, and Neil Houlsby.
\newblock Large scale learning of general visual representations for transfer.
\newblock \emph{arXiv preprint arXiv:1912.11370}, 2019.

\bibitem[Kusner et~al.(2017)Kusner, Loftus, Russell, and
  Silva]{kusner2017counterfactual}
Matt~J Kusner, Joshua Loftus, Chris Russell, and Ricardo Silva.
\newblock Counterfactual fairness.
\newblock In \emph{Advances in neural information processing systems}, pages
  4066--4076, 2017.

\bibitem[Lu et~al.(2020)Lu, Mardziel, Wu, Amancharla, and Datta]{lu2020gender}
Kaiji Lu, Piotr Mardziel, Fangjing Wu, Preetam Amancharla, and Anupam Datta.
\newblock Gender bias in neural natural language processing.
\newblock In \emph{Logic, Language, and Security}, pages 189--202. Springer,
  2020.

\bibitem[Lundberg and Lee(2017)]{lundberg2017unified}
Scott~M Lundberg and Su-In Lee.
\newblock A unified approach to interpreting model predictions.
\newblock In \emph{Advances in neural information processing systems}, pages
  4765--4774, 2017.

\bibitem[Madras et~al.(2018)Madras, Creager, Pitassi, and
  Zemel]{madras2018learning}
David Madras, Elliot Creager, Toniann Pitassi, and Richard Zemel.
\newblock Learning adversarially fair and transferable representations.
\newblock In \emph{International Conference on Machine Learning}, pages
  3384--3393, 2018.

\bibitem[Madry et~al.(2018)Madry, Makelov, Schmidt, Tsipras, and
  Vladu]{madry2018towards}
Aleksander Madry, Aleksandar Makelov, Ludwig Schmidt, Dimitris Tsipras, and
  Adrian Vladu.
\newblock Towards deep learning models resistant to adversarial attacks.
\newblock In \emph{International Conference on Learning Representations}, 2018.

\bibitem[Maguolo and Nanni(2020)]{maguolo2020critic}
Gianluca Maguolo and Loris Nanni.
\newblock A critic evaluation of methods for covid-19 automatic detection from
  x-ray images.
\newblock \emph{arXiv preprint arXiv:2004.12823}, 2020.

\bibitem[Mitsuhara et~al.(2019)Mitsuhara, Fukui, Sakashita, Ogata, Hirakawa,
  Yamashita, and Fujiyoshi]{Mitsuhara2019EmbeddingHK}
Masahiro Mitsuhara, H.~Fukui, Yusuke Sakashita, T.~Ogata, Tsubasa Hirakawa,
  T.~Yamashita, and H.~Fujiyoshi.
\newblock Embedding human knowledge in deep neural network via attention map.
\newblock \emph{ArXiv}, abs/1905.03540, 2019.

\bibitem[M{\"u}ller et~al.(2019)M{\"u}ller, Kornblith, and
  Hinton]{muller2019does}
Rafael M{\"u}ller, Simon Kornblith, and Geoffrey~E Hinton.
\newblock When does label smoothing help?
\newblock In \emph{Advances in Neural Information Processing Systems}, pages
  4694--4703, 2019.

\bibitem[Rieger et~al.(2020)Rieger, Singh, Murdoch, and
  Yu]{rieger2020interpretations}
Laura Rieger, Chandan Singh, William Murdoch, and Bin Yu.
\newblock Interpretations are useful: penalizing explanations to align neural
  networks with prior knowledge.
\newblock In \emph{International Conference on Machine Learning}, pages
  8116--8126. PMLR, 2020.

\bibitem[Ross and Doshi-Velez(2018)]{ross2018improving}
Andrew Ross and Finale Doshi-Velez.
\newblock Improving the adversarial robustness and interpretability of deep
  neural networks by regularizing their input gradients.
\newblock In \emph{Proceedings of the AAAI Conference on Artificial
  Intelligence}, volume~32, 2018.

\bibitem[Ross et~al.(2017)Ross, Hughes, and Doshi-Velez]{ross2017right}
Andrew~Slavin Ross, Michael~C Hughes, and Finale Doshi-Velez.
\newblock Right for the right reasons: training differentiable models by
  constraining their explanations.
\newblock In \emph{Proceedings of the 26th International Joint Conference on
  Artificial Intelligence}, pages 2662--2670, 2017.

\bibitem[Sagawa et~al.(2020)Sagawa, Koh, Hashimoto, and
  Liang]{Sagawa2020Distributionally}
Shiori Sagawa, Pang~Wei Koh, Tatsunori~B. Hashimoto, and Percy Liang.
\newblock Distributionally robust neural networks.
\newblock In \emph{International Conference on Learning Representations}, 2020.
\newblock URL \url{https://openreview.net/forum?id=ryxGuJrFvS}.

\bibitem[Selvaraju et~al.(2017)Selvaraju, Cogswell, Das, Vedantam, Parikh, and
  Batra]{selvaraju2017grad}
Ramprasaath~R Selvaraju, Michael Cogswell, Abhishek Das, Ramakrishna Vedantam,
  Devi Parikh, and Dhruv Batra.
\newblock Grad-cam: Visual explanations from deep networks via gradient-based
  localization.
\newblock In \emph{Proceedings of the IEEE international conference on computer
  vision}, pages 618--626, 2017.

\bibitem[Simpson et~al.(2019)Simpson, Dutil, Bengio, and
  Cohen]{simpson2019gradmask}
Becks Simpson, Francis Dutil, Yoshua Bengio, and Joseph~Paul Cohen.
\newblock Gradmask: Reduce overfitting by regularizing saliency.
\newblock In \emph{International Conference on Medical Imaging with Deep
  Learning--Extended Abstract Track}, 2019.

\bibitem[Singh et~al.(2018)Singh, Murdoch, and Yu]{singh2018hierarchical}
Chandan Singh, W~James Murdoch, and Bin Yu.
\newblock Hierarchical interpretations for neural network predictions.
\newblock In \emph{International Conference on Learning Representations}, 2018.

\bibitem[Viviano et~al.(2019)Viviano, Simpson, Dutil, Bengio, and
  Cohen]{viviano2019underwhelming}
Joseph~D Viviano, Becks Simpson, Francis Dutil, Yoshua Bengio, and Joseph~Paul
  Cohen.
\newblock Underwhelming generalization improvements from controlling feature
  attribution.
\newblock \emph{arXiv preprint arXiv:1910.00199}, 2019.

\bibitem[Wah et~al.(2011)Wah, Branson, Welinder, Perona, and
  Belongie]{WahCUB_200_2011}
C.~Wah, S.~Branson, P.~Welinder, P.~Perona, and S.~Belongie.
\newblock {The Caltech-UCSD Birds-200-2011 Dataset}.
\newblock Technical Report CNS-TR-2011-001, California Institute of Technology,
  2011.

\bibitem[Xiao et~al.(2020)Xiao, Engstrom, Ilyas, and Madry]{xiao2020noise}
Kai Xiao, Logan Engstrom, Andrew Ilyas, and Aleksander Madry.
\newblock Noise or signal: The role of image backgrounds in object recognition.
\newblock \emph{arXiv preprint arXiv:2006.09994}, 2020.

\bibitem[Young et~al.(2019)Young, Booth, Simpson, Dutton, and
  Shrapnel]{young2019deep}
Kyle Young, Gareth Booth, Becks Simpson, Reuben Dutton, and Sally Shrapnel.
\newblock Deep neural network or dermatologist?
\newblock In \emph{Interpretability of Machine Intelligence in Medical Image
  Computing and Multimodal Learning for Clinical Decision Support}, pages
  48--55. Springer, 2019.

\bibitem[Yu et~al.(2018)Yu, Lin, Yang, Shen, Lu, and Huang]{yu2018cagan}
Jiahui Yu, Zhe Lin, Jimei Yang, Xiaohui Shen, Xin Lu, and Thomas~S Huang.
\newblock Generative image inpainting with contextual attention.
\newblock In \emph{Proceedings of the IEEE Conference on Computer Vision and
  Pattern Recognition}, pages 5505--5514, 2018.

\bibitem[Zech et~al.(2018)Zech, Badgeley, Liu, Costa, Titano, and
  Oermann]{zech2018confounding}
John~R Zech, Marcus~A Badgeley, Manway Liu, Anthony~B Costa, Joseph~J Titano,
  and Eric~K Oermann.
\newblock Confounding variables can degrade generalization performance of
  radiological deep learning models.
\newblock \emph{arXiv preprint arXiv:1807.00431}, 2018.

\bibitem[Zemel et~al.(2013)Zemel, Wu, Swersky, Pitassi, and
  Dwork]{zemel2013learning}
Rich Zemel, Yu~Wu, Kevin Swersky, Toni Pitassi, and Cynthia Dwork.
\newblock Learning fair representations.
\newblock In \emph{International Conference on Machine Learning}, pages
  325--333, 2013.

\bibitem[Zhang et~al.(2018)Zhang, Cisse, Dauphin, and
  Lopez-Paz]{zhang2018mixup}
Hongyi Zhang, Moustapha Cisse, Yann~N Dauphin, and David Lopez-Paz.
\newblock mixup: Beyond empirical risk minimization.
\newblock In \emph{International Conference on Learning Representations}, 2018.

\bibitem[Zhou et~al.(2014)Zhou, Lapedriza, Xiao, Torralba, and
  Oliva]{zhou2014learning}
Bolei Zhou, Agata Lapedriza, Jianxiong Xiao, Antonio Torralba, and Aude Oliva.
\newblock Learning deep features for scene recognition using places database.
\newblock 2014.

\bibitem[Zhuang et~al.(2019)Zhuang, Cai, Wang, Zhang, and
  Zheng]{zhuang2019care}
Jiaxin Zhuang, Jiabin Cai, Ruixuan Wang, Jianguo Zhang, and Weishi Zheng.
\newblock Care: Class attention to regions of lesion for classification on
  imbalanced data.
\newblock In \emph{International Conference on Medical Imaging with Deep
  Learning}, pages 588--597, 2019.

\bibitem[Zmigrod et~al.(2019)Zmigrod, Mielke, Wallach, and
  Cotterell]{zmigrod2019counterfactual}
Ran Zmigrod, Sebastian~J Mielke, Hanna Wallach, and Ryan Cotterell.
\newblock Counterfactual data augmentation for mitigating gender stereotypes in
  languages with rich morphology.
\newblock In \emph{Proceedings of the 57th Annual Meeting of the Association
  for Computational Linguistics}, pages 1651--1661, 2019.

\end{thebibliography}
}


\appendix

\section{Training details}
\label{appx:training_details}

\subsection{Hyperparameters}
Here we describe the hyperparameters to train our model in Table~\ref{table:hyperparams}.
We follow the training details in the original paper accompanied with these datasets as close as possible which is why there are some differences of hyperparameters among datasets.
We set our maximum epochs to be large enough such that the performance saturates.
We use floating point $16$ to speed up the training.
To our surprise in waterbirds when doing finetuning, floating point 16 is crucial to get superior performance and we use it for all our experiments.
We release our code in \url{https://github.com/zzzace2000/robust_cls_model}.

\setlength\tabcolsep{0.5pt} 
\begin{table}[!h]
\caption{Training hyperparameters for each dataset}
\begin{tabular}{c|ccc}
 & IN-9         & Waterbirds         & \makecell{Caltech\\Camera Trap} \\ \toprule
Model                & BiT-S-R50x1 & BiT-S-R50x1 & BiT-S-R50x1         \\ \midrule
Epochs               & 25          & 40                              & 50                  \\ \midrule
LR                   & 0.05        & 8.00E-04                        & 0.003               \\ \midrule
Optimizer &
  \makecell{SGD with\\momentum 0.9} &
  \makecell{SGD with\\momentum 0.9} &
  \makecell{RMSProp with\\momentum 0.9} \\ \midrule
Weight decay         & 1e-4        & 1.00E-04                        & 0                   \\ \midrule
LR Scheduling &
  \makecell{Decay 1/10\\after 6, 12,\\18 epochs} &
  \makecell{Reduce LR on\\Plateau with\\patience 1} &
  \makecell{Decay 1/10\\after 15, 30,\\45 epochs} \\ \midrule
Batch size           & 32          & 32                              & 64                  \\ \midrule
Early stopping       & Yes         & Yes         & Yes                 \\ \midrule
Finetuning & No & Yes & No \\
\end{tabular}
\label{table:hyperparams}
\end{table}

\subsection{Tiled background generation}
\label{appx:tile_generation}

Here we show how to generate tiled background in Algorithm~\ref{alg:tile}.

\begin{algorithm}[htbp]
  \caption{Tiled background generation}
  \label{alg:tile}
\begin{algorithmic}
  \STATE {\bfseries Input:} An image $x$ and important region $r$ \\
  \textbf{Output:} Tiled image $\phi_{tile}$ \\
   
  $A \leftarrow$ the largest rectangular regions that $r=0$ \\
  $w, h \leftarrow$ $x$ width, $x$ height \\
  $a_w, a_h \leftarrow$ $A$ width, $A$ height \\
  \IF{$a_w < w$}
        \STATE
          repeat A $\lceil \frac{w}{a_w}\rceil$ times horizontally \\
  \ENDIF
  \IF{$a_h < h$}
        \STATE
          repeat A $\lceil\frac{h}{a_h}\rceil$ times vertically \\
  \ENDIF
   
  $A \leftarrow A[:w, :h]$ \\
  $\phi_{tile} = x \dot r + A \dot (1 - r) $ \\
\end{algorithmic}
\end{algorithm}

\section{Additional results}

\subsection{Qualitative examples}

\begin{figure*}[tbp]
\caption{Random examples that the best model (Grey) (CF(CAGAN)+F(Shuffle)+Sal) predicts correctly while Original model (Red) fails in IN9 Mixed-Next dataset. Here we show their top 5 predictions.  
}
\centering
\begin{tabular}{cccccc}
  \includegraphics[width=0.166\linewidth]{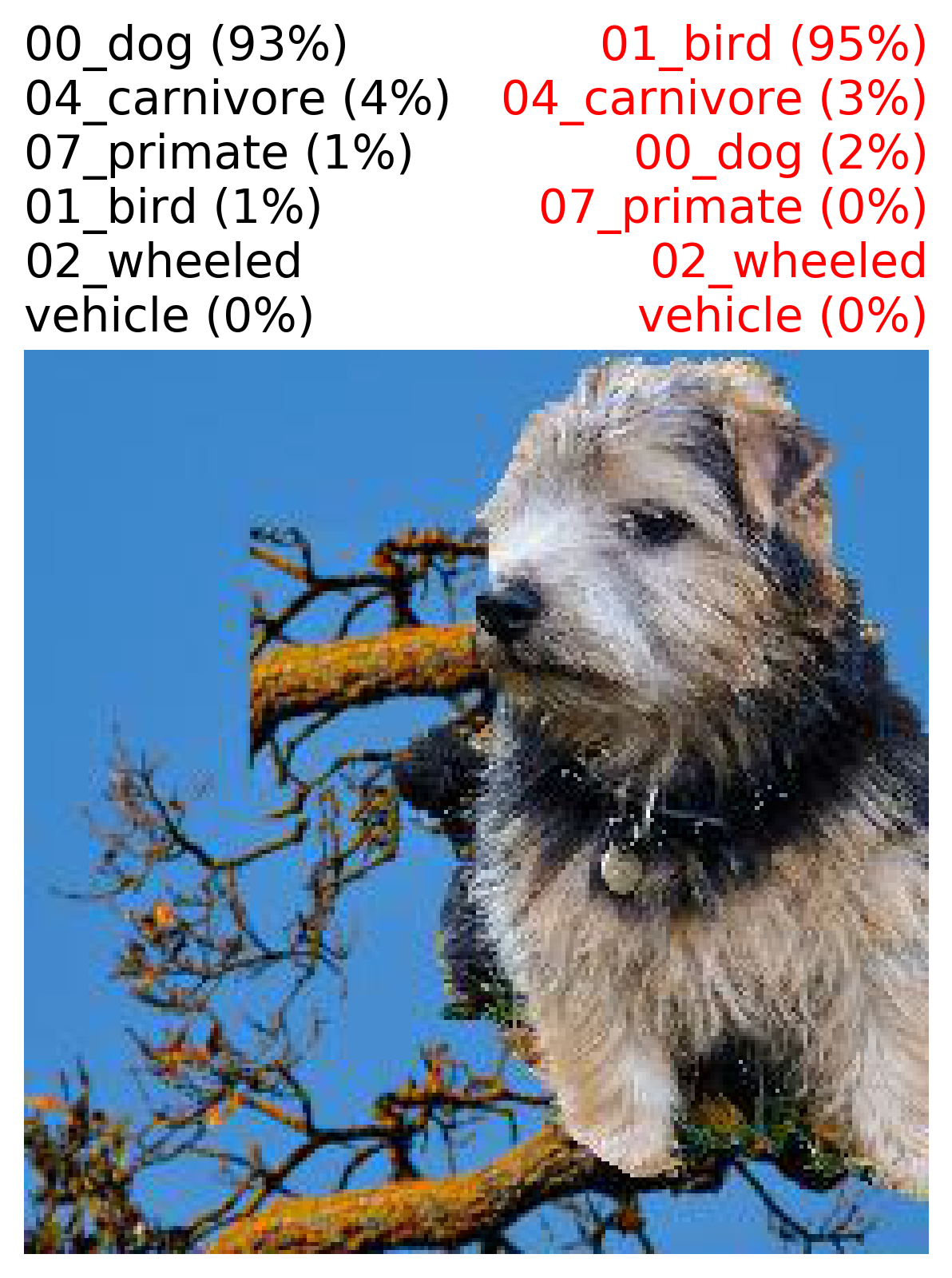} & 
  \includegraphics[width=0.166\linewidth]{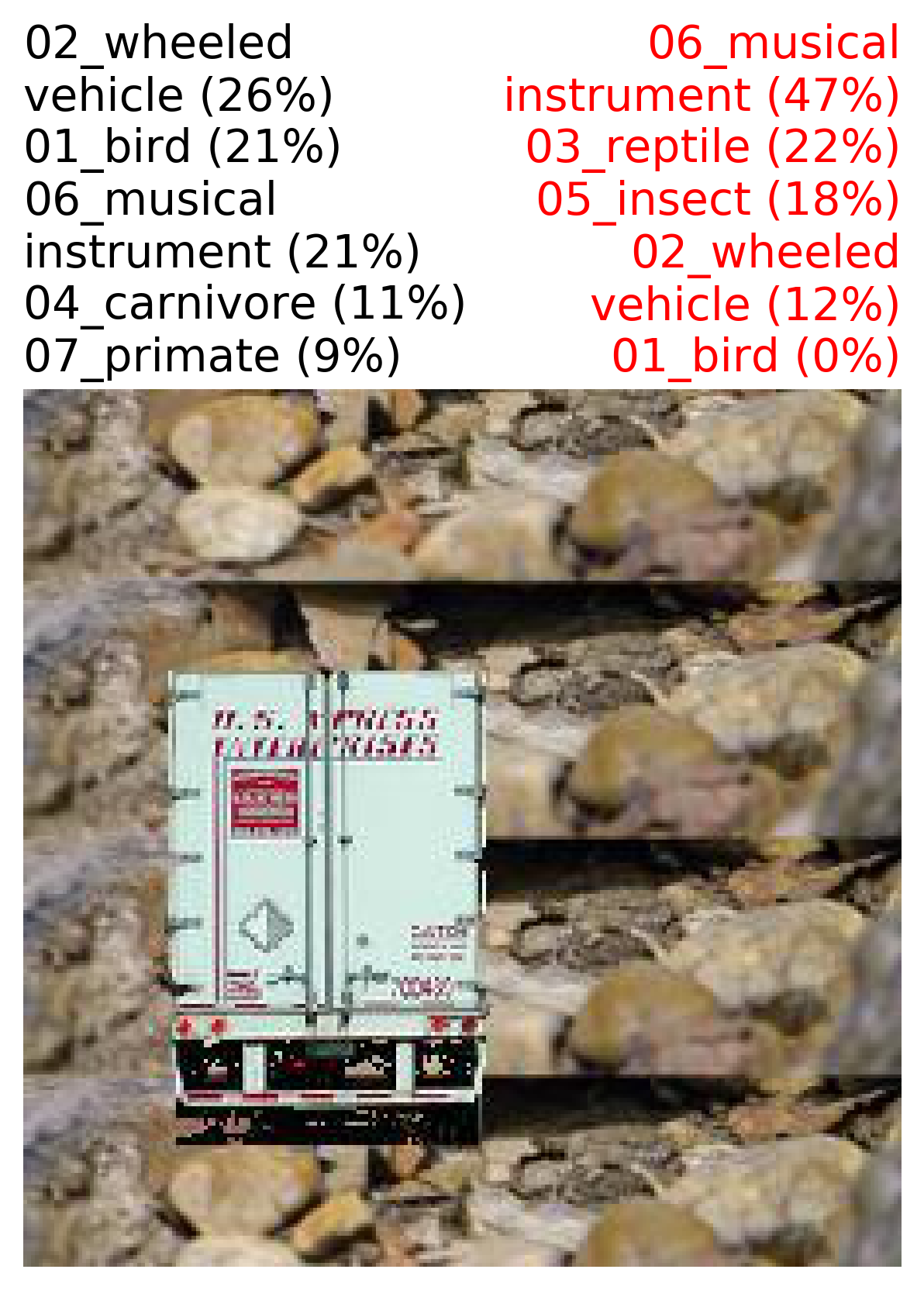} & 
  \includegraphics[width=0.166\linewidth]{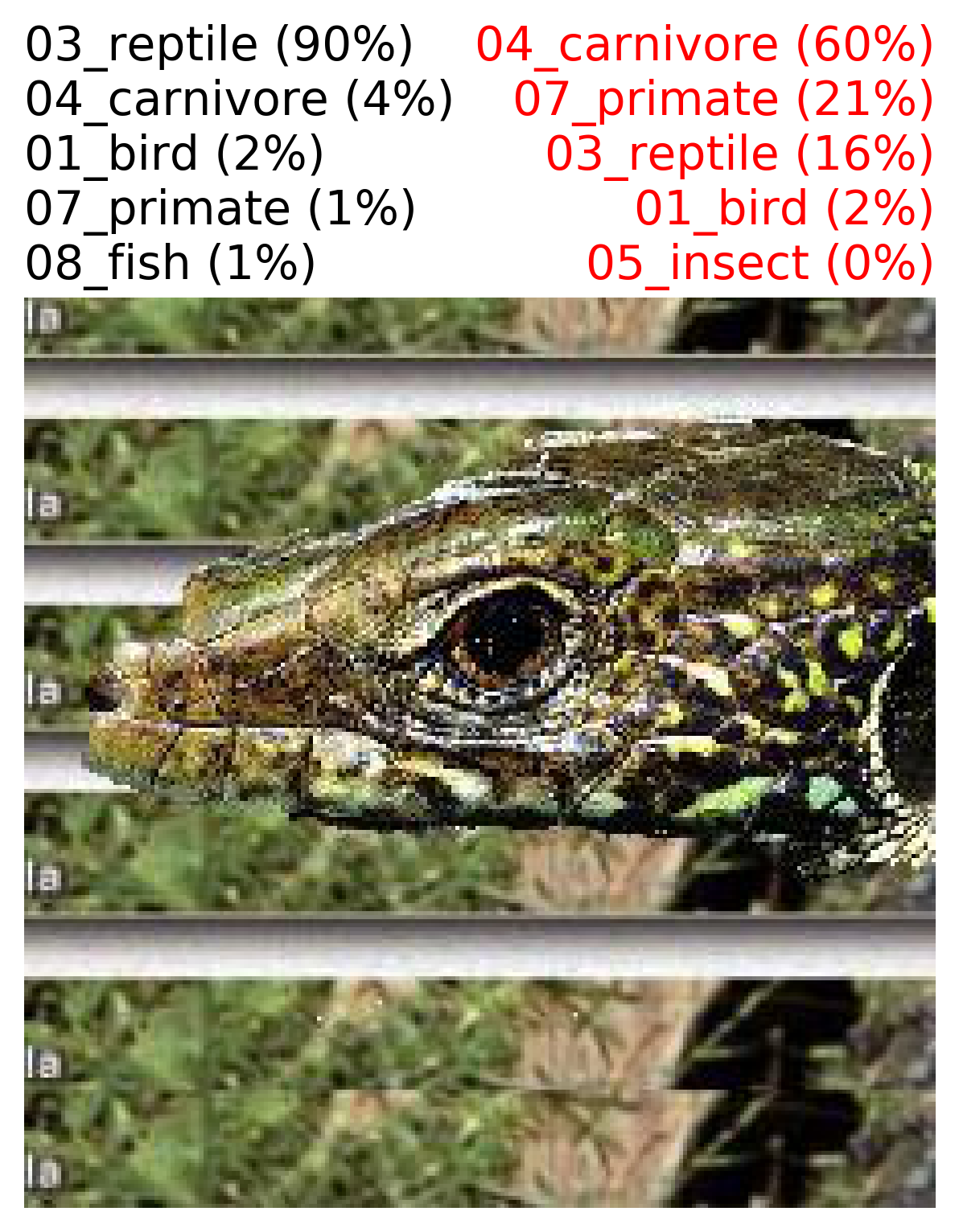} & 
  \includegraphics[width=0.166\linewidth]{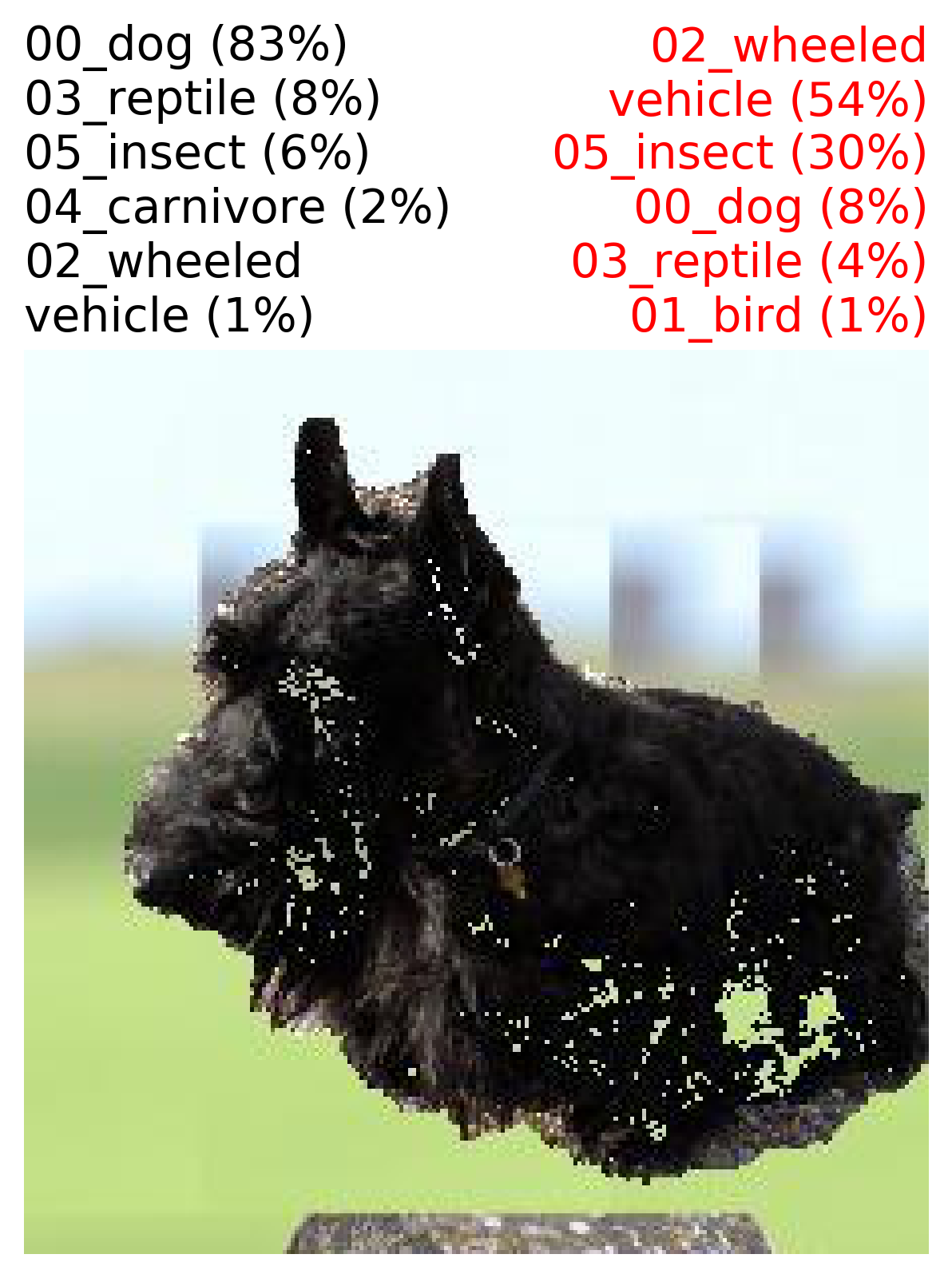} &
  \includegraphics[width=0.166\linewidth]{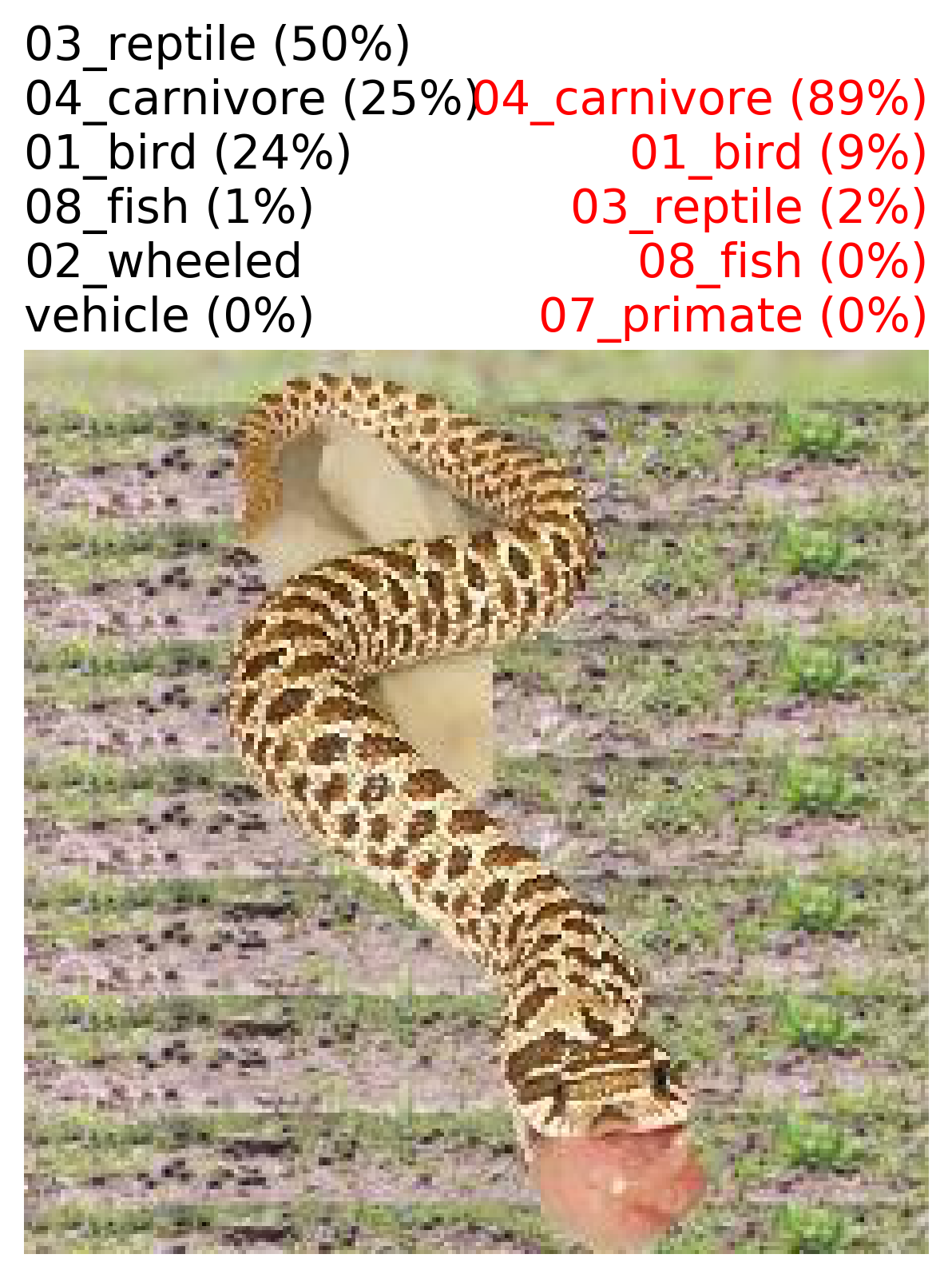} & 
  \includegraphics[width=0.166\linewidth]{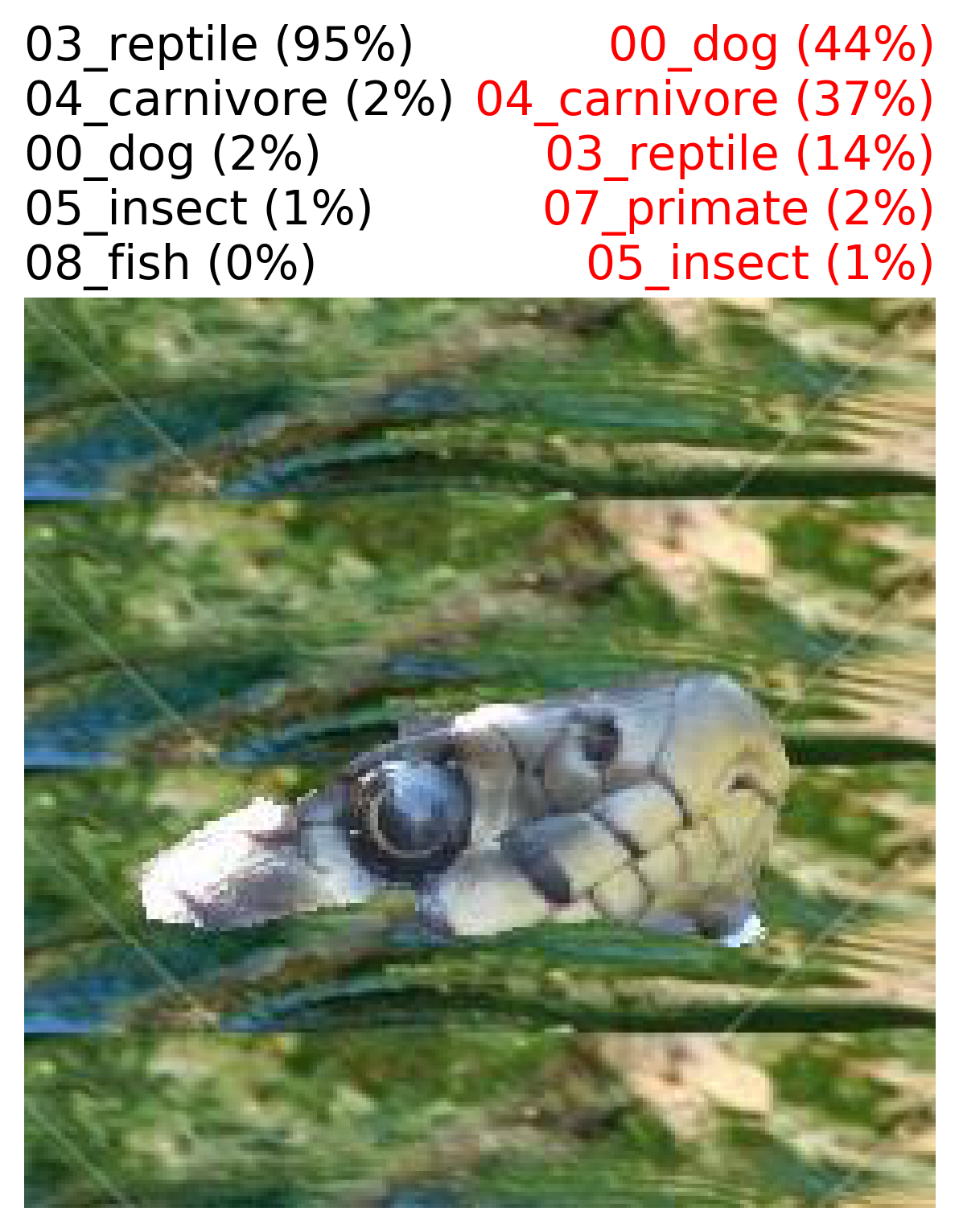} \\
  \includegraphics[width=0.166\linewidth]{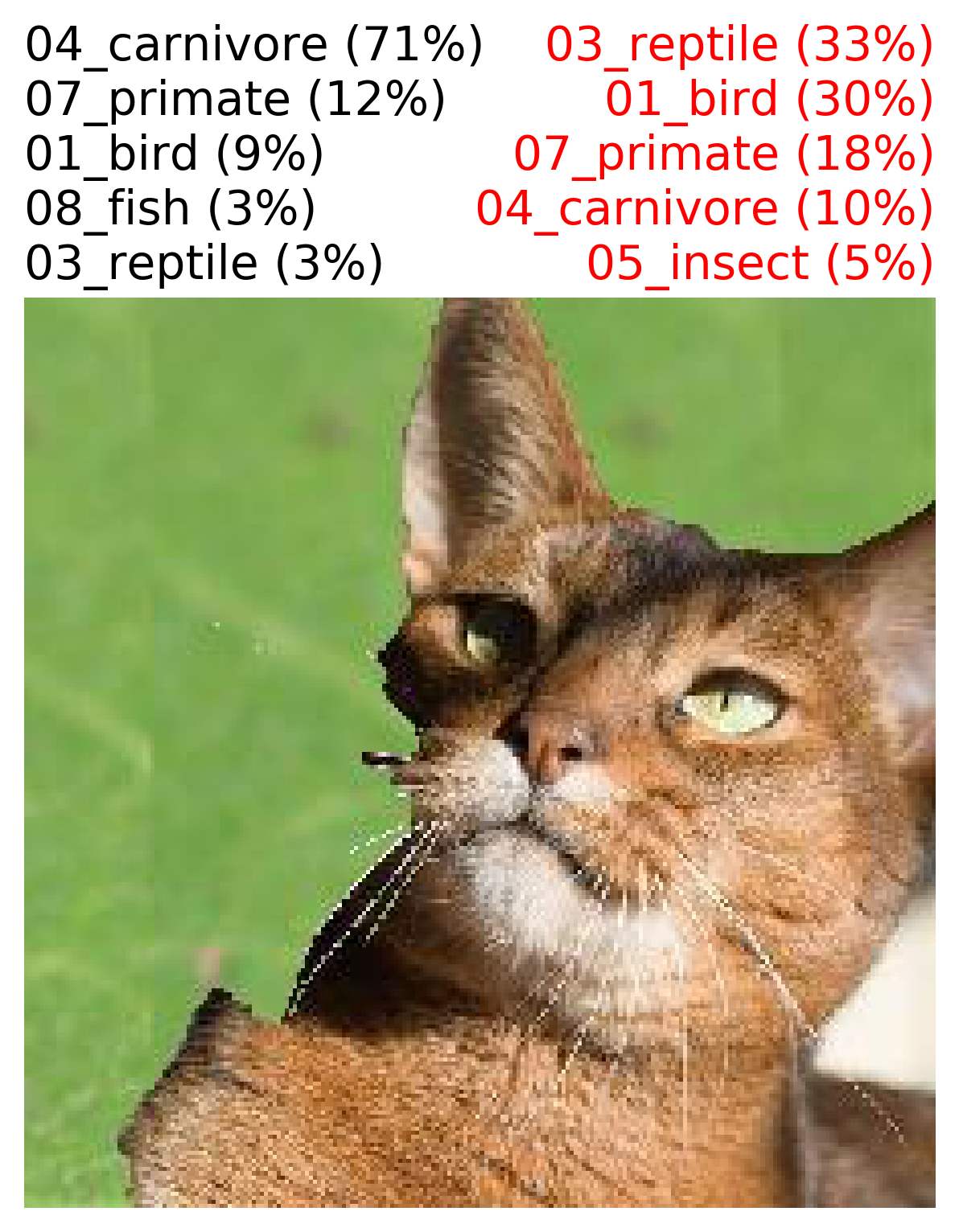} & 
  \includegraphics[width=0.166\linewidth]{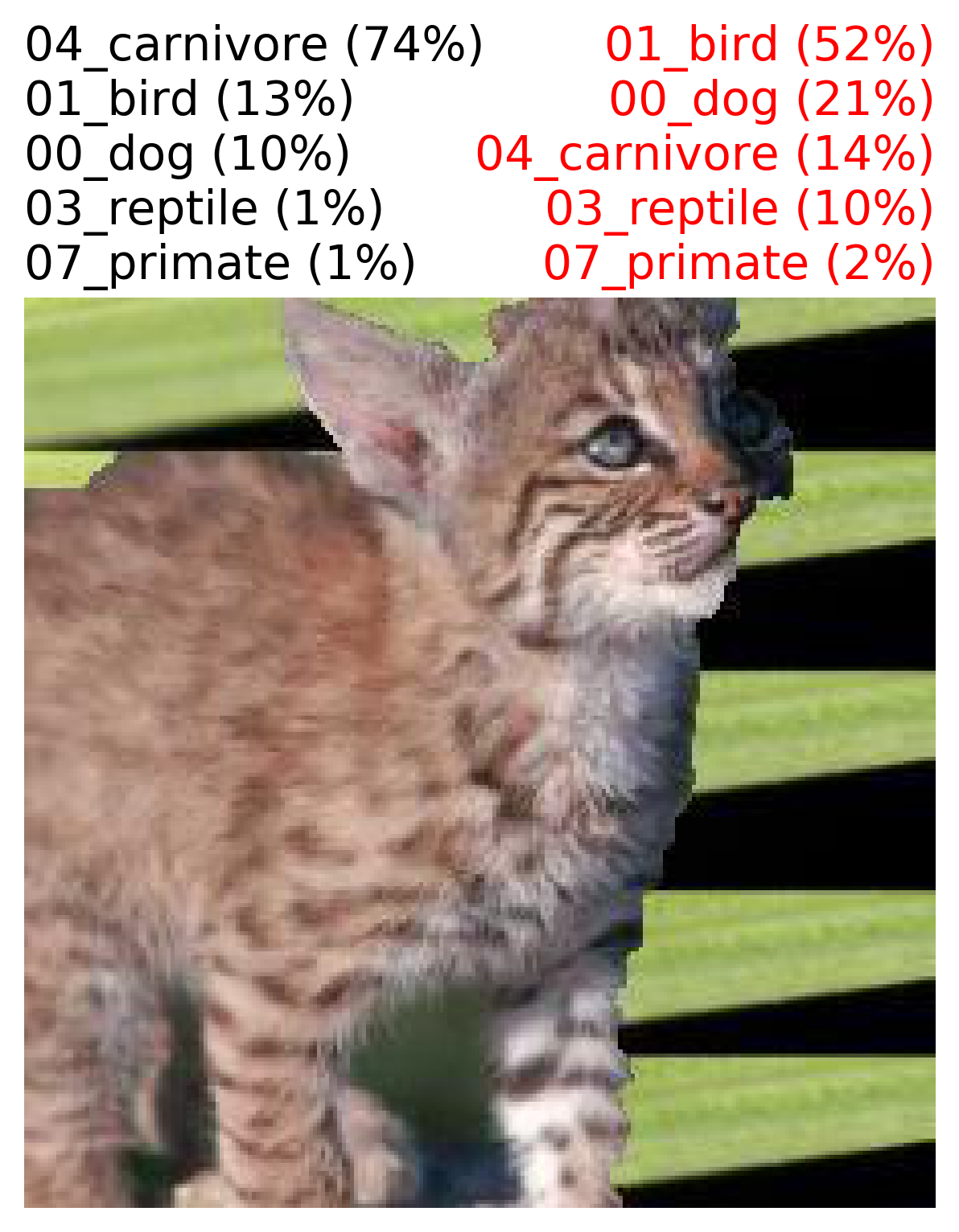} & 
  \includegraphics[width=0.166\linewidth]{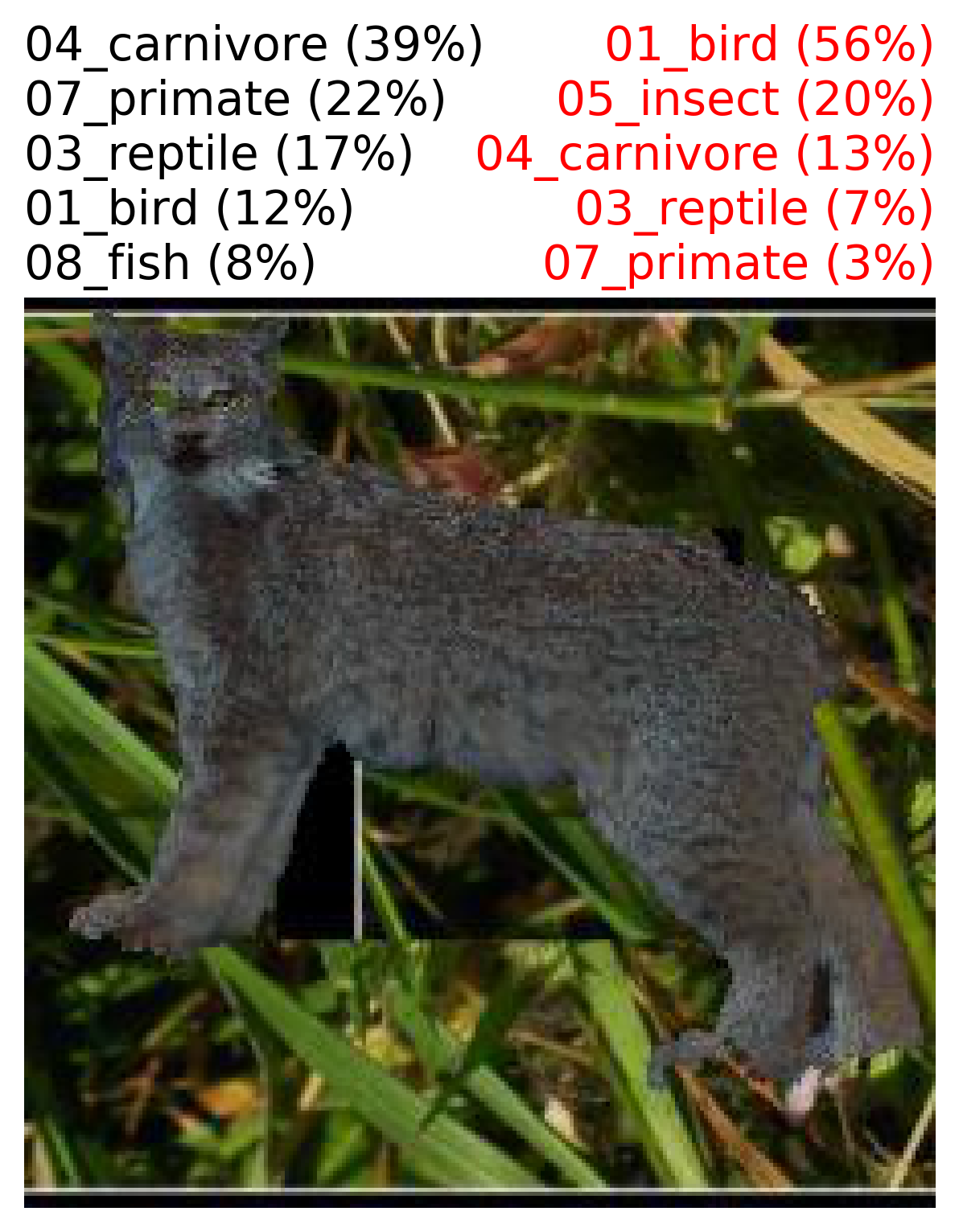} & 
  \includegraphics[width=0.166\linewidth]{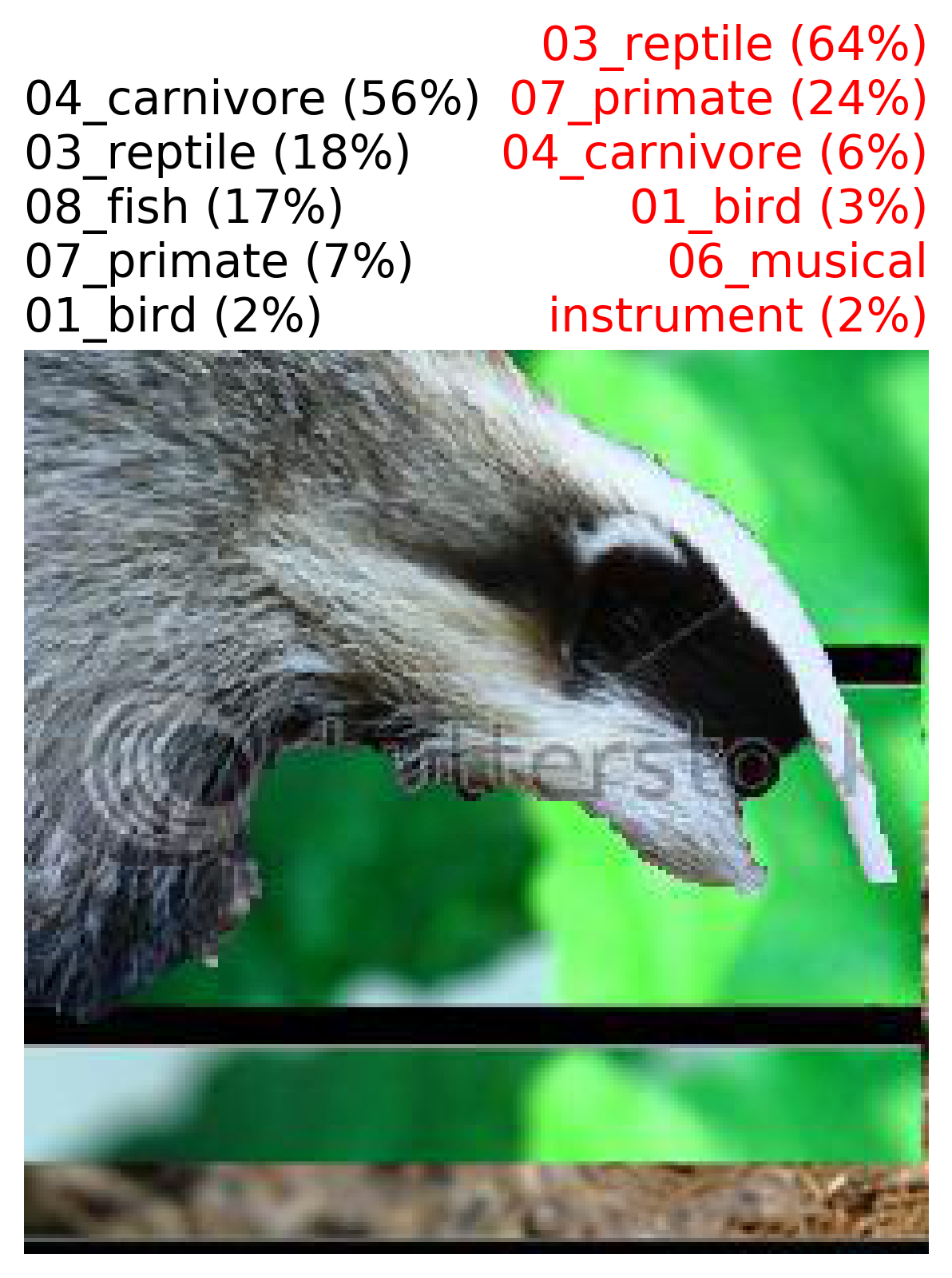} &
  \includegraphics[width=0.166\linewidth]{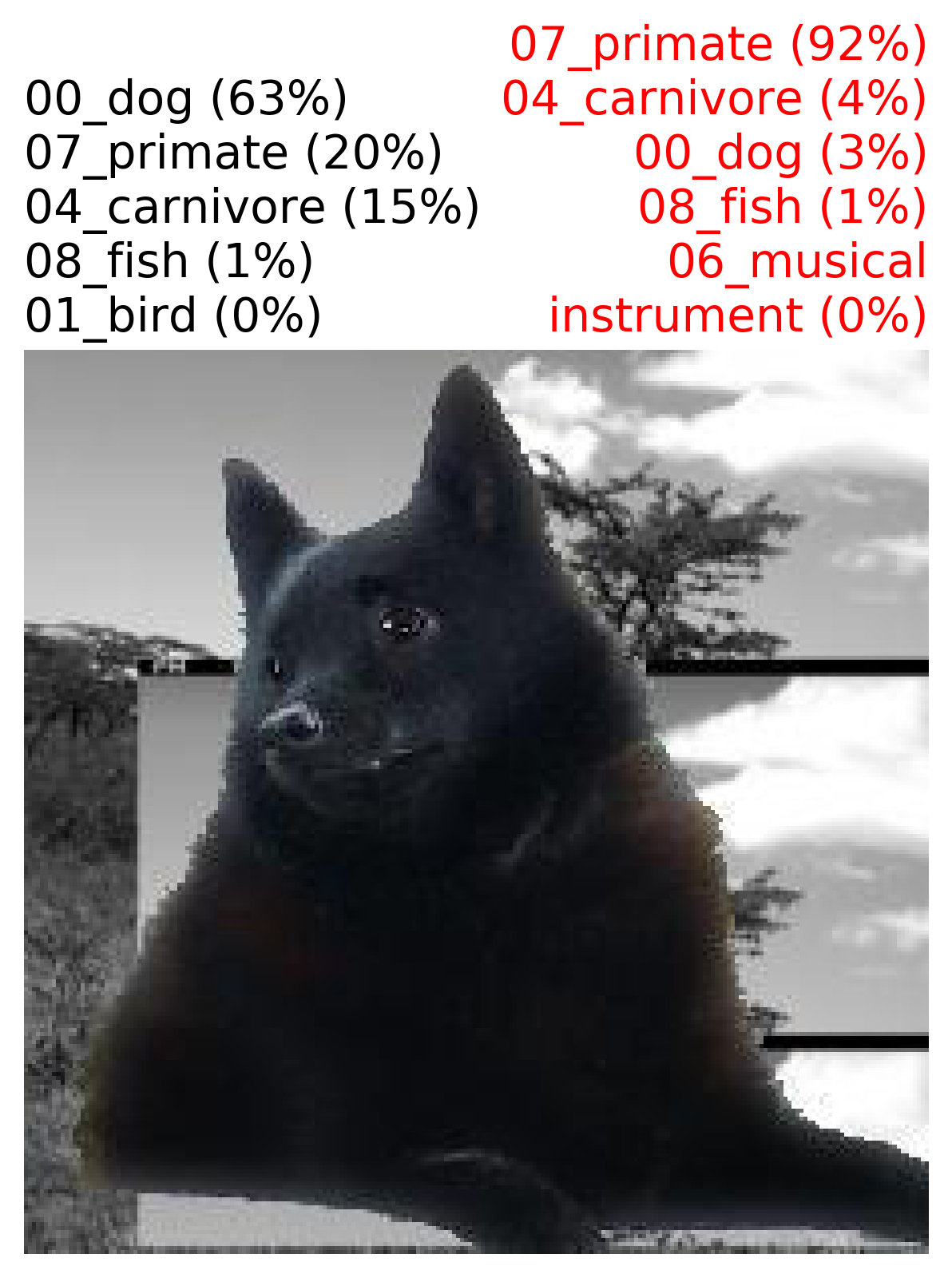} & 
  \includegraphics[width=0.166\linewidth]{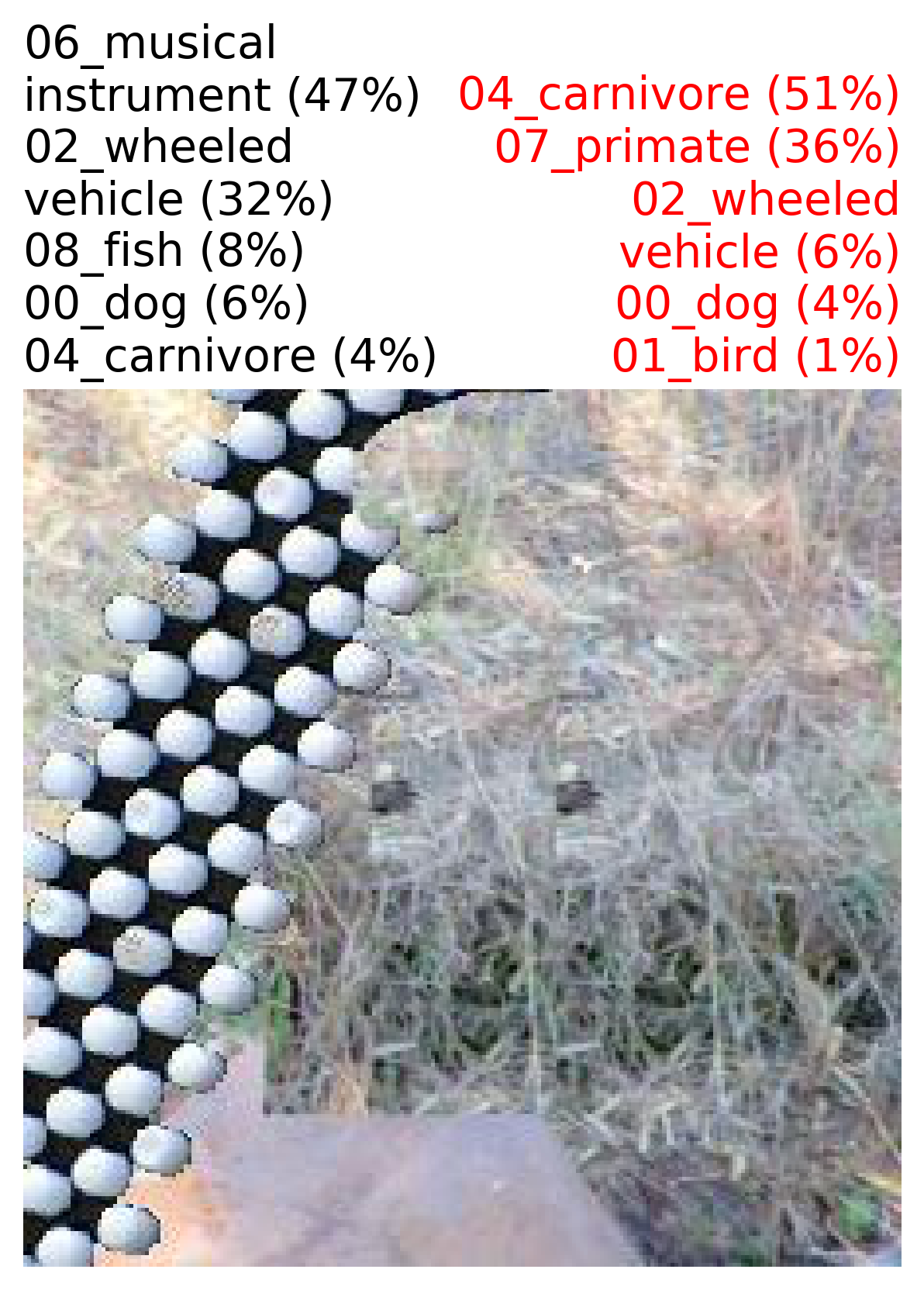} \\
\end{tabular}
\label{fig:in9_winning_cases_rand}
\end{figure*}

\begin{figure*}[tbp]
\caption{Random examples of CCT Trans-Test that the best model (+CF(Tile)+Sal) predicts correctly but Original model fails. 
The grey and red text are the top 5 predictions from the best and Original model's predictions respectively.}
\centering
\begin{tabular}{cccccc}
  \includegraphics[width=0.166\linewidth]{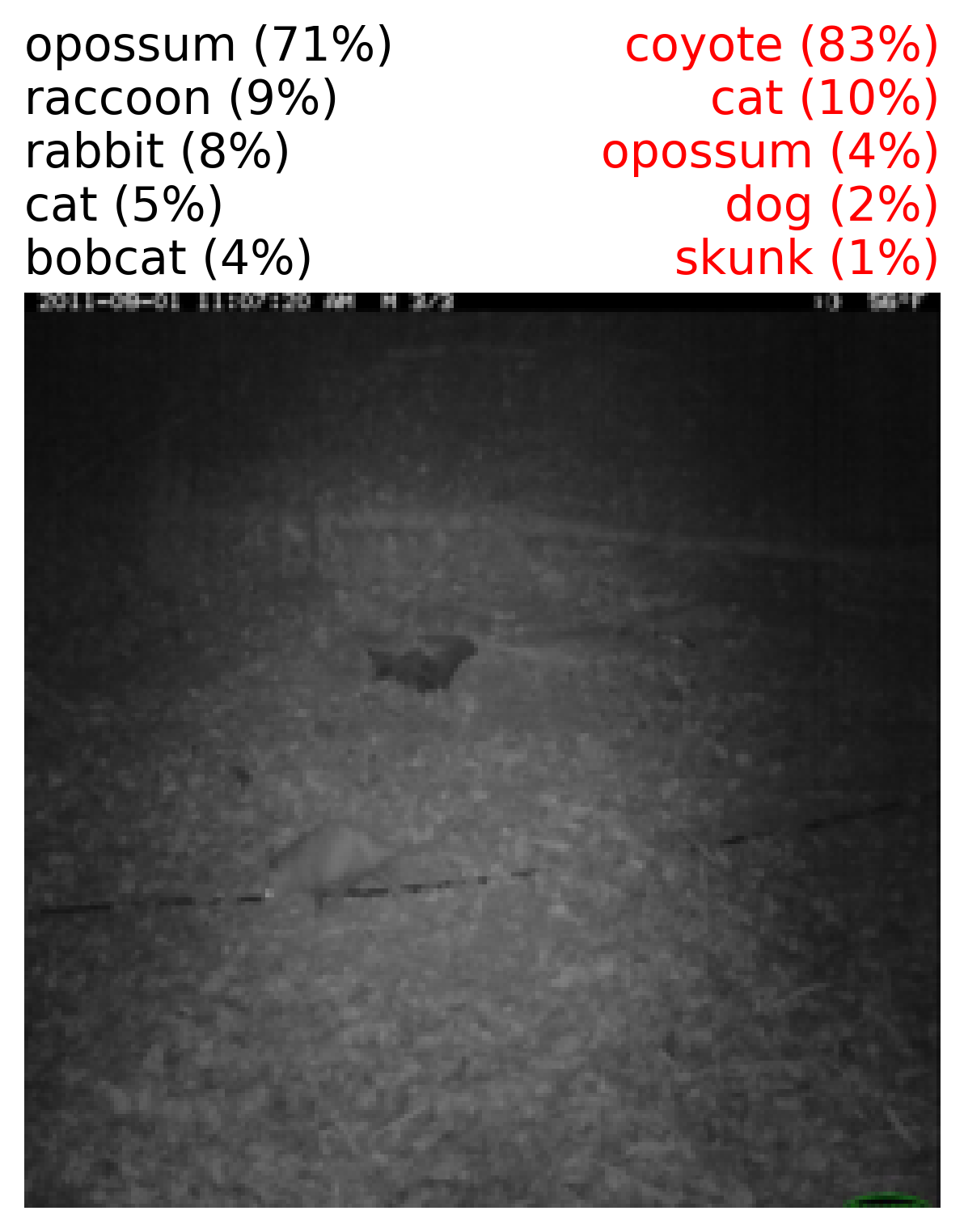} & 
  \includegraphics[width=0.166\linewidth]{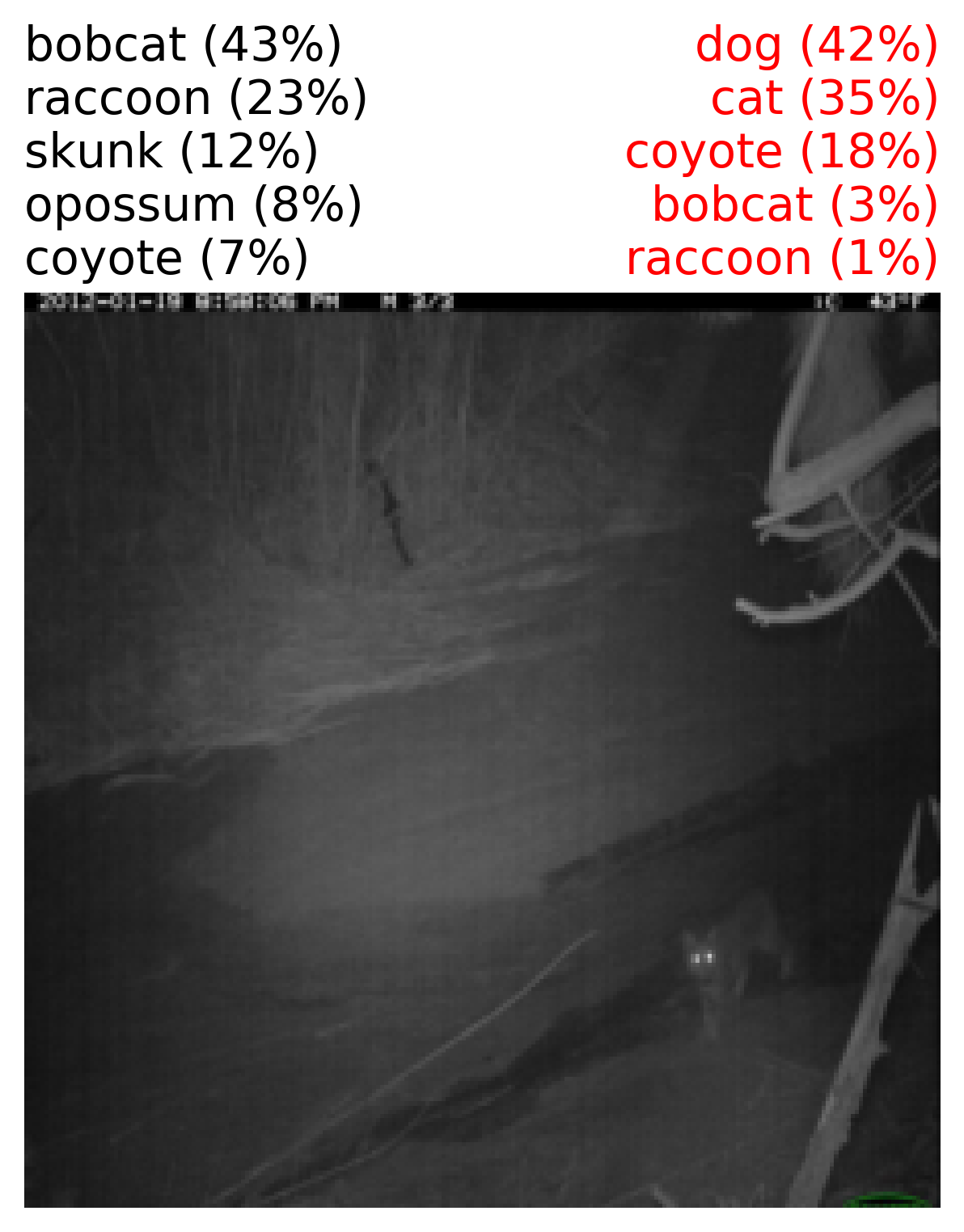} &
  \includegraphics[width=0.166\linewidth]{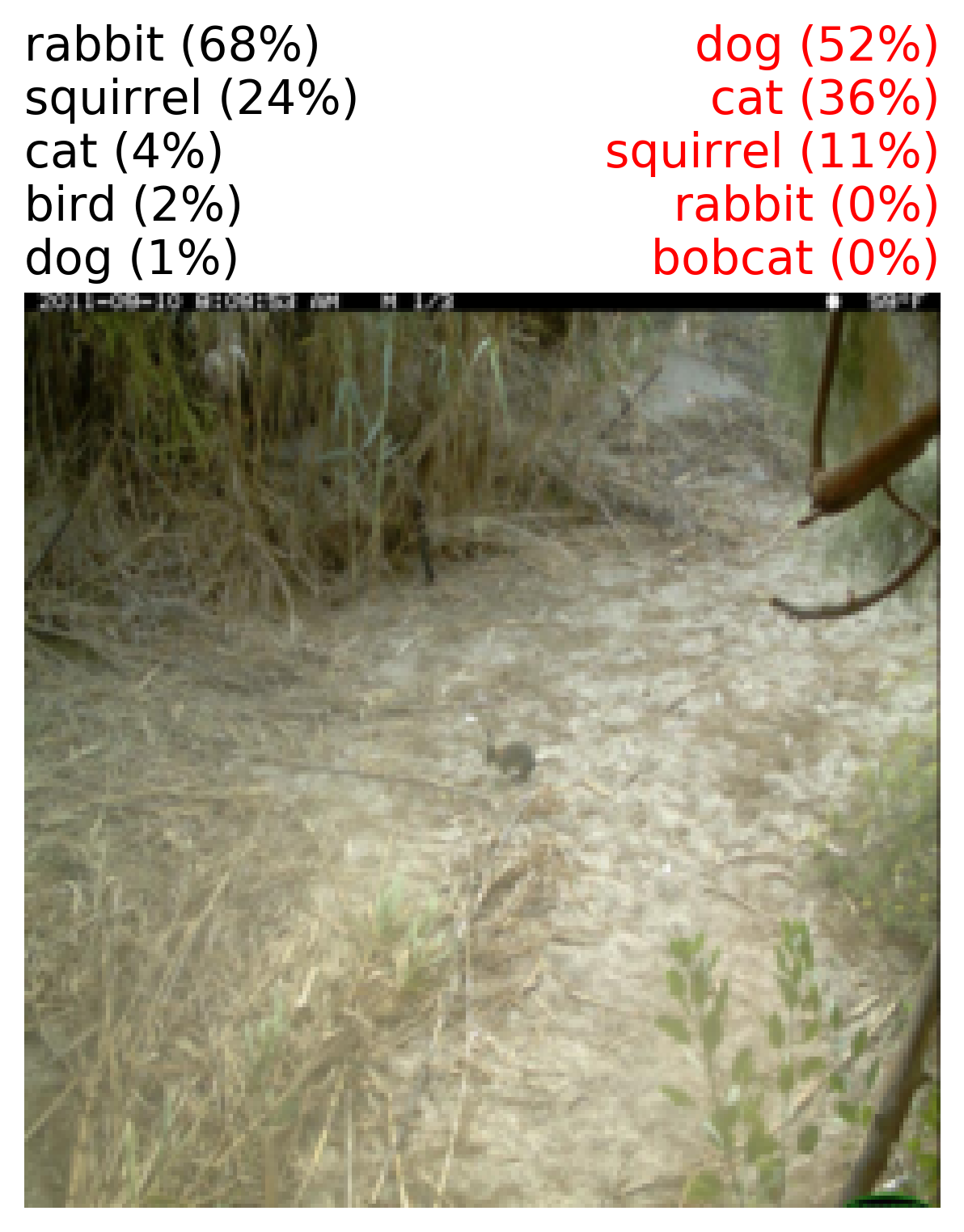} & 
  \includegraphics[width=0.166\linewidth]{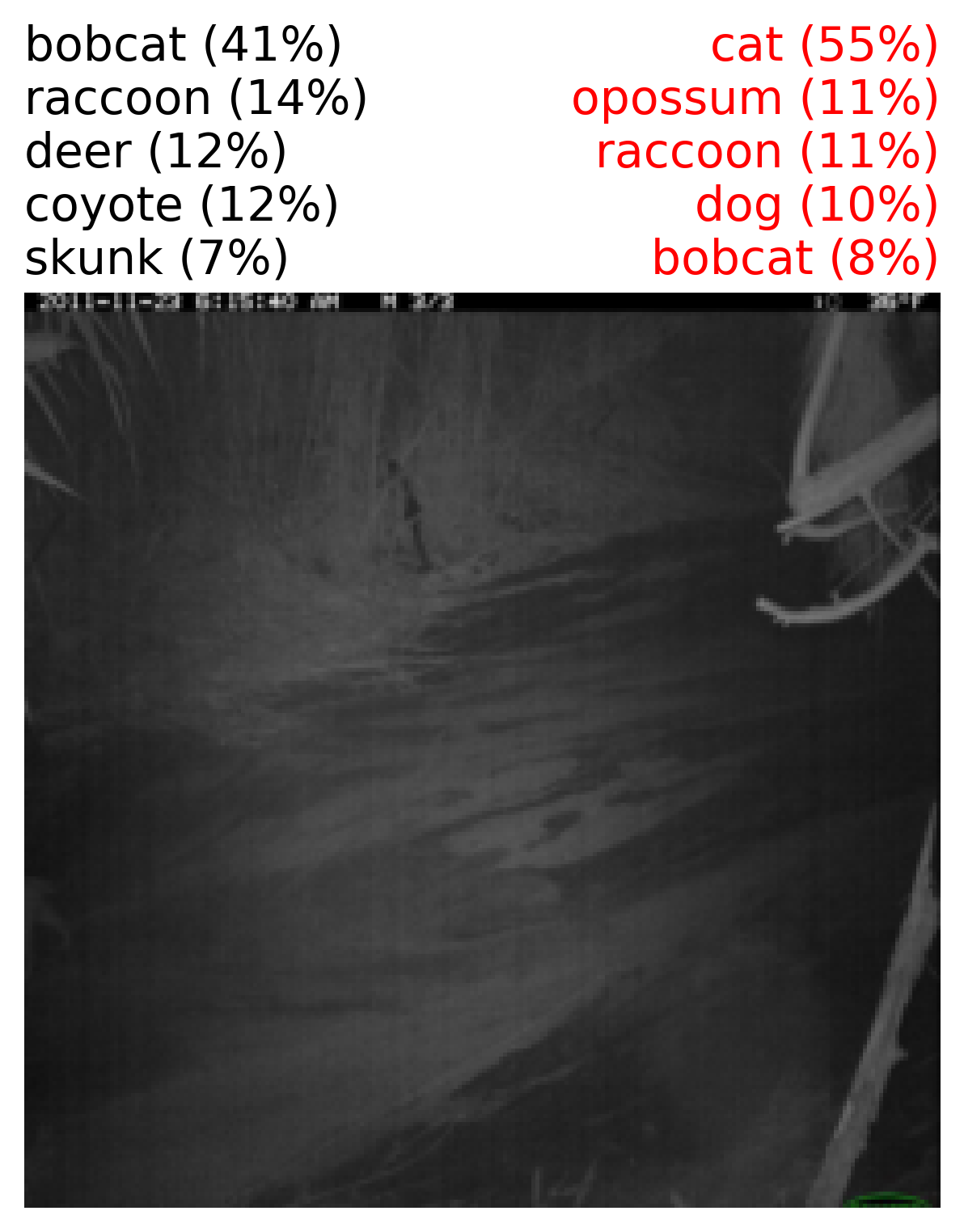} & 
  \includegraphics[width=0.166\linewidth]{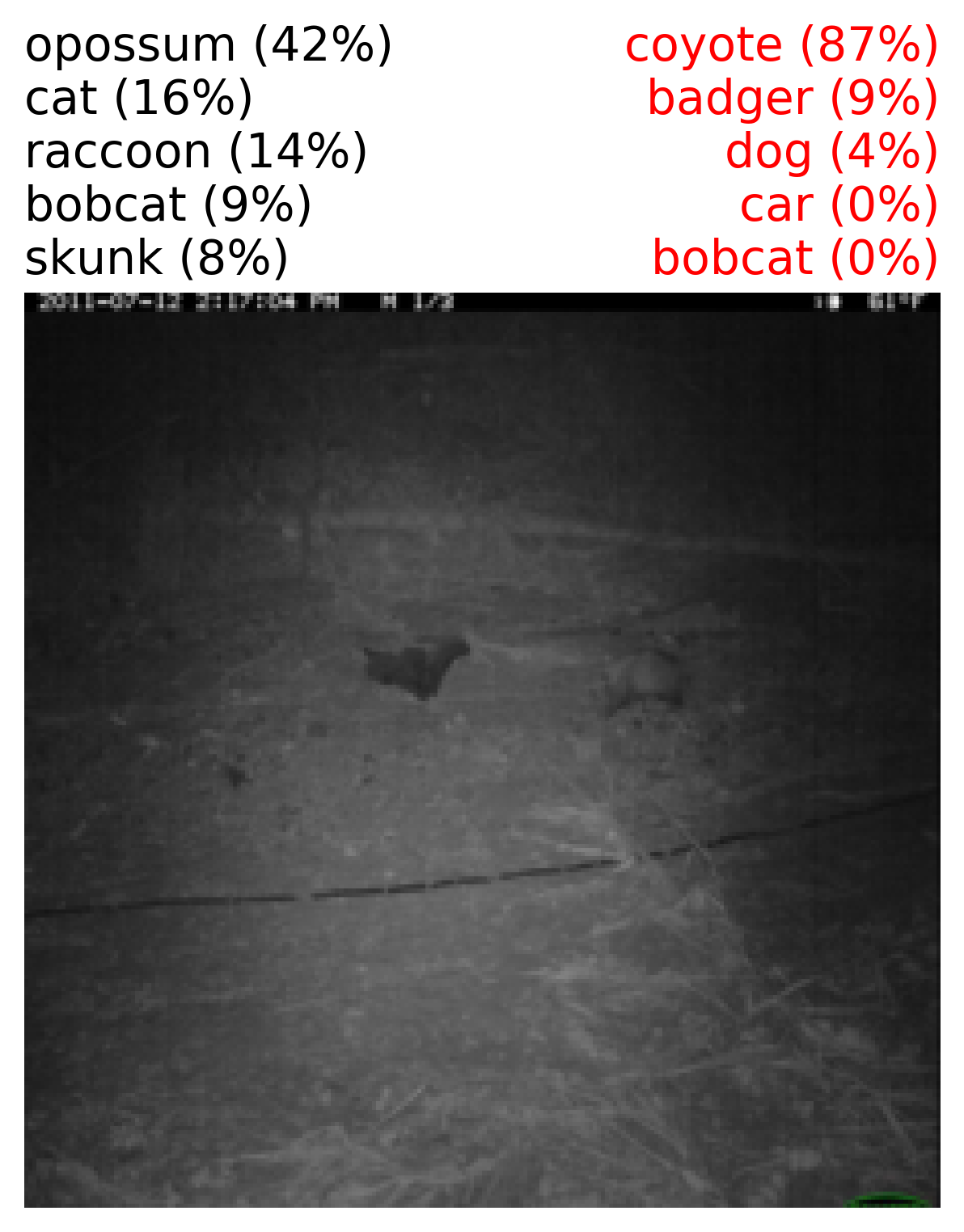} &
  \includegraphics[width=0.166\linewidth]{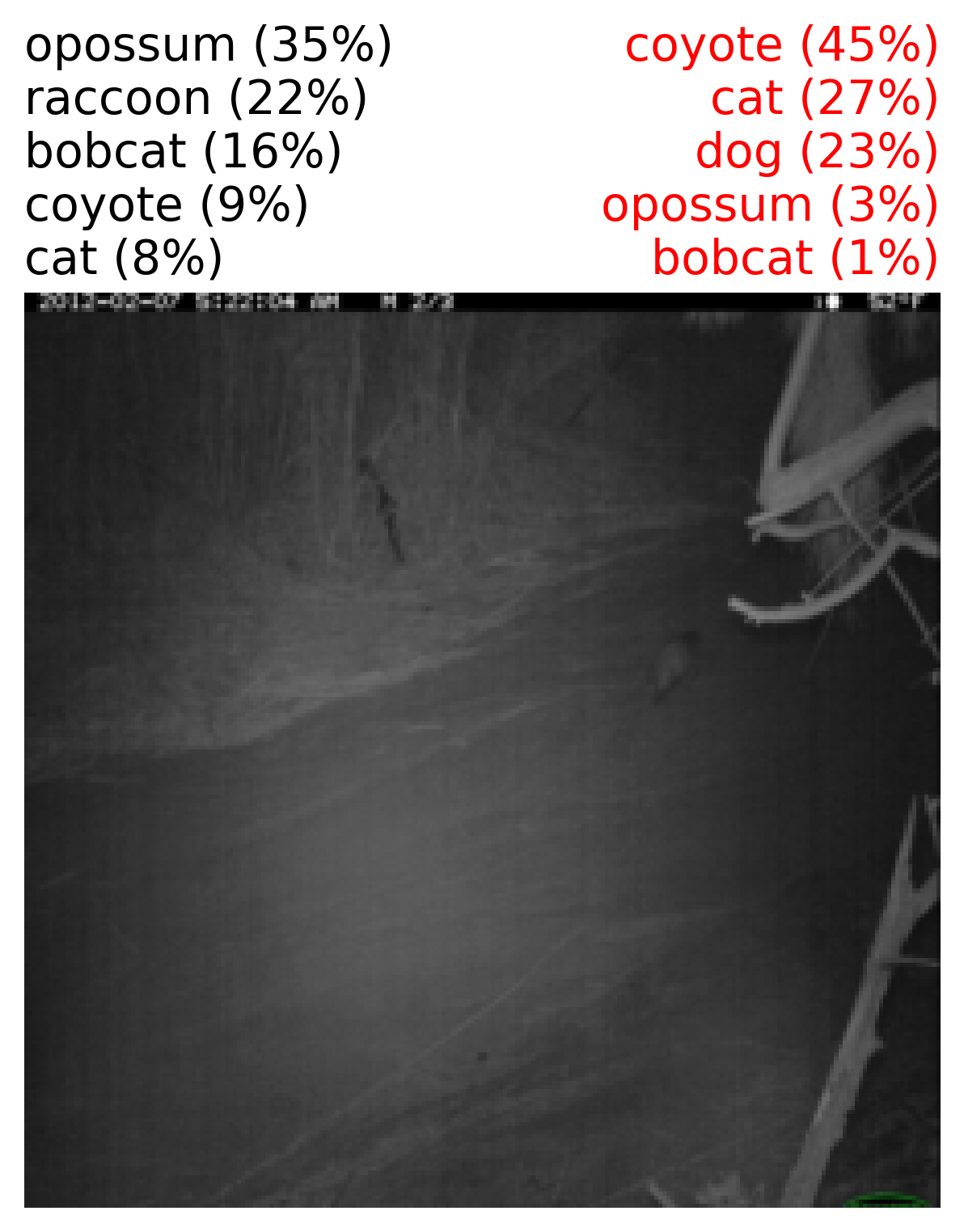} \\ 
  \includegraphics[width=0.166\linewidth]{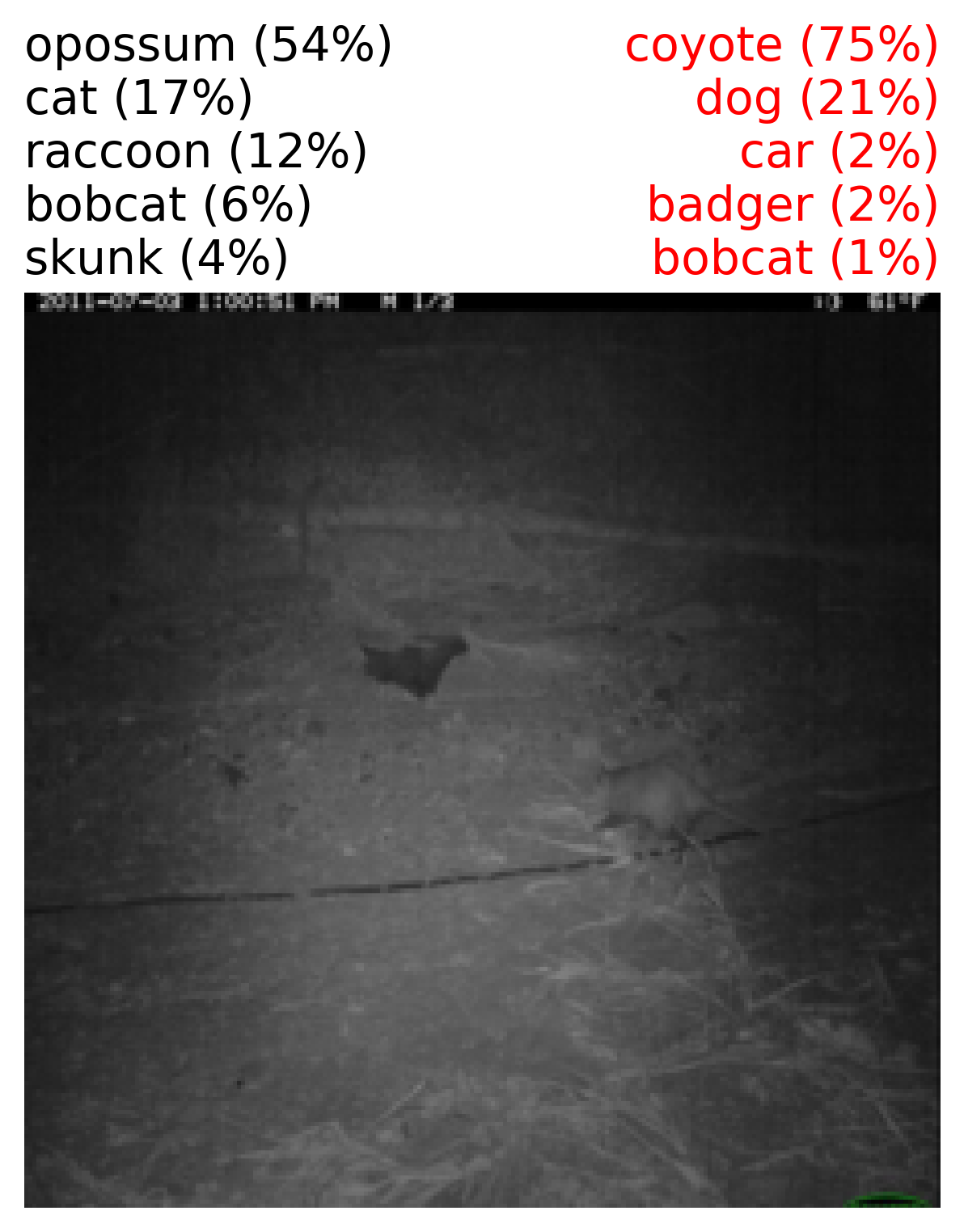} & 
  \includegraphics[width=0.166\linewidth]{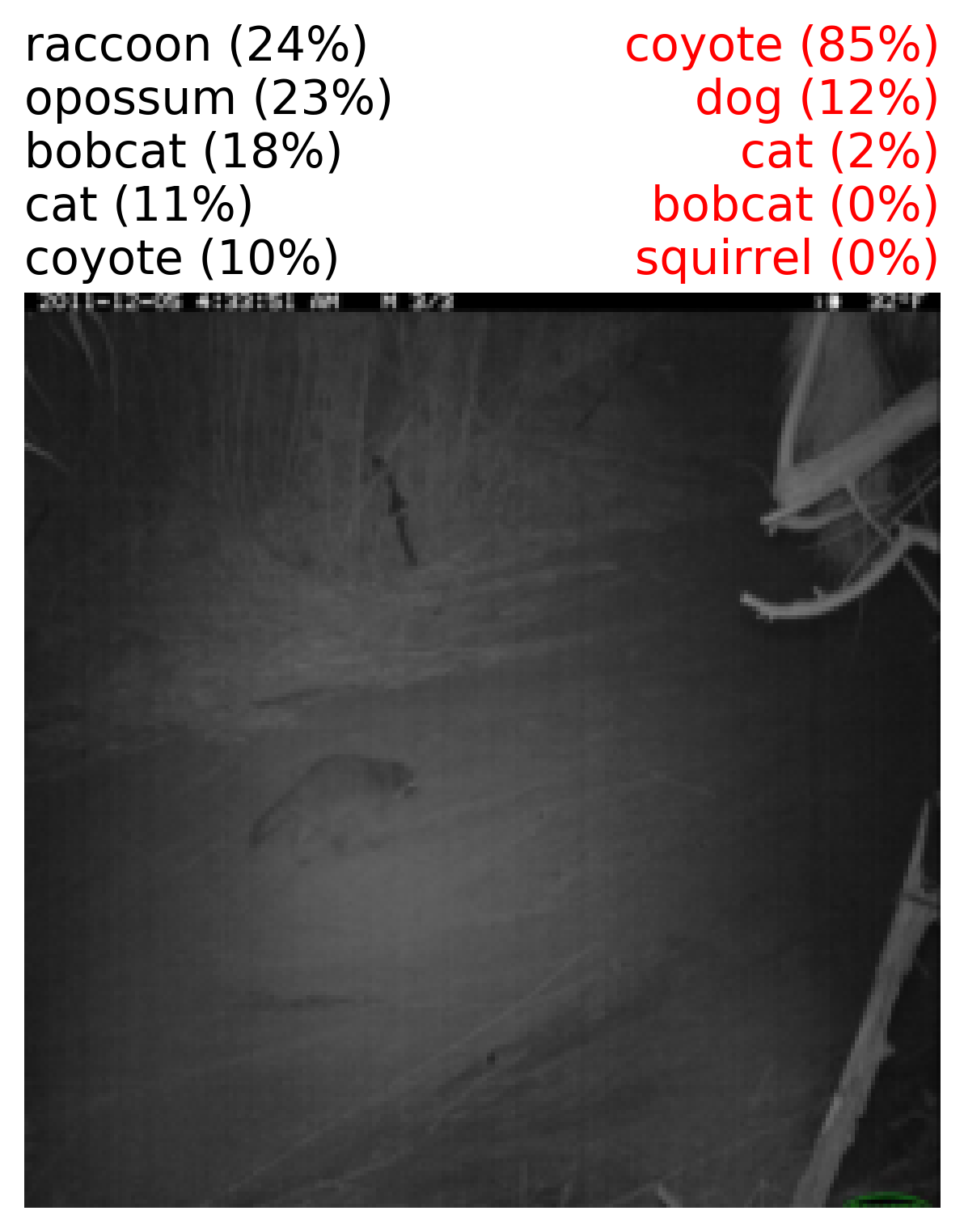} &
  \includegraphics[width=0.166\linewidth]{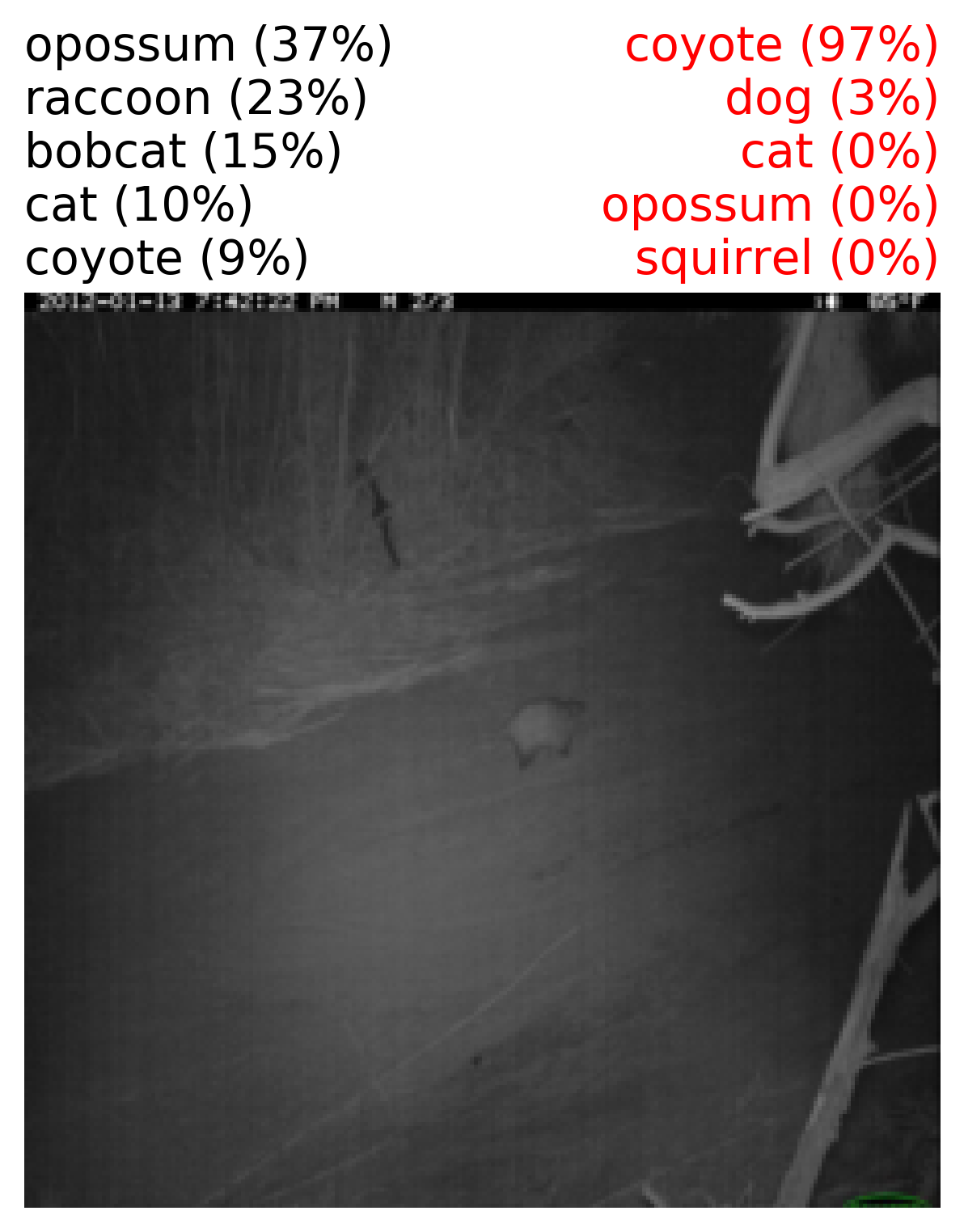} & 
  \includegraphics[width=0.166\linewidth]{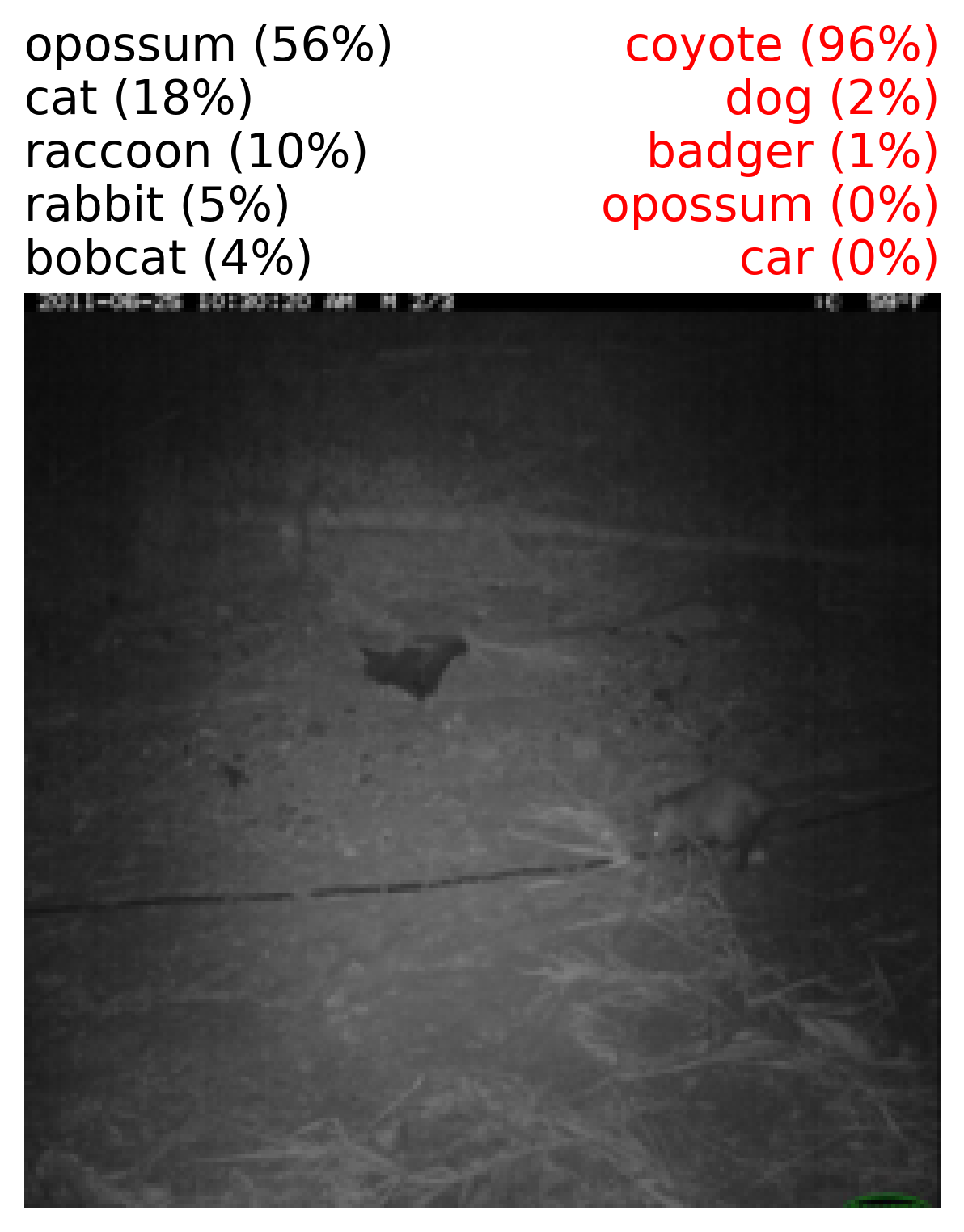} & 
  \includegraphics[width=0.166\linewidth]{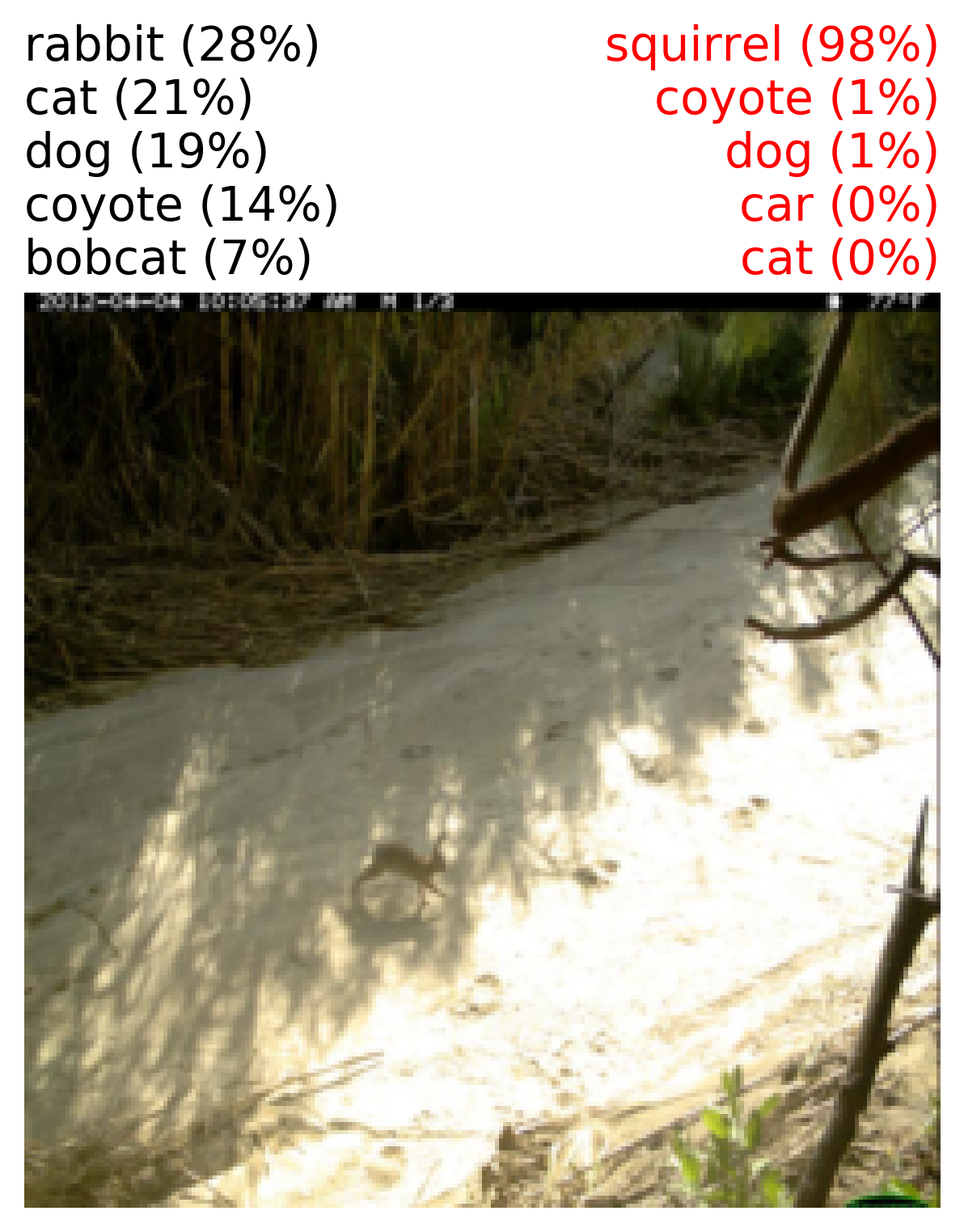} &
  \includegraphics[width=0.166\linewidth]{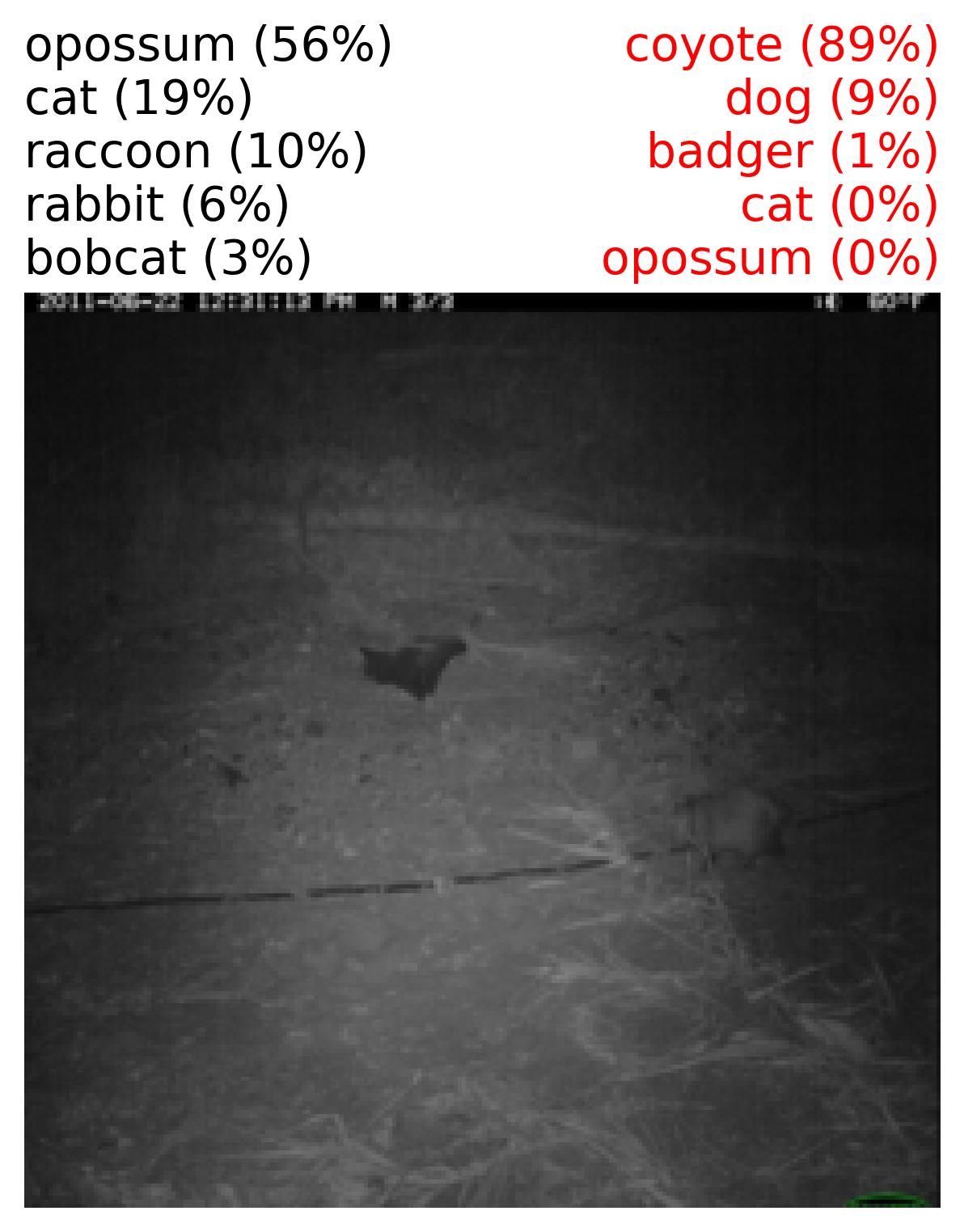} \\
\end{tabular}
\label{fig:cct_winning_cases_rand}
\end{figure*}

We \emph{randomly} pick images in IN9 Mixed-Next test set that our best model (CF(CAGAN)+F(Shuffle)+Sal) predicts correctly while Original model fails in Figure~\ref{fig:in9_winning_cases_rand}.
We also \emph{randomly} pick images in CCT Trans test set that our best model (CF(Tile)+Sal) predicts correctly while Original model fails in Figure~\ref{fig:cct_winning_cases_rand}.

\setlength\tabcolsep{0pt} 
\setlength\tabcolsep{6pt} 

\setlength\tabcolsep{0pt} 
\setlength\tabcolsep{6pt} 

\end{document}